\documentclass[11pt]{article}
\usepackage{amsthm}
\usepackage{graphicx} 
\graphicspath{{./}{../}}
\usepackage{array} 
\usepackage{mathtools}
\usepackage{amsmath, amssymb, amsfonts, verbatim,bbm}
\usepackage{hyphenat,epsfig,subcaption,multirow}
\usepackage{nicefrac}
\usepackage{paralist}
\usepackage[utf8]{inputenc}
\usepackage{algorithm}

\usepackage{booktabs,tabularx,array,threeparttable}

\usepackage[font=small, labelfont=bf]{caption}

\usepackage[dvipsnames,table]{xcolor}

\DeclareFontFamily{U}{mathx}{\hyphenchar\font45}
\DeclareFontShape{U}{mathx}{m}{n}{
      <5> <6> <7> <8> <9> <10>
      <10.95> <12> <14.4> <17.28> <20.74> <24.88>
      mathx10
      }{}
\DeclareSymbolFont{mathx}{U}{mathx}{m}{n}
\DeclareMathSymbol{\bigtimes}{1}{mathx}{"91}
\usepackage{tcolorbox}
\tcbuselibrary{skins,breakable}
\tcbset{enhanced jigsaw}

\usepackage[normalem]{ulem}
\usepackage[compact]{titlesec}

\definecolor{DarkRed}{rgb}{0.1,0.1,0.8}
\definecolor{DarkBlue}{rgb}{0.1,0.1,0.5}

\usepackage{nameref}
\definecolor{ForestGreen}{rgb}{0.1333,0.5451,0.1333}
\definecolor{Red}{rgb}{0.9,0,0}
\usepackage[linktocpage=true,
	pagebackref=true,colorlinks,
		urlcolor=black,
	linkcolor=DarkRed, citecolor=ForestGreen,
	bookmarks,bookmarksopen,bookmarksnumbered]
	{hyperref}
\usepackage{aliascnt}
\usepackage[noabbrev,nameinlink]{cleveref}
\crefname{property}{property}{Property}
\creflabelformat{property}{(#1)#2#3}
\crefname{equation}{eq}{Eq}
\creflabelformat{equation}{(#1)#2#3}

\usepackage{bm}
\usepackage{url}
\usepackage{xspace}
\usepackage[mathscr]{euscript}
\usepackage{float}
\usepackage[numbers]{natbib}
\AtBeginDocument{\let\cite\citep}

\setcitestyle{maxcitenames=2} 

\usepackage{tikz}
\usetikzlibrary{arrows}
\usetikzlibrary{arrows.meta}
\usetikzlibrary{shapes}
\usetikzlibrary{backgrounds}
\usetikzlibrary{positioning}
\usetikzlibrary{decorations.markings}
\usetikzlibrary{patterns}
\usetikzlibrary{calc}
\usetikzlibrary{fit}
\usetikzlibrary{decorations.pathmorphing}

\usepackage{mdframed}

\usepackage[noend]{algpseudocode}
\makeatletter
\def\BState{\State\hskip-\ALG@thistlm}
\makeatother

\usepackage{cite}
\usepackage{enumitem}
\usepackage{authblk}
\usepackage[margin=1in]{geometry}

\newtheorem{theorem}{Theorem}

\newtheorem{lemma}{Lemma}[section]

\newtheorem{proposition}[lemma]{Proposition}
\newtheorem{corollary}[lemma]{Corollary}
\newtheorem{claim}[lemma]{Claim}

\newtheorem{definition}{Definition}[section]
\crefname{definition}{Definition}{Definitions}
\Crefname{definition}{Definition}{Definitions}

\newtheorem{problem}{Problem}

\newtheorem*{claim*}{Claim}
\newtheorem*{proposition*}{Proposition}
\newtheorem*{lemma*}{Lemma}
\newtheorem*{problem*}{Problem}

\crefname{lemma}{Lemma}{Lemmas}
\crefname{claim}{Claim}{Claims}

\crefname{claim}{Claim}{Claims}

\newtheorem{hassumption}{Hardness Assumption}
\crefname{hassumption}{Assumption}{Assumptions}

\theoremstyle{definition}
\newaliascnt{remark}{lemma}
\newtheorem{remark}[remark]{Remark}
\aliascntresetthe{remark}
\crefname{remark}{Remark}{Remarks}
\Crefname{remark}{Remark}{Remarks}

\newtheoremstyle{restate}{}{}{\itshape}{}{\bfseries}{~(restated).}{.5em}{\thmnote{#3}}
\theoremstyle{restate}
\newtheorem*{restate}{}

\theoremstyle{definition}
\newtheorem{mdalg}{Algorithm}

\allowdisplaybreaks

\renewcommand{\qed}{\nobreak \ifvmode \relax \else
      \ifdim\lastskip<1.5em \hskip-\lastskip
      \hskip1.5em plus0em minus0.5em \fi \nobreak
      \vrule height0.75em width0.5em depth0.25em\fi}

\newcommand{\E}{\mathbb{E}}
\newcommand{\Var}{\text{Var}}

\newcommand{\Z}{\mathbb{Z}}
\newcommand{\R}{\mathbb{R}}

\newcommand{\eps}{\varepsilon}

\newcommand{\sign}{\text{sign}}

\def\zo{0\textnormal{-}1}
\def\prefix{\textnormal{pfx}}

\def\simiid{\sim_{iid}}

\newcommand{\norm}[1]{\lVert #1 \rVert}

\DeclareSymbolFont{rsfs}{U}{rsfs}{m}{n}
\DeclareSymbolFontAlphabet{\mathscrsfs}{rsfs}

\def\VCdim{\mathrm{VCdim}}

\def\spn{{\rm span}}

\def\lin{{\rm lin}}

\def\Unif{{\rm Unif}}
\def\lin{{\rm lin}}

\def\spn{{\rm span}}

\def\cV{{\mathcal V}}
\def\cG{{\mathcal G}}

\def\cP{{\mathcal P}}

\def\cC{{\mathcal C}}

\def\cL{{\mathcal L}}
\def\cF{{\mathcal F}}

\def\cS{{\mathcal S}}

\def\cV{{\mathcal V}}
\def\cG{{\mathcal G}}

\def\cP{{\mathcal P}}

\def\cH{{\mathcal H}}
\def\cA{{\mathcal A}}

\def\M{\mathsf{M}_\star}

\def\reals{{\mathbb R}}

\def\naturals{{\mathbb N}}

\def\poly{\mathrm{poly}}

\def\reals{{\mathbb R}}

\def\naturals{{\mathbb N}}

\def\th{\tilde{h}}

\def\cD{{\mathcal D}}
\def\cX{{\mathcal X}}
\def\cY{{\mathcal Y}}

\def\passk{\mathsf{pass@}k}
\def\passkappa{\mathsf{pass@}\kappa}
\def\cF{{\mathcal F}}
\def\cS{{\mathcal S}}

\def\cZ{{\mathcal Z}}

\def\lin{{\rm lin}}

\def\ind{\mathbbm{1}}

\def\th{\textnormal{thr}}

\def\of{\overline{f}}
\def\sCoT{{\sf CoT}}
\def\sete{{\sf e2e}}

\newcommand{\ete}[2]{#1^{\sete\textnormal{-}{#2}}}

\newcommand{\CoT}[2]{#1^{\sCoT\textnormal{-}{#2}}}

\newcommand{\Cons}{\textnormal{\textsc{Consistent}}}

\newcommand{\Conscot}{\Cons_{\sCoT}}
\newcommand{\Consete}{\Cons_{\sete}}

\def\zo{\textnormal{0-1}}
\def\iid{\sim_{iid}}

\def\cFlin{\mathcal{F}_{\scriptscriptstyle d,\lin}}

\newcommand{\abs}[1]{\left\lvert{#1}\right\rvert}

\DeclareMathOperator{\Bern}{Bernoulli}

\DeclareMathOperator{\Score}{score}
\DeclareMathOperator{\Bij}{Bij}
\newcommand{\F}{\mathbb{F}}
\newcommand{\id}{\mathsf{id}}
\newcommand{\z}{\mathfrak{z}}

\newcommand{\gapsvp}{\mathsf{GapSVP}}
\newcommand{\sivp}{\mathsf{SIVP}}
\newcommand{\klocal}{K}

\newcommand{\NEARMID}{\operatorname{NEAR-MID}}
\newcommand{\dsum}{\operatorname{sum}}
\newcommand{\scrC}{\mathcal{C}}

\usepackage{bm}

\def\vb{{b}}

\def\vg{{g}}

\def\vp{{p}}

\def\vv{{u}}
\def\vw{{w}}
\def\vx{{x}}

\def\vz{{z}}

\def\gA{{\mathcal{A}}}
\def\gB{{\mathcal{B}}}

\def\gD{{\mathcal{D}}}

\def\gF{{\mathcal{F}}}
\def\gG{{\mathcal{G}}}
\def\gH{{\mathcal{H}}}

\def\gL{{\mathcal{L}}}

\def\gS{{\mathcal{S}}}

\def\gU{{\mathcal{U}}}

\def\gX{{\mathcal{X}}}
\def\gY{{\mathcal{Y}}}

\renewcommand{\r}{\right}
\renewcommand{\l}{\left}

\newcommand{\ellcons}{\ell_{\text{cons}}}
\newcommand{\vxi}{\vx^{(i)}}

\newcommand{\ord}[1]{#1^{\mathrm{th}}}
\def\thr{\textnormal{thr}}
\newcommand{\summ}[2]{\sum_{#1 = 1}^{#2}}

\newcommand{\disthard}{\gD_{(N,\klocal)\textnormal{-}\text{hard}}}
\newcommand{\classdnfhard}{\Psi_{N,\klocal}}

\def\val{{\rm val}}

\title{
Learning to Think from Multiple Thinkers:\\[0.1em]
{
How Chain-of-Thought Heterogeneity Affects Efficient Learnability}}

\author[1]{Nirmit Joshi}
\author[2]{ \quad Roey Magen}
\author[1]{\quad Nathan Srebro}
\author[1]{\quad Nikolaos Tsilivis}
\author[2]{\quad Gal Vardi}

\affil[1]{\small Toyota Technological Institute at Chicago. \texttt{\{nirmit,nsrebro,ntsilivis\}@ttic.edu}}
\affil[2]{\small Weizmann Institute of Science. \texttt{\{roey.magen,gal.vardi\}@weizmann.ac.il}}

\date{}
\begin{document}
\maketitle
\begin{abstract}

Chain-of-thought (CoT) supervision makes computationally hard learning problems tractable by exposing intermediate steps that are easy to learn via next-token prediction. However, in practice, CoT datasets draw on heterogeneous sources with different reasoning styles, and substantial effort goes into curating supervision suitable for next-token training. What role does this curation play in enabling the computational advantage of CoT supervision? In this work, we take a first step toward understanding this question from a learning-theoretic perspective. We introduce several models of \emph{heterogeneous CoT supervision} from multiple sources, each providing correct traces with the same final answer but different intermediate steps. Assuming that fitting a next-token predictor to \emph{homogeneous} traces from any single source is computationally tractable, we ask when this suffices for efficient learning from heterogeneous supervision. 

We show \emph{cryptographic hardness} of learning from \emph{passively} collected traces from multiple sources, even though learning from homogeneous CoTs is easy. This hardness can stem either from \emph{hard-to-learn contribution biases} determining which sources contribute traces on different parts of the domain, or from receiving an \emph{unfiltered} heterogeneous mixture (though a filtered subset of homogeneous traces is sufficient for efficient learning). On the positive side, inspired by iterative and adaptive fine-tuning of language models, we give a computationally efficient learning algorithm that \emph{actively} and \emph{adaptively} collects CoTs from multiple sources with a per-source CoT budget completely independent of the target error $\eps$. 
We argue that our results illuminate data collection and curation practices central to making CoT supervision effective for learning.
\end{abstract}

\paragraph{AI methodology.} All main technical ideas were developed by the authors without AI assistance. After establishing the structure of the arguments, we used OpenAI models to assist with selected calculations and technical details. We subsequently reviewed and refined the arguments and their presentation. We also used these models to polish the writing and improve its clarity. We take full responsibility for all parts of the paper, including the correctness of results and references.

\newpage
 
\section{Introduction}\label{sec:intro}
Large language models (LLMs) have acquired remarkable capabilities for solving problems that require complex reasoning. Given a question, the models \emph{autoregressively} generate a sequence of tokens, and chaining them together—referred to as Chain of Thought (CoT)—allows them to think. Such behavior is elicited because these models are trained on examples containing rich reasoning traces. Such supervision is explicitly available during supervised fine-tuning (SFT) and implicitly during pretraining and mid-training; see \Cref{sec:related-work} for further discussion.\footnote{Even reinforcement learning (RL), which has driven major recent advances, crucially builds on a foundation acquired during these earlier phases of training \citep[see discussions in][]{chu2025sft,guo2025deepseek}.} CoT supervision is thus 
a key mechanism
through which modern LLMs acquire complex reasoning abilities and has led both industry labs \cite[e.g.,][]{guo2025deepseek,abdin2025phi} and the open-source community \cite[e.g.,][]{li2024numinamath,guha2025openthoughts} to invest heavily in constructing high-quality CoT datasets.


    On the theoretical front, a growing line of work studies how CoT supervision enables efficient learning of complex reasoning \citep{malach2023auto,joshi2025theory,altabaa2025cot,pmlr-v336-rajaraman26a,hanneke2026sample}. In the theoretical framework, a next-token generator is applied autoregressively to produce a CoT and a final answer. The resulting end-to-end concept can be computationally hard to learn from final answers alone. However, CoT supervision exposes the correct next token at each intermediate step, making the end-to-end concept efficiently learnable when next-token prediction is tractable \citep{malach2023auto,joshi2025theory}. Thus, CoT supervision breaks a hard-to-learn computation into easy-to-learn steps.\footnote{In fact, this principle dates back nearly four decades to the work of Rivest and Sloan \cite{rivest1988learning}.}

Yet this theory begins exactly where data curation ends: it assumes that \emph{homogeneous} CoT supervision from a \emph{single} generator (i.e., thinker) is already available, on which next-token prediction is easy to implement. In practice, CoT datasets draw on \emph{heterogeneous} sources, combining traces with different reasoning styles, and substantial effort goes into making the collected traces suitable for next-token training \citep[e.g.,][]{guha2025openthoughts,abdin2025phi}. For example, NuminaMath-CoT aggregates solutions from school exercises, mathematical olympiads, online forums, existing benchmarks, and synthetic data, then translates and realigns them into a uniform CoT format \citep{li2024numinamath}. Practitioners also filter traces for desired styles and formats \citep[e.g.,][]{grattafiori2024llama,guo2025deepseek}. We discuss these and other curation practices as the presentation progresses; see also \Cref{sec:related-work} for a broader discussion.

Once such supervision is curated, learning reduces, in both theory and practice, to next-token prediction. But these practices suggest that realizing the computational advantage of CoT supervision depends not merely on the correctness of individual traces, but also on whether and how heterogeneous supervision can be curated into a homogeneous form to facilitate next-token training \citep[e.g.,][]{grattafiori2024llama,guo2025deepseek,zhang2026best}. Therefore, in our view, a theory of learnability with CoT is incomplete without understanding how heterogeneity in reasoning traces affects efficient learning. We ask:
\begin{center}
    \emph{How do equally correct and valid but heterogeneous CoT traces affect efficient learnability?}
\end{center}

\paragraph{In this work.} Inspired by CoT data curation practices, we define models of supervision from multiple CoT generators, ranging from passive to active data collection, extending the autoregressive learning framework of \cite{joshi2025theory}. As in the line of work discussed above, we study classes that are computationally hard to learn without CoT but are efficiently learnable under homogeneous CoT supervision via a next-token prediction oracle. Our goal is to understand how data collection from multiple sources 
affects
the computational benefits of CoT supervision.

We show that learning can remain cryptographically hard under \emph{passively} collected heterogeneous CoT supervision from multiple sources, all representing the same end-to-end target concept. To model heterogeneity, we introduce two passive data-collection models that differ in how sources contribute across inputs and whether traces have known provenance of the source or are provided as an unfiltered mixture. 
Our first main result highlights how computationally indistinguishable contribution biases (i.e., which generators contribute to which parts of the domain) can obstruct efficient learning (\Cref{thm:hardness-with-two-gens}). 
Our next result establishes that a small amount of well-filtered homogeneous supervision can enable efficient learning, whereas even a polynomially large pool of unfiltered heterogeneous CoT traces can remain computationally hard to learn from (\Cref{thm:filtering-hardness}). 


Our final result (\Cref{thm:boosting}) shows the power of \emph{active} and \emph{adaptive} CoT collection from multiple sources, inspired by iterative and adaptive fine-tuning of language models. We present a generic computationally efficient learning algorithm that achieves arbitrarily small error $\eps$ by collecting CoTs from multiple sources, with a per-source budget completely independent of $\eps$.

\subsection{Background: a framework of learning concepts with autoregressive CoT}\label{subsec:CoT-Framework}
Inspired by learning in autoregressive transformers, \cite{joshi2025theory} introduced a framework for learning with autoregressive Chain-of-Thought (CoT), which we begin by recalling. A sequence-to-next-token generator (i.e., a thinker) is a function $f:\{0,1\}^* \to \{0,1\}$. For any such function, we define the apply-and-append map $\of(x): x \mapsto \textnormal{append}(x, f(x)) = [x, f(x)] \in \{0,1\}^{\abs{x}+1}$. Autoregressive generation produces a sequence of tokens by iteratively applying $\of$, namely $\of \circ \cdots \circ \of (x)$.

For any sequence-to-next-token generator $f$, and a generation (i.e., thinking) length $T$, we define the end-to-end mapping realized after $T$ steps as a function $\ete{f}{T}:\{0,1\}^* \to \{0,1\}$, given by
\begin{equation}\label{eq:fete}
    \ete{f}{T}(x) := \underbrace{\of \circ \of \circ \ldots \circ \of}_{\textrm{\small $T$ times}}(x)[-1]\,,
\end{equation}
where $\z[-1]$ denotes the last bit of any string $\z\in \{0,1\}^*$. We refer to all the intermediate tokens leading up to $h=\ete{f}{T}(x)$ as the CoT of $f$ on $x$, given by a mapping $\CoT{f}{T}:\{0,1\}^* \to \{0,1\}^T$,
\begin{equation}\label{eq:fCoT}
    \CoT{f}{T} (x) := [\ete{f}{1}(x),\dots,\ete{f}{T}(x)]\,.
\end{equation}
For any base class of next-token generators $\cF \subseteq \{0,1\}^{\{0,1\}^*}$, our goal is to learn the end-to-end class realized after $T$ steps, given by
\begin{equation}\label{eq:ete-class}
    \ete{\cF}{T}:=\{\ete{f}{T}:\{0,1\}^* \to \{0,1\} \mid f \in \cF\}\,.
\end{equation}
Although these functions are defined on all finite binary strings, each learning problem has an input domain $\cX\subseteq\{0,1\}^*$.

\paragraph{E2E vs.\ (Homogeneous) CoT supervision.}
The framework of \cite{joshi2025theory} studied the gap between learning from only the end result and learning with the intermediate CoT. In end-to-end (E2E) supervision, the learner observes examples $(x^{(i)},h_*(x^{(i)}))$, where the inputs are drawn i.i.d.\ from an unknown distribution $\cD\in \Delta(\cX)$ and $h_*\in\ete{\cF}{T}$ is the target concept. Homogeneous CoT supervision additionally provides $z^{(i)}=\CoT{f_*}{T}(x^{(i)})$ from a single generator $f_*\in\cF$ satisfying $\ete{f_*}{T}=h_*$. The precise PAC-style definition of $(\eps,\delta)$ learning is given in \Cref{sec:preliminary}.

For computational complexity, we use a parameter $d$ to specify the problem size. We consider input domains $\cX_d\subseteq \{0,1\}^{\leq d}$ and a family of next-token generator classes $\cF=\bigcup_{d\geq 1}\cF_d$, where $\cF_d\subseteq\{0,1\}^{\{0,1\}^*}$.\footnote{Although inputs lie in $\cX_d$, the domain of the base class $\cF_d$ is all finite binary strings $\{0,1\}^*$: generating a CoT of length $T$ requires evaluating prefixes of length up to $d+T-1$.} We write $\ete{\cF}{T}_d$ for the end-to-end class generated by $\cF_d$ after $T$ steps on $\cX_d$.

\paragraph{Tractable Consistent oracle.}
The CoT supervision reveals intermediate tokens that break a hard-to-learn end-to-end concept into steps that can be easy to learn. Therefore, efficiently learning $\ete{\cF}{T}_d$ reduces\footnote{Our focus is computational tractability. We assume that $\VCdim(\ete{\cF}{T}_d)$ is bounded by $\poly(d,T)$. This ensures that the class is learnable with $\poly(d,T,\eps^{-1},\log\delta^{-1})$ E2E samples, ignoring computation.} to finding a predictor in $\cF_d$ consistent with the observed prefix--next-token pairs, i.e., achieving zero empirical $0$--$1$ loss on next-token prediction \citep{joshi2025theory}. The assumption that this next-token consistency problem is computationally easy is captured by the following oracle.
\begin{definition}[Tractable $\Cons$ oracle]\label{def:cons-oracle}
We say that a base hypothesis class $(\cF_d)_d$ has a tractable $\Cons$ oracle if there is an algorithm that, given a training set
\( S=\{(\z^{(i)},f_*(\z^{(i)})) \mid  i \in [m]\}
\)
of $m$ examples of sequence and next-token pairs from a \textbf{single} generator $f_* \in \cF_d$, on sequences $\z^{(i)}$ of length at most $n$, finds some $\widehat{f}\in \cF_d$ such that for all $i \in [m]$, we have $\widehat{f}(\z^{(i)})=f_*(\z^{(i)})$, and runs in time $\poly(d,n,m)$.
\end{definition}
 
Relating this definition to practice, although transformers are not efficiently learnable in the worst case, next-token prediction on the input--next-token pairs obtained from sufficiently rich and homogeneous CoT supervision is believed to be easy in practice. The tractable $\Cons$ oracle (\Cref{def:cons-oracle}) captures this assumption. 
\begin{figure}[]
    \centering
\includegraphics[width=1\linewidth]{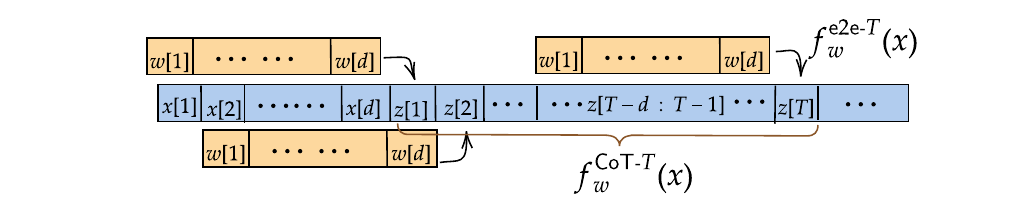}
\vspace{-20pt}
    \caption{A visual help on how a linear threshold is applied autoregressively.}
    \vspace{-10pt}
    \label{fig:intro-AR-linear-thresholds}
\end{figure}
\paragraph{The class of autoregressive linear thresholds.}\label{subsec:intro-AR-linear-threshold}\label{app:AR-linear} 
To contrast with efficient learnability results under homogeneous supervision, we will establish our hardness results for the class of \emph{autoregressive linear thresholds} \cite[e.g.,][]{joshi2025theory,hanneke2026sample}. For window size $d$, the base class is $\cFlin=\{f_w \mid w\in\R^d\}$, where
\[
f_w(\z)=\th\left(\sum_{i=1}^{\min\{d,|\z|\}}w[-i]\z[-i]\right),
\qquad
\th(a)=\ind\{a\ge 0\}.
\]
Negative indices count coordinates from the end. See also \Cref{fig:intro-AR-linear-thresholds}. This base class admits a provably efficient $\Cons$ oracle (see details in \Cref{sec:preliminary}), and thus under homogeneous CoTs from a single generator $f_*\in \cFlin$, the class after $T$ autoregressive steps $\ete{\cFlin}{T}$ is efficiently learnable.\footnote{By Sauer's lemma, $\VCdim(\ete{\cFlin}{T})\leq\log|\cFlin|=O(d^2)$; see \cite[Lemma 4.1]{joshi2025theory}. Thus, $O(\eps^{-1}(d^2+\log \delta^{-1}))$ E2E samples suffice statistically, yet learning without CoT supervision remains computationally hard under cryptographic assumptions \cite[Theorem 4.4]{joshi2025theory}.}

\begin{remark}[Alternative stochastic modeling choices]\label{rem:modeling-choices}
In this paper, we use classes of deterministic next-token generators, following prior learning-theoretic treatments of learning with CoTs. In practice, transformer-based language models predict distributions over tokens and are trained by next-token likelihood maximization, suggesting a stochastic formulation with policies $\pi:\{0,1\}^*\to\Delta(\{0,1\})$. Even under this formulation, the questions about heterogeneity remain: an efficient next-token likelihood-maximization oracle would still be assumed only for data generated by a single policy, while supervision could come from multiple policies under our different collection settings.  Our hardness results already hold in the deterministic formulation, where likelihood maximization on realizable data reduces to the consistency problem in \Cref{def:cons-oracle}. In \Cref{rem:ub-stochastic}, we further discuss how our main algorithmic result can be extended to a stochastic formulation. Thus, stochastic modeling offers a natural extension, but the questions studied here remain relevant.
\end{remark}

\section{Learning from multiple thinkers: definitions and results}\label{sec:main-results}\label{sec:results}\label{subsec:multiple-CoT-supervision-intro}

In this section, we introduce models of CoT supervision from multiple generators (i.e., thinkers) and the corresponding results, along with their motivation (\Cref{subsec:hard-to-learn-contribution-bias,subsec:hardness-without-filter,subsec:active-adaptive-collection}).

\begin{figure*}[t]
    \centering
    \includegraphics[width=\textwidth]{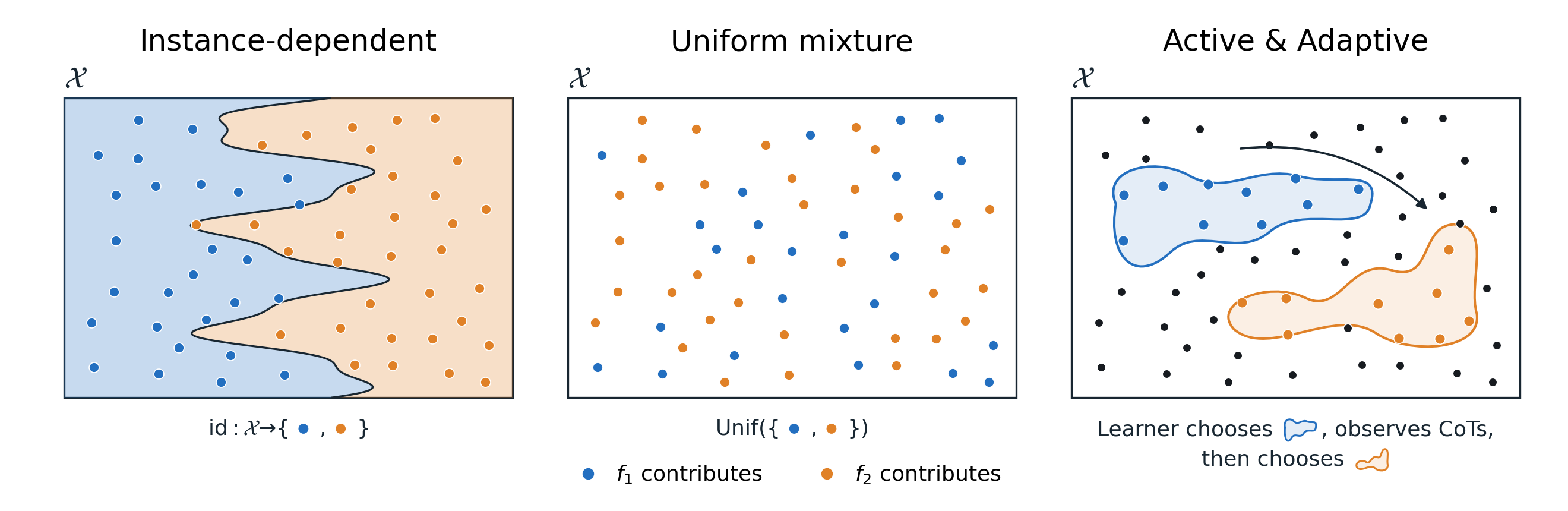}
    \vspace{-25pt}
    \caption{A visualization of how generators contribute CoTs across the domain under instance-dependent selection, a uniform mixture, and active and adaptive collection, discussed in \Cref{subsec:hard-to-learn-contribution-bias,subsec:hardness-without-filter,subsec:active-adaptive-collection}, respectively.}
    \vspace{-10pt}
    \label{fig:generator-settings}
\end{figure*}

A base class $\cF\subseteq\{0,1\}^{\{0,1\}^*}$, domain $\cX$, and generation length $T$ specify a learning problem for the end-to-end class $\ete{\cF}{T}$ on $\cX$. The adversary chooses an unknown realizable distribution over input--label pairs, with input marginal $\cD$ and labels $y=h_*(x)$ for some unknown $h_*\in\ete{\cF}{T}$. The learner receives CoTs from $\kappa$ generators $(f_1,\dots,f_\kappa)\in\cF^\kappa$ realizing the same target end-to-end:
\begin{equation}\label{eq:e2e-agreement}
   \ete{f_1}{T}(x)=\cdots=\ete{f_\kappa}{T}(x)=h_*(x)
   \qquad\text{almost surely for }x\sim\cD\,.
\end{equation}
We will soon introduce different models of CoT collection from these generators; see \Cref{fig:generator-settings}. Unless otherwise stated, the goal is to find a predictor $\widehat h:\cX\to\{0,1\}$ with small end-to-end error:
\begin{equation}\label{eq:loss}
   \cL_{\cD,h_*}(\widehat h)
   =\Pr_{x\sim\cD}\left(\widehat h(x)\neq h_*(x)\right)\,.
\end{equation}
The main distinction from prior line of work is that the CoT supervision is no longer guaranteed to come from a single generator $f_*\in\cF$.

Importantly, we still assume an efficient $\Cons$ oracle \emph{only for prefix-next-token examples realizable by a single generator} $f_*\in\cF$, as in \Cref{def:cons-oracle}. This restriction reflects our practical motivation: next-token prediction remains the standard training objective, while substantial effort goes into curating homogeneous reasoning traces to make this training effective. The oracle captures the tractability of fitting homogeneous traces. We ask when and how this computational primitive alone suffices for efficient learning from heterogeneous CoT supervision.

\subsection{Hardness of learning under instance-dependent contribution biases}\label{subsec:hard-to-learn-contribution-bias}\label{subsec:passive-known-sources}
Our first data collection model is inspired by the common practice of assembling larger datasets from specialized ones. For example, MathInstruct contains different reasoning styles, including step-by-step arithmetic explanations, competition-style mathematical solutions, and executable programs \citep{yue2024mammoth}. This variety arises because MathInstruct combines 13 rationale subsets (including GSM8K, AQuA-RAT, MATH, and TheoremQA) for different types of mathematical questions. Thus, seeking broad coverage by pooling specialized datasets can make the reasoning style depend on which source contributed the question; see \Cref{sec:related-work} for other examples.

To model how different sources contribute on different parts of the domain, consider an unknown partition $\cX=\bigsqcup_{k=1}^{\kappa}\cX_k$, where generator $f_k$ provides the CoTs for inputs in $\cX_k$. Equivalently, a partition function $\id:\cX\to[\kappa]$, with $\id(x)=k$ for $x\in\cX_k$, specifies the contributing generator.

\begin{definition}\label{def:instance-dependent-bias}
    The adversary chooses $\cD \in \Delta(\cX),h_*\in \ete{\cF}{T}$, CoT generators $(f_1,\dots,f_\kappa)\in \cF^\kappa$ satisfying \eqref{eq:e2e-agreement}, and the partition  $\id: \cX \to [\kappa]$. The learner receives the source-labeled CoT samples
\[
S_{\sCoT}
=
\left\{
\left(x^{(i)},j^{(i)},z^{(i)}\right):
j^{(i)}=\id(x^{(i)}),\;
z^{(i)}=\CoT{f_{j^{(i)}}}{T}(x^{(i)}), \, \forall i \in [m]
\right\},
\]
where $x^{(1)},\dots,x^{(m)}\simiid\cD$. We call this the \emph{\textbf{instance-dependent contribution bias}} model with known provenance.
\end{definition}
We emphasize that the provenance of every training trace is known to the learner. Receiving an unfiltered mixture $(x^{(i)},z^{(i)})$ without the source identities $j^{(i)}$ can only make learning harder, and we will return to this distinction in the next subsection. When provenance is known, as in the MathInstruct example, the learner can identify constituent subsets such as MATH, TheoremQA, and GSM8K and train on each homogeneous subset separately. But what can we say about \emph{generalization} across the \emph{entire} input domain?

In our framework, the learner can find predictors $(\widehat{f}_1,\dots,\widehat{f}_\kappa)$ consistent with the respective subsets $S_k=\{(x^{(i)},z^{(i)}): j^{(i)}=k\}$. Each $\widehat{f}_k$ generalizes well under $\cD\mid x\in\cX_k$, but has no such guarantee elsewhere. On a test input $x\sim\cD$, the predictor $\widehat{f}_k$ corresponding to $x\in\cX_k$ is guaranteed to generalize well, but it is unclear how to determine this correct routing from training samples.

Our first result shows that such instance-dependent contribution biases can 
make learning the shared end-to-end target computationally hard (even weakly, i.e., to constants $\eps<0.5$ and $\delta<1$), already for $\kappa=2$ generators. Our hardness results rely either on the subexponential-time quantum hardness of worst-case lattice problems or the existence of Goldreich's local pseudorandom generators with polynomial stretch \cite{goldreich2011candidate}. Both assumptions, formally stated in \Cref{subsec:crypto}, have also been used in the hardness-of-learning literature; see \Cref{sec:related-work}.

\begin{tcolorbox}[colback=white]
\begin{theorem}\label{thm:contribution-bias-hardness}\label{thm:hardness-with-two-gens}
Assuming the quantum hardness of $\gapsvp$ or $\sivp$ (\Cref{assump:lattice}), for any non-decreasing $g(d)=\omega_d(\log d \log \log d)$ computable in $\poly(d)$ time, there is no $\poly(d)$-time algorithm that weakly learns $\ete{\cFlin}{T(d)}$ for all $T(d)\leq g(d)$, from the CoT supervision of $\kappa=2$ generators under the \textbf{instance-dependent contribution bias} model with known provenance (\Cref{def:instance-dependent-bias}).\\
Assuming the existence of local PRGs with polynomial stretch (\Cref{assumption:prg_exists}), the same conclusion holds for any non-decreasing $g(d)=\omega_d(1)$ computable in $\poly(d)$ time.
\end{theorem}
\end{tcolorbox}
Recall that the class $\ete{\cFlin}{T}$ is learnable in $\poly(d,T,\eps^{-1},\log\delta^{-1})$ time from homogeneous CoT supervision by a single $f_*\in\cFlin$ \cite[Corollary 4.3]{joshi2025theory}. Yet instance-dependent contribution biases from just two sources can make learning hard. On the other hand, if the source-selection function $\id:\cX\to[\kappa]$ is efficiently learnable, an efficient $\Cons$ oracle suffices for efficient learning. Thus, hardness of learning the contribution biases is necessary for the hardness in \Cref{thm:contribution-bias-hardness}. In our MathInstruct example \citep{yue2024mammoth}, the questions alone do not easily reveal the contributing source, since distinct questions of a \emph{similar mathematical nature} receive \emph{different} styles of supervision (e.g., natural-language derivations in AQuA-RAT versus executable program solutions in MathQA).

While hardness of learning the contribution biases is necessary, it is not sufficient to establish hardness of learning the target. The learner may still exploit the CoTs to predict the target without identifying the contributing source; see \Cref{subsec:prf-ove-hardness-of-two,subsec:parity-main-paper} for supporting evidence. Our reduction strategy must  \emph{control} the information leaked through the CoTs, which we explain in \Cref{subsec:prf-ove-hardness-of-two}.

\paragraph{Achieving good $\passkappa$-accuracy.} 
We highlight another important guarantee: even when the contribution biases are hard to learn, there is still an efficient learner under the $\passkappa$ metric, where the learner outputs $\kappa$ candidate responses and is evaluated on the best one.\footnote{It is important to realize that, to discuss the $\passkappa$ metric, we must shift our focus to CoT agreement (with one of the generators), since evaluating for only the final answer is trivial when the answer space is small (e.g, binary). We also note that evaluating the best of multiple responses is common in LLM benchmarking \cite[e.g.,][]{chen2021evaluating}.}
Indeed, the learner can partition the training sample using the observed source labels and fit predictors $\widehat f_1,\ldots,\widehat f_\kappa$. On a test input $x$, it outputs
\[
\widehat{\mathcal Z}(x)
:=
\left\{
\CoT{\widehat f_1}{T}(x),\ldots,
\CoT{\widehat f_\kappa}{T}(x)
\right\}\,.
\]
Roughly speaking, each $\widehat f_k$ generalizes under $\cD_k :=\cD\mid x\in\cX_k$, provided its training subset is sufficiently large. Overall, with probability at least $1-\delta$ over a training sample of size $\poly(\kappa,\VCdim(\cF_d),T,\eps^{-1},\log\delta^{-1})$, we have the guarantee that
\[
\Pr_{x\sim\cD}
\left[
\CoT{f_{\id(x)}}{T}(x)\in\widehat{\mathcal Z}(x)
\right]
\geq 1-\eps.
\]
The formal guarantee is provided in \Cref{thm:passk}. This is also referred to as \emph{coverage}: the learner can produce a small set containing a correct trace without knowing which candidate is correct. Such coverage provides a foundation for subsequent reinforcement learning or test-time scaling methods (e.g., Best-of-$N$) \cite{chen2026coverage}. To our knowledge, this is the first (non-trivial) computational separation between coverage and accuracy: under instance-dependent contribution biases, the learner can efficiently produce a set containing a valid CoT, while learning to directly output even the final answer remains computationally intractable.


\begin{remark}[E2E hardness with small $T$]\label{rem:e2e-hardness-for-small-T}
    Our supervision model (\Cref{def:instance-dependent-bias}) is strictly stronger than E2E supervision, where the learner observes only input--label pairs $(x,h_*(x))$. 
    Therefore, \Cref{thm:contribution-bias-hardness} also establishes E2E learning hardness for 
    $T(d)\leq g(d)$, for any $g(d)=\omega_d(\log d\log\log d)$ under \Cref{assump:lattice}, or $g(d)=\omega_d(1)$ under \Cref{assumption:prg_exists}. 
    This improves upon the hardness at $T=\Theta(d)$ established by \cite{joshi2025theory}. 
    The improvement comes from a more efficient  simulation of depth-two threshold circuits by autoregressive linear thresholds; see \Cref{sec:technical-overview}. This construction yields a hardness result for E2E learning at short generation lengths $T$, assuming hardness of learning small depth-$2$ threshold circuits (\Cref{thm:ete-hardness}).
\end{remark}

\subsection{Hardness of learning from unfiltered uniform mixture}
\label{subsec:hardness-without-filter}
Our next supervision model is inspired by another common data-curation practice that filters heterogeneous CoT traces to obtain homogeneous supervision. Practitioners rely on heuristic filtering to obtain sufficiently consistent supervision from such mixtures \citep[e.g.,][]{guo2025deepseek,zhang2026best}. These efforts often go beyond checking answer correctness. For example, Llama 3 \citep{grattafiori2024llama} uses a model-based judge to score coding responses for correctness as well as code style. Can such filtering be necessary for CoT supervision to allow efficient learning, even when every available trace is valid and reaches the correct answer?

To study this question, we consider an \emph{unfiltered} heterogeneous mixture of CoTs with \emph{unknown} provenance. Unlike in \Cref{subsec:hard-to-learn-contribution-bias}, the generator for each input is chosen uniformly from $[\kappa]$, independently of the input, so \emph{each} generator contributes examples from the input distribution $\cD$.
\begin{definition}\label{def:uniform-mixture}
The adversary chooses $\cD\in\Delta(\cX)$, $h_*\in\ete{\cF}{T}$, and CoT generators $(f_1,\dots,f_\kappa)\in\cF^\kappa$ satisfying \eqref{eq:e2e-agreement}. The learner receives the CoT samples
\[
S_{\sCoT}
=
\left\{
\left(x^{(i)},z^{(i)}\right):
z^{(i)}=\CoT{f_{j^{(i)}}}{T}(x^{(i)}),\,
\forall i\in[m]
\right\},
\]
where $x^{(1)},\dots,x^{(m)}\simiid\cD$ and $j^{(1)},\dots,j^{(m)}\simiid\Unif[\kappa]$ are sampled independently.  We call this setting learning from an \emph{\textbf{unfiltered uniform mixture}}.
\end{definition}
Unlike in \Cref{def:instance-dependent-bias}, there are no contribution biases, but the challenge is that the learner does not know which examples form a homogeneous subset, contributed by a single generator.

\paragraph{A filtering oracle reduces learning to a single generator.}
Suppose an oracle allows the learner to filter the traces contributed by any single generator:
\begin{equation}\label{eq:filtered-subset}
    S_k
=
\left\{(x^{(i)},z^{(i)}) \in S_{\sCoT}:j^{(i)}=k\right\}.
\end{equation}
In the absence of contribution biases, $j^{(i)}$ is independent of $x^{(i)}$, and thus the inputs in $S_k$ remain i.i.d.\ from $\cD$. Therefore, a linear overhead in $\kappa$ in the total sample size suffices to obtain a filtered subset $S_k$ of size $O(\eps^{-1}(\VCdim(\cF_d)\log T\log(2/\eps)+\log\delta^{-1}))$. This subset provides homogeneous CoT supervision from $f_k$ on the target input distribution $\cD$. Calling the efficient $\Cons$ oracle on $S_k$ then yields an efficient learner for $\ete{f_k}{T}=h_*$. However, without access to such a filtering oracle, learning from such a mixture can be computationally hard. 

Our next result shows that, under the same cryptographic assumptions, no polynomial time learner can learn from an \emph{unfiltered} uniform mixture of \emph{a few} sources.
\begin{tcolorbox}[colback=white]
\begin{theorem}\label{thm:filtering-hardness}
Assuming the quantum hardness of $\gapsvp$ or $\sivp$ (\Cref{assump:lattice}), for every non-decreasing sequence $g(d)=\omega_d(\log d \log \log d)$ that is computable in $\poly(d)$ time, there is no $\poly(d)$-runtime algorithm that weakly learns $\ete{\cFlin}{T(d)}$ from an \textbf{unfiltered uniform mixture} of $\kappa(d)$ generators (\Cref{def:uniform-mixture}), for all $T(d),\kappa(d)\leq g(d)$.\\
Assuming the existence of local PRGs with polynomial stretch (\Cref{assumption:prg_exists}), the same conclusion holds for any non-decreasing $g(d)=\omega_d(1)$ computable in $\poly(d)$ time.
\end{theorem}
\end{tcolorbox}
The theorem gives a sharp contrast: a small homogeneous subset suffices, while any polynomially many samples from the unfiltered mixture do not suffice for efficient learnability. This echoes the emphasis on data quality over quantity in practical training pipelines \citep[e.g.,][]{touvron2023llama,wang2023data,abdin2025phi,ye2025limo}. While quality can refer to many properties in practice, our framework focuses on homogeneity: every trace is already valid and reaches the correct answer, yet without filtering a dataset, it can remain hard to learn efficiently.

Finally, \Cref{thm:filtering-hardness} also implies that a filtering oracle cannot be implemented efficiently for the class $\cFlin$. Although this guarantee may be pessimistic, it highlights the computational role of filtering in making CoT supervision useful for efficient learning. Moreover, even when heuristic filtering approaches succeed in practice, there may be contribution biases (\Cref{subsec:hard-to-learn-contribution-bias}), which may still make learning hard (\Cref{thm:contribution-bias-hardness}), providing another obstacle despite successful filtering.

\subsection{An algorithm under active \& adaptive collection from multiple sources}
\label{subsec:efficient-algorithm-via-boosting}\label{subsec:active-adaptive-collection}\label{subsec:boostin-based-learner}

Our next supervision model is inspired by iterative and adaptive fine-tuning of language models, where they benefit from CoT supervision from multiple sources \citep[e.g.,][]{grattafiori2024llama,li2025learning}, through successive rounds of data collection and training. Across these rounds, the student's current weaknesses can guide which examples receive further supervision \citep[e.g.,][]{lee2024llm2llm,kessler2025towards}. The teacher providing this supervision may also change, either evolving through training on earlier traces \citep{zelikman2022star,singh2023beyond}, or being selected for the particular capabilities  depending on the questions, as in Llama 3 \citep{grattafiori2024llama}. 

Motivated by these practices, we ask whether a learner can benefit from small CoT datasets that are individually homogeneous but collectively heterogeneous. We study active and adaptive CoT collection, where the learner selects a small number of examples on which to request homogeneous CoTs in each round, while the generating source may depend on the selected examples and previous rounds. We formalize this protocol below.

\begin{definition}[Active \& Adaptive collection]\label{def:active-adaptive-collection}
    The learner has a budget of $\M$ CoTs per round.
\begin{itemize}
\vspace{-6pt}
    \item The adversary chooses $\cD \in \Delta(\cX)$ and $h_*\in \ete{\cF}{T}$. The learner receives a passive E2E training set $S_{\sete}=\{(x^{(i)},y^{(i)}) \mid y^{(i)}=h_*(x^{(i)}) \,\forall i \in [m]\}$ where $x^{(1)},\dots,x^{(m)}\simiid \cD$.
    \vspace{-5pt}
    \item Active \& Adaptive CoT collection proceeds in rounds $k=1,\dots,\kappa$. In every round $k\in [\kappa]$:
    \vspace{-5pt}
    \begin{enumerate}
    \item The learner selects a subset $\cS_k \subseteq [m]$ of size at most $\M$, on which it requests CoTs from a single source.
    \item The adversary then chooses a source $f_k\in\cF$ such that $\ete{f_k}{T}\equiv h_*$, and reveals the CoTs $\{\CoT{f_k}{T}(x^{(i)}) \mid i \in \cS_k\}$ to the learner. The choice can depend on the entire history $\cS_1,\dots,\cS_{k-1},f_1,\dots,f_{k-1}$, as well as the current subset $\cS_k$.
    \end{enumerate}
\end{itemize}
\end{definition}
The learner controls which examples receive CoTs, but not which generator provides them. In each round, it submits all queries together and receives a homogeneous batch on which it can use the $\Cons$ oracle. Across rounds, the supervision may be arbitrarily heterogeneous and adaptive. We show that under the protocol in \Cref{def:active-adaptive-collection}, the learner can achieve arbitrarily small error $\eps$, while the number of CoTs collected from any single $f_k$ is independent of $\eps$. 
\begin{tcolorbox}[colback=white]
\begin{theorem}[Simplification of \Cref{thm:boosting-model-aggregation}]\label{thm:boosting} There is an algorithm (\Cref{alg:adaboost-cot}) under the protocol in \Cref{def:active-adaptive-collection} such that for every base class hierarchy $(\cF_d)_d$ and generation length $T$, with probability at least $0.99$ over the algorithm's internal randomness and the sampling of an E2E training set of size $m = \Tilde{O}( \eps^{-1}\VCdim(\ete{\cF}{T}_d))$, the algorithm returns a predictor $\widehat{h}:\cX_d \to \{0,1\}$ such that:
$$\cL_{\cD,h_*}(\widehat{h})=\Pr_{x\sim \cD}\left(\widehat{h}(x) \neq h_*(x) \right) \leq \eps\,.$$
The algorithm actively and adaptively collects the CoTs from different generators under the protocol in \Cref{def:active-adaptive-collection}, and has the following guarantee:
\begin{itemize}
    \item It collects CoTs over $\kappa=\Tilde{O}(\log m)=\Tilde{O}\left(\log \eps^{-1}+\log \VCdim(\ete{\cF}{T}_d)\right)$ rounds;
    \item Demands at most $\M=O(\VCdim(\ete{\cF}{T}_d))$ CoTs in a single round;
    \item And runs in time $\poly(d,T,\eps^{-1})$ as long as $(\cF_d)_d$ has a tractable $\Cons$ oracle (\Cref{def:cons-oracle}) and $\VCdim(\ete{\cF}{T}_d)=\poly(d,T)$.
\end{itemize}
\end{theorem}
\end{tcolorbox}
The algorithm uses \emph{boosting} to combine the predictors learned in different rounds as weak learners into an overall strong learner \cite{schapire1990strength,freund1997decision}. The theorem shows that small homogeneous CoT batches from \emph{different} sources can be combined across rounds to achieve arbitrarily small end-to-end error, even when the sources change adaptively. This captures a feature of iterative fine-tuning, where successive rounds use traces from evolving models or different teacher checkpoints \citep[e.g.,][]{zelikman2022star,singh2023beyond,grattafiori2024llama}. Our result provides a theoretical basis for continued improvement from such varying sources with a CoT budget per round that is independent of the target accuracy. An interesting open question is whether a similar benefit (a per-source CoT budget independent of $\eps$) is possible without adaptivity or perhaps even without active collection, for example, under a uniform mixture with known provenance (\Cref{def:uniform-mixture}), which allows traces to be filtered by source.

\section{Discussion}\label{sec:discussion-future-work}\label{subsec:future-work}

Our work brings data collection and curation into the theoretical study of learning with CoT supervision by studying heterogeneity in the collected traces. While practical data curation addresses many aspects of supervision quality, we focus on one: homogeneity of the traces. To study its role in efficient learning, we introduced several new models of CoT supervision from multiple sources.

Although such heterogeneous supervision consists entirely of correct traces, its computational benefit depends on how they are collected. Our main results establish the computational gap between \emph{coverage} and \emph{accuracy} under instance-dependent contribution biases, show why filtering can be necessary for efficient learning, and demonstrate how active and adaptive collection, as in iterative fine-tuning on multiple sources, can be very powerful. Our results isolate different aspects of learning from heterogeneous supervision that may even coexist in practical scenarios.

From a theoretical perspective, the landscape ranges from cryptographic hardness under passive data collection to strong algorithmic guarantees under active and adaptive collection. Distribution-free hardness results, including ours, are often criticized for not reflecting the practical difficulty of machine learning problems. Nevertheless, they can guide the search for assumptions and formulations that admit provably efficient learning. In our setting, they highlight the properties of supervision that matter for computational tractability and motivate alternative data-collection models. Overall, shifting the focus to heterogeneous supervision, in the theoretical study of learning with CoTs, allows us to illuminate several prevalent practices in data curation.

\paragraph{Future Directions.} We begin with some immediate questions arising from our results, and then discuss a few broader directions.
\begin{itemize}

\item In practice, models can benefit from supervision collected from multiple sources without adaptivity \citep[e.g.,][]{mukherjee2023orca,zhang2026best}. Can active but non-adaptive CoT collection match the guarantees of \Cref{alg:adaboost-cot}, with a per-source budget $\M$ independent of $\eps$? More generally, is there a fundamental tradeoff between adaptivity and the total number of CoT queries required for efficient learning?

\item Does this benefit extend to passive collection from an already-filtered uniform mixture without contribution biases? Here, the traces are grouped into homogeneous subsets with known provenance. Can an efficient learner achieve error $\eps$ using $o(1/\eps)$ CoTs per source, up to polynomial factors in $d$ and $T$? Can the per-source budget be even independent of $\eps$?

   \item Two technical questions concern strengthening our hardness results. (i) Our hardness result for learning from an unfiltered mixture in \Cref{thm:filtering-hardness} requires growing $\kappa$; we conjecture that mixture of $\kappa=2$ sources suffices to establish the hardness. (ii) Can we also establish hardness of end-to-end learning with short generation length $T=O(1)$? This would improve the dependence on $T$ from $T=\omega_d(1)$ in \Cref{rem:e2e-hardness-for-small-T}; see \Cref{sec:technical-overview} for further discussion.
\end{itemize}
Adjacent to our study is the question of sample complexity for learning from multiple sources. The main goal is to identify the combinatorial structures that characterize the optimal sample complexity under the different data-collection settings considered here. Within each setting, one can study end-to-end loss, CoT-agreement loss (requiring agreement with any one of the valid CoTs), and $\passk$ loss, which is closely related to list learnability \citep{brukhim2022characterization,charikar2023characterization}. We believe that progress toward characterizing these problems could reveal interesting combinatorial notions beyond those identified for single-source CoT supervision in \cite{hanneke2026sample,li2026optimal}.

Beyond the settings studied here, it would be interesting to understand how other forms of imperfect CoT supervision affect the computational benefits of learning with CoTs. For example, missing steps can be modeled by passing CoTs through an erasure channel that reveals only part of each trace. When does learning remain hard, and under what assumptions is it easy? An interesting setting is when only prefixes of long CoTs are provided when actively queried by a learner. Such approaches have shown empirical promise~\cite[e.g.,][]{Ama+25,huang2025blending}. A natural theoretical goal is to identify concept classes that can be learned efficiently under such partial CoT supervision.

\paragraph{Acknowledgment.} NJ thanks Nived Rajaraman for a discussion that led us to explore the active and adaptive learning setting (\Cref{subsec:active-adaptive-collection}). Part of this work was done as part of the Simons/NSF Sponsored Collaboration on the Theoretical Foundations of Deep Learning and the NSF TRIPOD
Institute for Data, Econometrics, Algorithms, and Learning (IDEAL). NJ is supported in part by the Apple AI/ML PhD fellowship. GV is supported by the Israel Science Foundation (grant No. 2574/25), by a research grant from Mortimer Zuckerman (the Zuckerman STEM Leadership Program), and by research grants from the Center for New Scientists at the Weizmann Institute of Science, and the Shimon and Golde Picker -- Weizmann Annual Grant.

\paragraph{Outline.} In \Cref{sec:technical-overview}, we provide a technical overview of the ideas behind the main results. We discuss related work in \Cref{sec:related-work} and present formal definitions, hardness assumptions, and technical preliminaries in \Cref{sec:preliminary}. In \Cref{sec:additional-results}, we formally state the additional results mentioned over the course of the presentation. Full proofs of all theorems are deferred to the appendices.

\section{Technical overview}\label{sec:technical-overview}

In this section, we provide a high-level overview of the ideas behind our main results. Recall the autoregressive learning framework from \Cref{subsec:CoT-Framework}.

It is well-known that cryptographic primitives (e.g., pseudorandom functions or secure public-key encryption) directly imply the existence of computationally hard learning problems. However, learning theory is concerned with whether natural learning problems are easy or hard. Here, we are concerned with the end-to-end functions computed by autoregressive linear thresholds, namely $\ete{\cFlin}{T}$. A common approach to establishing cryptographic hardness for standard \emph{supervised learning} problems is to show that the concept class can express a hard-to-learn cryptographic function class \citep[e.g.,][]{kearns1994cryptographic,klivans2009cryptographic}. An efficient learner for the former would then learn the latter, contradicting the cryptographic assumption. Since learning $\ete{\cFlin}{T}$ from $(x^{(i)},h_*(x^{(i)}))$, without any CoT, is a standard supervised learning problem, this expressivity argument suffices to establish E2E hardness.

However, establishing hardness with (heterogeneous) CoT supervision is more delicate. CoTs can reveal enough information about the target to make learning easy, as the results for homogeneous supervision demonstrate \cite{joshi2025theory}. We must control what each generator reveals through its autoregressive trace while computing the target function. To this end, our starting point is a \emph{CoT-leakage-preserving} expressivity result for $\ete{\cFlin}{T}$, which we discuss next. 

\subsection{CoT leakage preserving simulation of depth two circuits}\label{subsec:expressivity}
 We show that autoregressive linear thresholds can simulate depth-$2$ threshold circuits, which include natural classes known to be hard to learn under cryptographic assumptions \citep[e.g.,][]{klivans2009cryptographic,daniely2021local}; see \Cref{sec:related-work} for further discussion. Crucially, the information leakage from the autoregressive CoT generated by $f_w$ is equivalent to the leakage from the gate outputs of the circuit being expressed.
\begin{lemma}[Formalized in \Cref{lemma:LT-expressivity-depth-2-d=n'}]\label{lem:informal-expressivity-wo-CoT}
There is an efficiently computable feature map $\phi:\{0,1\}^n\to\{0,1\}^d$, with $d=\Theta(ns)$, such that every depth-$2$ threshold circuit $C$ of size $s$ on $n$ inputs can be simulated by some $f_w\in\cFlin$ in $T=2s-1$ steps as follows: for every $x\in\{0,1\}^n$,
\begin{equation}\label{eq:CoT-preservation}
    \CoT{f_w}{T}(\phi(x))
=
[y_1,\dots,y_{s-1},0^{s-1},y_s],
\end{equation}
where $y_1,\dots,y_{s-1}$ are the hidden-gate outputs and $y_s=C(x)$.
\end{lemma}

The construction is technical in nature.\footnote{Roughly speaking, we arrange the circuit's weights and biases in a single vector $w$ to simulate each gate at its designated autoregressive step. The main difficulty is that since the same threshold $f_w$ is reused and its alignment shifts by only one position per step, input padding and inserted zero outputs are needed to align the second-layer weights without interference from the first-layer weights.} Full details appear in \Cref{sec:additional-results}. The CoT-leakage-preserving property\footnote{Another interesting property of the simulation is that it uses dimension $d=\Theta(ns)$ and thinking length $T=2s-1$, compared with $d,T=\Theta(ns^2)$ in \cite{joshi2025theory}. Their construction has $T\propto d$, whereas ours allows us to establish hardness for much smaller values of $T$, as discussed in \Cref{rem:e2e-hardness-for-small-T}. This expressivity alone yields E2E learning hardness from hardness of learning depth-$2$ circuits; see \Cref{thm:ete-hardness} for the formal statement.} in \eqref{eq:CoT-preservation} ensures that, beyond the final answer, the trace reveals exactly the intermediate gate outputs $[y_1,\dots,y_{s-1}]$.

\subsection{Reduction for \texorpdfstring{\Cref{thm:contribution-bias-hardness}}{}: hiding leakage via a distribution fork}\label{subsec:prf-ove-hardness-of-two}
Recall that, in the instance-dependent contribution bias model (\Cref{def:instance-dependent-bias}), hardness requires the contribution biases themselves to be hard to learn (\Cref{subsec:hard-to-learn-contribution-bias}). However, this is not sufficient to establish \Cref{thm:contribution-bias-hardness}, since the CoTs may reveal enough information to learn the target without identifying the source.\footnote{We explicitly demonstrate this in \Cref{sec:parity} for the problem of \emph{learning noisy parities}, which has received attention in learning with CoT because of its end-to-end hardness \citep[e.g.,][]{WLS23,malach2023auto,kim2024transformers,tsilivis2025reinforcement}.}

\begin{figure}
    \centering
    \includegraphics[width=1.03\linewidth]{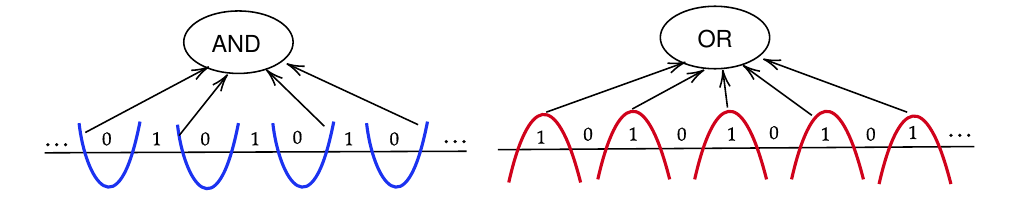}
    \caption{Two representations of the same decryption function of Regev's cryptosystem. Left: construction of \cite{klivans2009cryptographic} using intersections of degree-$2$ PTFs. Right: our construction using unions of degree-$2$ PTFs.}
    \label{fig:KS-vs-us}
\end{figure}

Our main idea is to construct a \textbf{\emph{distribution fork}}: two generators $f_0,f_1\in\cFlin$ realizing the same target $h_*$ end-to-end, where $f_0$ has trivial CoTs on inputs where $h_*(x)=0$ and $f_1$ has trivial CoTs on inputs where $h_*(x)=1$. Here, trivial means reconstructible from the hard-to-learn E2E labels alone. The useful CoT information from each generator is confined to the part of the distribution on which the other generator contributes. By the CoT-leakage-preserving simulation in \Cref{lem:informal-expressivity-wo-CoT}, it suffices to construct two circuits with this property.

One side already holds in existing supervised-learning hardness constructions: for intersections of halfspaces, all hidden gates output $1$ when $h_*(x)=1$. The challenge is to represent the \emph{same} function with trivial hidden-gate outputs on inputs labeled $0$. Under \Cref{assump:lattice}, we achieve this for decryption functions of  Regev's cryptosystem \citep{regev2009lattices}. Klivans and Sherstov \citep{klivans2009cryptographic} represent these functions as intersections of degree-$2$ polynomial threshold functions (PTFs) that exclude the intervals labeled $0$. A quadratic feature map then turns these PTFs into halfspaces:
\[
\text{Regev's decryption function}
\;\longrightarrow\;
\text{Intersection of degree-$2$ PTFs}
\;\longrightarrow\;
\text{Intersection of halfspaces}.
\]
The interval structure of a decryption function (cf. \Cref{fig:KS-vs-us}) also allows us to recognize the intervals labeled $1$ using a union of degree-$2$ PTFs, and hence a union of halfspaces under the same feature map. This gives the complementary side of the fork. We emphasize that this is not an application of De Morgan's law, since we have to express the same function, not its negation.
\begin{lemma}[Informal; formalized in \Cref{lem:two-ways=for-regev}]
Each decryption function of Regev's cryptosystem has two representations as depth-$2$ circuits with degree-$2$ PTFs in the hidden layer, both computing the same function. On inputs labeled $1$, all hidden gates in the intersection representation output $1$. On inputs labeled $0$, all hidden gates in the union representation output $0$.
\end{lemma}
Setting $\id(x)=h_*(x)$ selects the generator with trivial CoTs on each input, so both the trace and its provenance can be reconstructed from the E2E label. An efficient learner would therefore break the cryptosystem. Under the local-PRG assumption (\Cref{assumption:prg_exists}), we obtain the same property using DNF and CNF representations of a hard-to-learn target; see \Cref{lem:hardness_of_dnf_adversarial}.

\subsection{Reduction for \texorpdfstring{\Cref{thm:filtering-hardness}}{}: hiding leakage via cyclic gate shifts}\label{subsec:flitering-pf-ovw}
In an unfiltered uniform mixture (\Cref{def:uniform-mixture}), the source is chosen independently of the input. Thus, the previous strategy of hiding CoT leakage through a distribution fork does not work for \Cref{thm:filtering-hardness}. Instead, we hide the leakage by uniformly mixing a fixed family of equivalent circuits whose mixed gate outputs can be sampled from the E2E label alone.

The key is another structural property of Regev's decryption function: consider the union-of-PTFs representation in \Cref{fig:KS-vs-us} (right). Each hidden gate tests membership in one interval labeled \(1\), and these intervals are disjoint. This gives the following property.

\begin{lemma}[Formalized in \Cref{lem:two-ways=for-regev}]\label{lem:informal-regev-single-zero}
Each Regev's decryption function has a depth-$2$ circuit representation with $s-1$ degree-$2$ PTFs in the hidden layer and an OR output gate. On inputs labeled $0$, all hidden gates output $0$. On inputs labeled $1$, \textbf{exactly one hidden gate} outputs $1$.
\end{lemma}

Fix such a circuit with hidden-gate outputs $y_1,\dots,y_{s-1}$. For each $r\in\{0,\dots,s-2\}$, construct an equivalent circuit whose hidden gates are cyclically shifted to produce
\[
[y_{r+1},\dots,y_{s-1},y_1,\dots,y_r].
\]
This gives $\kappa=s-1$ fixed circuits computing the same function. On an example with $h_*(x)=1$, a uniformly chosen circuit places the unique $1$ at a uniformly random position. On an example with $h_*(x)=0$, every circuit produces the all-zero vector. Thus, the mixed distribution of hidden-gate outputs depends only on the final label.

By the CoT-leakage-preserving simulation in \Cref{lem:informal-expressivity-wo-CoT}, the corresponding CoTs can therefore be generated from E2E labels with exactly the distribution of an unfiltered uniform mixture of $\kappa$ generators in $\cFlin$. An efficient learner would then break the cryptosystem.\footnote{The reduction also establishes hardness when the learner receives CoTs from all $\kappa$ sources on every example as an unordered multiset without provenance.} We exploit a similar structural property under the local-PRG assumption (\Cref{assumption:prg_exists}) to obtain improved dependence on $\kappa$ and $T$.

\subsection{Idea behind the algorithmic result (\texorpdfstring{\Cref{thm:boosting}}{}): aggregation by boosting}
\label{subsec:technical-boosting}

Recall the active and adaptive collection protocol in \Cref{def:active-adaptive-collection}. The learner receives a training set of E2E examples and selects its subsets on which to request homogeneous CoTs in each round. The sources across rounds may be arbitrarily heterogeneous and adaptive.

\paragraph{CoTs make consistency computationally tractable.} By the fundamental theorem of statistical learning (\Cref{prop:fundamental-VC}), any predictor consistent with
\[
m=O\left(\frac{\VCdim(\ete{\cF}{T}_d)\log(2/\eps)+\log\delta^{-1}}{\eps}\right)
\]
i.i.d.\ E2E examples has error at most $\eps$ with probability at least $1-\delta$. Finding such a predictor can be computationally hard, even for $\cFlin$ \citep{joshi2025theory}. Homogeneous CoTs make this efficient: applying $\Cons$ (\Cref{def:cons-oracle}) to their prefix--next-token pairs yields a generator consistent with the traces and hence the E2E labels. The statistical guarantee holds for all consistent E2E predictors, including the generator that is chosen after observing the sampled inputs $(x^{(1)},\dots,x^{(m)})$.

\paragraph{Combining weak predictors from different sources via boosting.}
The preceding argument achieves error $\eps$ using homogeneous CoTs from a single source, without active or adaptive collection. Our protocol, however, treats CoTs from any single source as a limited resource. Rather than request enough CoTs from one source to reach error $\eps$, we request only enough to obtain a \emph{weak predictor}, and combine predictors from different sources through \emph{boosting} \citep{schapire1990strength,freund1997decision}. The algorithm is simple once this connection to boosting is established.

We begin with an E2E sample of size $m=\Tilde{O}(\eps^{-1}\VCdim(\ete{\cF}{T}_d))$ and maintain a distribution over its examples. In each round, we sample $\M=O(\VCdim(\ete{\cF}{T}_d))$ examples from the current distribution and request CoTs from a single source. By the preceding argument, invoking $\Cons$ on these traces yields, with constant probability, an E2E predictor with error at most $1/4$ under the current distribution, even when the source is chosen after observing the subsample. Thus, \emph{one round of homogeneous CoT collection and finding a consistent predictor} implements the computationally efficient weak learner needed for boosting.

We evaluate the predictor on the E2E sample, increase the weight of misclassified examples, and draw the next batch of $\M$ examples from the updated distribution. Boosting only requires each weak predictor to perform well under the current distribution, so the sources may change adaptively across rounds while the per-source CoT budget remains independent of $\eps$. After $O(\log m\log\log m)=O_{d,T}(\log\eps^{-1}\log\log\eps^{-1})$ rounds, the algorithm produces a weighted majority consistent with the full E2E sample. Uniform convergence over the class of these weighted-majority predictors then gives population error at most $\eps$.

The algorithm combines small homogeneous batches from different sources into an arbitrarily accurate predictor. This captures a key feature of iterative fine-tuning: successive rounds of training on actively selected examples and traces from evolving models or different teacher checkpoints can yield continued improvement \citep{zelikman2022star,singh2023beyond,grattafiori2024llama}.

\section{Related Work}\label{sec:related-work}
\paragraph{CoT supervision across training phases and curation practices.}
Training Transformer-based language models through next-token prediction was popularized by models such as GPT-2 \cite{Rad+19}. Since then, this objective has been used extensively to train models on CoT data at different stages of training.
Pretraining exposes models to mathematical and scientific text containing derivations and worked solutions, while mid-training can emphasize targeted reasoning data and synthetic traces \citep[e.g.,][]{lewkowycz2022solving,abdin2024phi}. During SFT, this supervision is provided explicitly through questions paired with step-by-step solutions \citep[e.g.,][]{abdin2025phi,guha2025openthoughts}.

Specialized datasets cover different question domains and difficulty levels, and combine reasoning formats ranging from natural-language explanations to executable programs \citep{yue2024mammoth,longpre2023flan}. Another example is the FLAN Collection, which brings together diverse task collections and reasoning datasets \citep{longpre2023flan}. Human-written solutions and responses from different teachers also vary in style and level of detail. Practical pipelines curate this supervision through format standardization, correctness checks, and heuristic or model-based filtering of response quality \citep{li2024numinamath,grattafiori2024llama,guo2025deepseek}. Selection can further account for compatibility with the student model \citep{zhang2026best}. Some studies \citep{touvron2023llama,zhang2026best,ye2025limo} also highlight the importance of quality over quantity: smaller, carefully selected datasets can outperform much larger collections. Such collection and curation practices are ubiquitous in current training pipelines and play an important role in realizing the benefits of CoT supervision.

\paragraph{Theoretical works on the power of CoT.} The ability of CoT supervision to make hard learning problems tractable has motivated a growing body of learning-theoretic work from different perspectives: the expressive power of transformers with CoT \cite{chen2025theoretical,li2024chain,merrill2023expressive}, gradient-based learning in transformers \cite{kim2024transformers,huang2025transformers}, and learnability in generic autoregressive concept classes \cite[e.g.,][]{malach2023auto,joshi2025theory,altabaa2025cot,pmlr-v336-rajaraman26a,hanneke2026sample,li2026optimal,magen2026surprising}. Our work follows the latter approach, focusing on heterogeneous CoT supervision. 

From a computational perspective, prior work studies consistency-based CoT learning, modeling vanilla fine-tuning and yielding efficient learning when next-token consistency is tractable \citep{malach2023auto,joshi2025theory,altabaa2025cot}. The work of \citep{pmlr-v336-rajaraman26a} studies adaptive fine-tuning from a single fixed teacher. Their algorithm achieves CoT query complexity $\poly\log(1/\eps)$, compared with the worst-case $\Omega(1/\eps)$ requirement for non-adaptive fine-tuning.\footnote{While we were working on other results, the authors of \cite{pmlr-v336-rajaraman26a} shared with us their active learning setup and their guarantee of learning with $\poly\log(1/\eps)$ active CoT queries. This inspired us to also consider active and adaptive learning from multiple sources.} Their algorithm, also based on boosting, is motivated by autocurriculum: adaptively selecting prompts to reduce the number of CoT queries to a fixed teacher. We develop ours from the perspective of model aggregation across heterogeneous sources, with adaptively varying contributions and a limited CoT budget per source.

On the sample complexity of learning from a single generator, initial sample-complexity bounds \citep{joshi2025theory} were subsequently improved \citep{hanneke2026sample}, with optimal bounds for realizable CoT learning established in \citep{li2026optimal}. \citep{doron2026stochastic} extends the framework to stochastic next-token generators and studies the sample and computational complexity of learning in this setting.

\paragraph{Hardness of supervised learning.}
The computational complexity of learning various hypothesis classes has been extensively studied in the past decades; see \cite{servedio2025probably} for a recent survey. Most relevant to our work are the distribution-free hardness results for improperly learning DNFs and intersections of halfspaces. \cite{klivans2009cryptographic} proved hardness of learning intersections of $O(n^\epsilon)$ halfspaces over $\{0,1\}^n$, for any $\epsilon>0$, assuming the hardness of worst-case lattice problems, such as the \emph{unique shortest vector problem}, as well as $\gapsvp$ and $\sivp$ which we discuss in \Cref{subsec:crypto}. Under a stronger assumption regarding the subexponential-time quantum hardness of $(\gapsvp)$ and $(\sivp)$ (\Cref{assump:lattice}), the construction of \cite{klivans2009cryptographic} implies hardness already for intersections of $\omega(\log n \log \log n)$ halfspaces; see \Cref{rem:tiegel-conclusion}. Moreover, \cite{tiegel2024improved} improved this construction and proved hardness for learning intersections of $\omega(\log\log n)$ halfspaces under \Cref{assump:lattice}, but when the domain is $\reals^n$. Hardness of learning DNFs is implied by \cite{applebaum2010public} under two assumptions: one concerning the planted dense subgraph problem in hypergraphs, and the other concerning local PRGs. \cite{daniely2016complexity} proved hardness of learning DNFs with $\omega(\log n)$ terms over $\{0,1\}^n$ assuming the hardness of refuting random $K$-SAT formulas. \cite{daniely2021local} proved hardness of learning DNFs with $\omega_n(1)$ terms under \Cref{assumption:prg_exists}. More recently, \cite{applebaum2025structured} proved hardness of learning DNFs assuming search hardness of sparse learning-parity-with-noise (sparse LPN). Note that these hardness results for learning DNFs also imply hardness of learning intersections of halfspaces (as a DNF with $q(n)$ terms can be expressed by the complement of an intersection of $q(n)$ halfspaces). Finally, we note that \Cref{assumption:prg_exists} was later used in \cite{daniely2023computational,attias2025hardness}, for establishing hardness of learning for various hypothesis classes.


\section{Preliminaries}\label{sec:preliminary}
In this section, we introduce all the preliminaries needed to understand our main theorems and proofs. The definitions of our learning problems are formalized in \Cref{subsec:def+notation}. We then go over some statistical complexity measures and standard learnability guarantees in the single generator case in \Cref{subsec:VC-of-Classes}. This is largely based on \cite{joshi2025theory}, and it is required for our learnability results. Finally, in \Cref{subsec:crypto} we state our cryptographic assumptions. We start with some notation.

\paragraph{Notation.} We use $O(\cdot)$, $\Omega(\cdot)$, and $\Theta(\cdot)$ to hide universal constants; subscripts indicate any additional parameters on which these constants depend. For $o(\cdot)$ and $\omega(\cdot)$, subscripts specify the asymptotic parameter when needed, e.g., $\omega_d(\cdot)$ and $o_d(\cdot)$. We write $\poly(\cdot,\dots,\cdot)$ for a fixed polynomial in the indicated parameters. In asymptotic upper bounds, logarithmic factors $\log u$ are understood as $\max\{1,\log u\}$. Finally, rarely, we use $\Tilde{O}$ notation for a fixed expression $\Psi$ to mean $\Tilde{O}(\Psi)=O(\Psi\,\poly \log \Psi)$. For any string $\z \in \{0,1\}^*$, we use $|\z|$ for its length and view it as indexed from left to right. In the context of learning with autoregressive CoT, we write $\z[\,i\,]$ for the $i^\mathrm{th}$ coordinate, and $\z[-i]:=\z[|\z|-i+1]$ for the $i^\mathrm{th}$ coordinate from the end. For coordinates not related to learning with CoT, we use the more familiar notation $\z_i$ for the $i^\mathrm{th}$ coordinate. We sometimes abbreviate the contiguous integers $(i,\dots,j)$ for $i \leq j$ by $i\!:\!j$.

\subsection{Definitions of learning scenarios}\label{subsec:def+notation}
We now formally describe our $(\eps,\delta)$-PAC learning definition for our learning scenarios.
 Recall the different models of supervision introduced in \Cref{subsec:multiple-CoT-supervision-intro}. We consider learning problems for any base hypothesis class $\cF\subseteq \{0,1\}^{\{0,1\}^*}$, domain $\cX \subseteq \{0,1\}^*$, generation length $T$, and the number of generators $\kappa$. For every learning algorithm, the adversary chooses a realizable joint distribution with input marginal $\cD$ and labels $y=h_*(x)$ for some unknown $h_*\in\ete{\cF}{T}$, together with $\kappa$ generators $(f_1,\dots,f_\kappa)\in\cF^\kappa$ that all realize $h_*$ end-to-end, as in \eqref{eq:e2e-agreement}.

We slightly generalize instance-dependent selection by allowing the contributing source to be chosen stochastically, according to a fixed, unknown conditional distribution $\cP(\cdot\mid x)$ over $[\kappa]$. This unifies \Cref{def:instance-dependent-bias} and \Cref{def:uniform-mixture}.

\begin{definition}[Models of supervision]\label{def:models-of-supervision}
Under the adversary's choices above, draw $m$ i.i.d. inputs $x^{(1)},\dots,x^{(m)}\simiid\cD$. Under \textbf{E2E supervision}, the learner observes only
\begin{equation}\label{eq:e2e-train-set-prelim}
    S_{\sete}
    =
    \left\{(x^{(i)},y^{(i)}): y^{(i)}=h_*(x^{(i)}), i\in[m]\right\}.
\end{equation}
Under \textbf{CoT supervision}, the adversary additionally fixes $\cP:\cX\to\Delta([\kappa])$ before the inputs are sampled. Conditional on these inputs, the source identities are drawn independently as $j^{(i)}\sim\cP(\cdot\mid x^{(i)})$, and the learner observes
\begin{equation}\label{eq:multiple-CoT-training-set-prelim}
    S_{\sCoT}
    =
    \left\{(x^{(i)},z^{(i)}):
    z^{(i)}=\CoT{f_{j^{(i)}}}{T}(x^{(i)}),\ i\in[m]\right\}.
\end{equation}
 The special case $\cP(\cdot\mid x)=\Unif([\kappa])$ for every $x\in\cX$ is called a \textbf{uniform mixture}.
We say that \textbf{provenance is known} if the learner additionally observes each $j^{(i)}$. Otherwise, provenance is unknown, and we call the traces \textbf{unfiltered}.
\end{definition}
When $\kappa=1$, all these supervision models of heterogeneous CoT supervision coincide with the homogeneous setting studied in \cite{joshi2025theory}. We study standard distribution-free PAC learning under different models of supervision.  
\begin{definition}[Distribution-free PAC learning]\label{def:distribution-free-PAC}
For any base class $\cF\subseteq \{0,1\}^{\{0,1\}^*}$, domain $\cX \subseteq \{0,1\}^*$, generation length $T$, and number of generators $\kappa$, we say that the class $\ete{\cF}{T}$ is learnable over $\cX$ under the respective model of supervision (cf. \Cref{def:models-of-supervision}), with sample complexity $m(\eps,\delta)$, if there exists a learning rule $\cA$ such that, for every $\eps,\delta \in (0,1)$ and every choice made by the adversary, given a training set of size $m\geq m(\eps,\delta)$ sampled as specified in the respective model of supervision in \Cref{def:models-of-supervision}, the learning rule $\cA$ returns a predictor $\widehat{h}:\cX \to \{0,1\}$ such that, with probability at least $1-\delta$ over the draw of the training set and the learner's internal randomness,
$$
\cL_{\cD,h_*}(\widehat{h}) = \Pr_{x\sim\cD}\left( \widehat{h}(x)\neq h_*(x) \right) \leq \eps \,.
$$
\end{definition}

\paragraph{Runtime of a learning algorithm.}
We use $d$ as the problem-size parameter and consider a hierarchy of learning problems over the growing domains $\cX_d\subseteq \{0,1\}^{\le d}$. Runtime includes both the time to output a representation of the hypothesis $\widehat{h}$ and the time to evaluate $\widehat{h}(x)$ on a new input $x\in\cX_d$. We measure runtime in terms of $d$ and explicitly state its dependence on $T,\kappa,\eps,\delta$, or how these parameters relate to $d$. For our learnability results, we assume that each generator $f\in\cF_d$ computes the next token $f(\z)$ in time $\poly(d,|\z|)$ for every $\z\in\{0,1\}^*$, as holds for $\cFlin$. 

\paragraph{Tractable consistency oracle for autoregressive linear thresholds.}
Recall the class of autoregressive linear thresholds without bias,
\[
\cFlin
=
\left\{
f_w: \z \mapsto \th\left(\sum_{i=1}^{\min\{d,|\z|\}}w[-i]\z[-i]\right) \mid 
w\in\mathbb{R}^d
\right\}.
\]
 Given realizable prefix--next-token pairs, consistency reduces to linear feasibility in $w$. Thus, $\Cons$ (\Cref{def:cons-oracle}) can be implemented in polynomial time using linear programming \citep{joshi2025theory}.

\begin{remark}[Polynomially bounded weights]\label{rem:poly-bounded-w}
For the restricted class with integer weights bounded in magnitude by a fixed polynomial in $d$, i.e., $O(\log d)$ bit complexity, even perceptron or gradient descent on the logistic loss finds a consistent $\widehat{w}$ in polynomial time, since the normalized margin is inverse-polynomial. Although our reductions do not formally track weight magnitudes, we believe that all our hardness results already hold under this restriction. In any case, our main focus remains the full class $\cFlin$, for which $\Cons$ can already be implemented in polynomial time.
\end{remark}

\subsection{Statistical complexity measures} \label{subsec:VC-of-Classes}
In this section, we define our statistical complexity measure and give an overview of statistical learnability results for our settings. All technical lemmas are borrowed from \cite{joshi2025theory}.
\paragraph{Growth function and VC dimension.} We begin by defining some standard terminology from supervised learning. Consider any abstract spaces $\cX$ and $\cY$, and a hypothesis class $\cH\subseteq \cY^\cX$.

\begin{definition}[Growth function]\label{def:growth-function}
For any $h:\cX \to \cY$ and any $S=(x^{(1)},\dots,x^{(m)}) \in \cX^m$, we define
\( 
h(S):=(h(x^{(1)}),\dots, h(x^{(m)}))\) and \(
\cH(S):=\{h(S): h \in \cH \}.
\) The growth function $\Gamma_{\cH}: \naturals \to \naturals$ is defined by
\[
\Gamma_{\cH}(m)= \max_{(x^{(1)}, \dots, x^{(m)}) \in \cX^m} \left| \{(h(x^{(1)}), \dots, h(x^{(m)})) : h \in \cH\} \right|
= \max_{S \in \cX^m} |\cH(S)|.
\]
We will also abuse notation and write $\Gamma_{\cH}(S)=|\cH(S)|$ for the number of behaviors of $\cH$ on $S$.
\end{definition}
When $\cY=\{0,1\}$ is binary, we can define the VC dimension of the class and its relationship with the growth function via Sauer's lemma; see \Cref{lem:sauer'slemma}.
\begin{definition}[VC dimension]\label{def:VCdim}
When $\cY=\{0,1\}$, for any $\cH \subseteq \{0,1\}^{\cX}$, we say that $\cH$ shatters a set of points $S \in \cX^m$ iff $\left| \cH(S)\right| = 2^{|S|}.$ The Vapnik--Chervonenkis dimension of the class, denoted by $\VCdim(\cH)$, is the largest integer $D$ such that $\Gamma_{\cH}(D)=2^D.$ If no largest such $D$ exists, then we say that $\VCdim(\cH)=\infty$\,.
\end{definition}
\paragraph{E2E and CoT consistent rules.} For our problems, consider two standard learning rules. First, under E2E supervision, given $S_{\sete}=\{(x^{(i)},y^{(i)})\mid i \in [m]\}$\,:
\begin{equation}\label{eq:ete-consistent}
    \text{Return $\ete{\widehat{f}}{T}$, for some $\widehat{f}\in \cF $ such that for all } i \in [m],\,\,  y^{(i)}=\ete{\widehat{f}}{T}(x^{(i)})\,. \tag{$\Consete$} 
\end{equation}
When the learner additionally observes CoTs from a single generator, it can further use this information and consider the following rule. Given
\( 
S_{\sCoT}=\{(x^{(i)},z^{(i)})\mid  i \in [m]\}
\)\,:
\begin{equation}\label{eq:CoT-Conc}
\text{Return $\ete{\widehat{f}}{T}$, for some $\widehat{f}\in \cF$ such that for all } i \in[m],\,\,\, z^{(i)}=\CoT{\widehat{f}}{T}(x^{(i)})\,\,.\tag{$\Conscot$}
\end{equation}
For a fixed domain $\cX$ and generation length $T$, consider $\VCdim(\ete{\cF}{T})$ (\Cref{def:VCdim}) over $\cX$. This characterizes learnability under E2E supervision using the learning rule \ref{eq:ete-consistent}. To analyze \ref{eq:CoT-Conc}, under the availability of CoTs from a single thinker, we consider the following CoT loss class:
\begin{equation}\label{eq:loss-class-main}
    \CoT{\cL}{T}(\cF):=\{ \ell_f:(x,z) \mapsto \ind\{z \neq \CoT{f}{T}(x)\} \mid f \in \cF\}\,.
\end{equation}
This is a binary class over the domain $\cX \times \{0,1\}^T$. \cite{joshi2025theory} provided a bound on the sample complexity of CoT from a single thinker in terms of the VC dimension of the loss class $\CoT{\cL}{T}(\cF)$. When CoTs from a single generator are provided, the resulting worst-case sample-complexity bound can improve upon the bound under E2E supervision; see \Cref{prop:VC-base-class-relation-ship}. For now, we state the statistical learnability results under E2E supervision and single-thinker CoT supervision in terms of these complexity measures.

In particular, we have the following sample complexity bounds, which follow directly from standard uniform convergence guarantees for realizable learning (\Cref{prop:fundamental-VC}), applied to the E2E function class $\ete{\cF}{T}$ and the CoT loss class $\CoT{\cL}{T}(\cF)$.

\begin{proposition}[Theorems 3.1, 3.2, 3.4 \cite{joshi2025theory}]\label{prop:UC-of-e2e-CoT} There exists a universal constant $c>0$ such that the following holds. For any base class $\cF \subseteq \{0,1\}^{\{0,1\}^*}$, thinking length $T$, domain $\cX\subseteq \{0,1\}^*$, the rule \ref{eq:ete-consistent} learns the class $\ete{\cF}{T}$ with sample complexity
$$m(\eps,\delta) \leq c \left( \frac{\min\{\log |\cF|,\VCdim(\ete{\cF}{T})\log(2/\eps)\}+\log \delta^{-1}}{\eps}\right)\,.$$
The rule \ref{eq:CoT-Conc} learns the class $\ete{\cF}{T}$ under the CoT supervision from a fixed single generator chosen before observing the sampled inputs $(x^{(1)},\dots,x^{(m)})$ with sample complexity
$$m(\eps,\delta) \leq c \left( \frac{\min\{\log |\cF|,\VCdim(\CoT{\cL}{T}(\cF))\log(2/\eps)\}+\log \delta^{-1}}{\eps}\right)\,.$$
\end{proposition}

\paragraph{Benefits of autoregressive function classes.} The central message of \cite{joshi2025theory} is the benefit of autoregressive learning. They show that, by applying the same function autoregressively, one obtains a form of parameter sharing, and thus the statistical complexity of learning the class $\ete{\cF}{T}$ when fixed single-thinker CoT is provided remains nearly independent of $T$, in terms of the VC dimension of the base class $\cF$. Note that $\VCdim(\cF)$ can be thought of as the statistical complexity measure for the corresponding single-step problem of learning to predict the next token. Formally, they prove the following:
\begin{proposition}[Theorems B.1, B.5 \cite{joshi2025theory}]\label{prop:VC-base-class-relation-ship}
For any $\cF\subseteq \{0,1\}^{\{0,1\}^*}$ and thinking length $T$,
$$\VCdim(\ete{\cF}{T}) \leq O\left(T\cdot \VCdim(\cF)\right)
\quad \text{and} \quad
\VCdim(\CoT{\cL}{T}(\cF))=O\left(\VCdim(\cF)\log T\right)\,.$$
Here, for any domain $\cX\subseteq \{0,1\}^*$, the quantity $\VCdim(\cF)$ is taken over the domain $\cX \times \{0,1\}^{\leq T-1}$ in order to bound the quantities on the left over the domain $\cX$.
\end{proposition}
We note that \cite[Theorem 3.3]{joshi2025theory} also shows that the rule $\Consete$ may require $\Omega(T\cdot \VCdim(\cF))$ samples for some class $\cF$, and thus the linear dependence on $T$ is necessary. On the other hand, combining \Cref{prop:VC-base-class-relation-ship,prop:UC-of-e2e-CoT} implies that, under supervision from a single thinker, the rule \ref{eq:CoT-Conc} learns the class $\ete{\cF}{T}$ with sample complexity depending logarithmically on $T$, as long as $\VCdim(\cF)$ is independent of $T$.

\subsection{Cryptographic assumptions}\label{subsec:crypto}
In this subsection, we state our cryptographic hardness assumptions. In particular, we assume (i) subexponential quantum hardness of some worst-case lattice problems, namely, $\gapsvp$ and $\sivp$, and (ii) the existence of Goldreich's local PRGs with polynomial stretch \cite{goldreich2011candidate}. We do not expect the reader to have prior knowledge on the technical aspects of these problems, cryptography, or the assumptions themselves, to understand the rest of the paper. 
\paragraph{Hardness of worst-case lattice problems \texorpdfstring{$\gapsvp$}{} and \texorpdfstring{$\sivp$}{}.}
An \(N\)-dimensional lattice \(L\) is a discrete additive subgroup of \(\R^N\). It can be specified by a basis \(B=[b_1,\dots,b_N] \in \R^{N \times N}\) as \(L = B\Z^N\). The special case we focus on is when $B$ is full-rank. For a lattice $L$ and \(1 \le i \le N\), define its $i^\mathrm{th}$ successive minimum as
\[
\lambda_i(L)
\;:=\;
\inf\Bigl\{r>0 : \dim\bigl(\spn(L \cap B_r(0))\bigr)\ge i\Bigr\},
\]
where \(B_r(0) = \{x \in \R^N : \|x\|_2 \le r\}\) denotes the Euclidean ball of radius \(r\) centered at the origin. Let $\alpha=p(N)$ be an arbitrary polynomial. We will consider two problems on lattices:
\begin{problem}[Gap Shortest Vector Problem $(\gapsvp)$]\label{prb:gap-SVP}
    For an approximation factor \(\alpha=\alpha(N)\), an instance of $\gapsvp_{\alpha}$ consists of an \(N\)-dimensional lattice \(L\) and a number \(\theta>0\), under the promise that either
\(
\lambda_1(L) \le \theta
\)
or
\(
\lambda_1(L) > \alpha \cdot \theta
\).
The goal is to decide which of the two cases holds.
\end{problem}
\begin{problem}[Shortest Independent Vector Problem $(\sivp)$]\label{prb:sivp} For an approximation factor $\alpha=\alpha(N)$, an instance of $\sivp_\alpha$ consists of an \(N\)-dimensional lattice \(L\). The goal is to output $N$ linearly independent lattice vectors, each of length at most $\alpha\cdot\lambda_N(L)$.
\end{problem}
We make the following assumption. 
\begin{hassumption}\label{assump:lattice}
For any polynomial $p(N)$, at least one of $\gapsvp_{\alpha=p(N)}$ (\Cref{prb:gap-SVP}) or $\sivp_{\alpha=p(N)}$ (\Cref{prb:sivp}) has no quantum algorithm that runs in time $2^{o(N)}$.
\end{hassumption}
All known classical and quantum algorithms for $\gapsvp_{\alpha}$ and $\sivp_{\alpha}$, for polynomial approximation factors, require time $2^{\Omega(N)}$. A falsification of the above assumption would therefore constitute a major breakthrough in lattice algorithms and cryptography; see, e.g., \cite{peikert2016decade} and the references therein. A weaker variant, assuming only the absence of polynomial-time algorithms for certain polynomial approximation factors, was used in the classical work of \cite{klivans2009cryptographic}, which gave the first representation-independent hardness result for learning intersections of $n^\epsilon$ halfspaces for any fixed $\epsilon$. The exact assumption was recently used in hardness-of-learning results by \cite{tiegel2023hardness,tiegel2024improved}. 

\paragraph{Existence of local Pseudo Random Generators (PRGs).} Fix integers \(N,M\) and a constant \(\klocal\). An \((N,M,\klocal)\)-hypergraph \(G\) consists of \(M\) ordered hyperedges
\[
S_1,\dots,S_M \in [N]^{\klocal},
\]
where each ordered hyperedge $S_i$ has $\klocal$ distinct coordinates. Let \(P : \{0,1\}^{\klocal} \to \{0,1\}\) be a Boolean predicate. Given a hypergraph \(G\), we define the \emph{Goldreich pseudorandom generator} \citep{goldreich2011candidate}
\( 
f_{P,G} : \{0,1\}^N \to \{0,1\}^M
\)
by: 
\[ 
f_{P,G}(x)
=
\bigl(P(x_{S_1}), \dots, P(x_{S_M})\bigr) \quad \text{ for all \(x \in \{0,1\}^N\)\,,}
\]
where \(x_{S_i}\) denotes the restriction of \(x\) to the coordinates indexed by \(S_i\).

The parameter \(\klocal\) is called the \emph{locality} of the generator. When \(\klocal\) is constant, both the predicate \(P\) and the generator \(f_{P,G}\) are said to be \emph{local}. The PRG is said to have \emph{polynomial stretch} if \(M = N^s\) for some constant \(s > 1\).

Let
\[
\cF_{P,N,M}
:=
\{ f_{P,G} : G \text{ is an } (N,M,\klocal)\text{-hypergraph} \}
\]
denote the family of generators. Let \(\cG_{N,M,\klocal}\) be the distribution over such hypergraphs obtained by sampling each hyperedge independently and uniformly at random from all ordered \(\klocal\)-tuples of distinct elements of \([N]\). We write \(x \sim\{0,1\}^N\) for a uniform draw from \(\{0,1\}^N\).

We say that \(\cF_{P,N,M}\) is an \(\epsilon\)-pseudorandom generator (or \(\epsilon\)-PRG) if, for every probabilistic polynomial-time algorithm \(\mathcal{A}\), the following holds for all sufficiently large \(N\):
\begin{align*}
\left|
\Pr_{\substack{G \sim \cG_{N,M,\klocal} \\ x \sim \{0,1\}^N}}
\bigl[ \mathcal{A}(G, f_{P,G}(x)) = 1 \bigr]
-
\Pr_{\substack{G \sim \cG_{N,M,\klocal} \\ y \sim \{0,1\}^M}}
\bigl[ \mathcal{A}(G, y) = 1 \bigr]
\right|
\le \epsilon.
\end{align*}
That is, given \(G\), the distinguishing advantage between a random output of \(f_{P,G}\) and a uniform \(y \in \{0,1\}^M\) is at most \(\epsilon\). We make the following assumption:
\begin{hassumption}[Existence of local PRG with polynomial stretch]\label{assumption:prg_exists} 
For every constant \(s>1\), there exists a constant \(\klocal\) and a predicate \(P : \{0,1\}^{\klocal} \to \{0,1\}\) such that \(\cF_{P,N,M}\), with \(M = N^s\), is a \(\tfrac{1}{3}\)-PRG.
\end{hassumption}
This assumption was used by \cite{daniely2021local,daniely2023computational,attias2025hardness} to prove hardness for several concept classes (see \cite{daniely2021local} for further discussion on the assumption).

\section{Additional results}\label{sec:additional-results}
In this section, we present additional results mentioned informally earlier in the presentation.

\subsection{A generalization bound and efficient learner for \texorpdfstring{$\passk$}{} error}\label{subsec:passk-efficient}
In this subsection, we formalize the efficient learner for $\passk$ error under instance-dependent contribution biases (\Cref{thm:passk}), discussed in \Cref{subsec:hard-to-learn-contribution-bias}. The main ingredient is a generalization bound for $\passk$ accuracy (\Cref{thm:satistical_upper_bound}).

Recall the generalized instance-dependent selection model in \Cref{def:models-of-supervision}: each input $x^{(i)}\sim\cD$ receives a trace $z^{(i)}=\CoT{f_{j^{(i)}}}{T}(x^{(i)})$, where $j^{(i)}\sim\cP(\cdot\mid x^{(i)})$. We derive our generalization bound even with \emph{unknown provenance}, where the learner observes $(x^{(i)},z^{(i)})$ but not $j^{(i)}$. Consider the following learning rule.

\begin{tcolorbox}
Given $S_{\sCoT}=\{(x^{(i)},z^{(i)}):i\in[m]\}$, find $\widehat f_1,\dots,\widehat f_\kappa\in\cF$ such that
\begin{equation}\label{eq:kappa-cons-learning-rule}
\forall i\in[m],\quad z^{(i)}\in\left\{\CoT{\widehat f_j}{T}(x^{(i)}):j\in[\kappa]\right\}.
\tag{$\kappa\textnormal{-}\Conscot$}
\end{equation}
On a test input $x\in\cX$:
\begin{itemize}
    \item For $\passkappa$ evaluation, return the candidate set $\widehat{\mathcal Z}(x)=\left\{\CoT{\widehat f_j}{T}(x):j\in[\kappa]\right\}$.
    \item For E2E prediction, additionally require $\ete{\widehat f_1}{T}\equiv\dots\equiv\ete{\widehat f_\kappa}{T}\equiv\widehat h$ and return the common final label $\widehat h(x)=\ete{\widehat f_1}{T}(x)=\dots=\ete{\widehat f_\kappa}{T}(x)$.
\end{itemize}
\end{tcolorbox}
We now bound the $\passk$ error of this rule in terms of the VC dimension of the following binary loss class over $\cX\times\{0,1\}^T$:
\begin{equation}\label{eq:loss-class-kappa}
\CoT{\cL}{T}(\cF;\kappa)
:=
\left\{
(x,z)\mapsto
\ind\!\left\{\forall j\in[\kappa],\ z\neq\CoT{g_j}{T}(x)\right\}
:
g_1,\dots,g_\kappa\in\cF
\right\}.
\end{equation}
The following theorem gives the resulting sample-complexity guarantee. Under the E2E agreement constraint, the bound also controls the error of the common E2E predictor.

\begin{theorem}[A generalization bound]\label{thm:satistical_upper_bound}
There is a universal constant $c>0$ such that the following holds. For any base class $\cF\subseteq\{0,1\}^{\{0,1\}^*}$, input domain $\cX\subseteq\{0,1\}^*$, and generation length $T$, consider instance-dependent CoT supervision with unknown provenance (\Cref{def:models-of-supervision}) from $\kappa$ generators $f_1,\dots,f_\kappa\in\cF$ satisfying $\ete{f_1}{T}\equiv\dots\equiv\ete{f_\kappa}{T}\equiv h_*$. For every $\eps,\delta\in(0,1)$, if
\[
m\geq c\,\frac{\VCdim(\CoT{\cL}{T}(\cF;\kappa))\log(2/\eps)+\log(1/\delta)}{\eps},
\]
then, with probability at least $1-\delta$ over the training set, the rule \ref{eq:kappa-cons-learning-rule} satisfies
\[
\Pr_{x\sim\cD,\,j\sim\cP(\cdot\mid x)}\left[\CoT{f_j}{T}(x)\notin\widehat{\mathcal Z}(x)\right]\leq\eps
\quad\text{ and }\quad
\cL_{\cD,h_*}(\widehat h)\leq\eps,
\]
where the second bound holds when the E2E agreement constraint is also imposed. Here, the loss class satisfies
\[
\VCdim(\CoT{\cL}{T}(\cF;\kappa))\leq c\,\kappa\VCdim(\cF)\log(2\kappa T).
\]
Here, $\VCdim(\cF)$ is taken over the prefix domain $\cX\times\{0,1\}^{\leq T-1}$.
\end{theorem}
To prove the theorem, we bound the VC dimension of the loss class in \eqref{eq:loss-class-kappa} and apply realizable uniform convergence (\Cref{prop:fundamental-VC}), since the rule has zero empirical loss with respect to this loss class. When E2E agreement is imposed, matching any target trace also yields the correct common final label. The full proof appears in \Cref{sec:sample-complexity-kappa<<T}.

\paragraph{Efficient $\passkappa$ learning with known provenance.}
Known provenance lets us identify homogeneous subsets and apply $\Cons$ separately to each, efficiently implementing the $\passkappa$ version of the learning rule \ref{eq:kappa-cons-learning-rule}, giving the following guarantee.

\begin{theorem}[Efficient learner for $\passkappa$]\label{thm:passk}
There is a learning algorithm that for any hierarchy of base classes $(\cF_d)_{d\geq1}$ admitting an efficient $\Cons$ oracle (\Cref{def:cons-oracle}) and satisfying $\VCdim(\cF_d)\leq\poly(d,T)$ over the domain $\cX_d\times \{0,1\}^{\leq T-1}$, under instance-dependent CoT supervision with known provenance (\Cref{def:models-of-supervision}), for any $\eps,\delta\in(0,1)$, with probability at least $1-\delta$ over the training sample, returns a predictor $\widehat{\cZ}:\cX_d\to 2^{\{0,1\}^T}$ satisfying $\max_{x\in \cX_d}|\widehat{\cZ}(x)|\leq\kappa$ and
\[
\Pr_{x\sim\cD,\,j\sim\cP(\cdot\mid x)}
\left[
\CoT{f_j}{T}(x)\notin\widehat{\cZ}(x)
\right]\leq\eps.
\]
The algorithm runs in time $\poly(d,T,\kappa,\eps^{-1},\log\delta^{-1})$.
\end{theorem}

\begin{proof} The learner observes
$S_{\sCoT}=\left\{(x^{(i)},j^{(i)},z^{(i)}):i\in[m]\right\}\,.$ With known provenance, the learner applies $\Cons$ to the prefix--next-token pairs from each generator's traces, obtaining $\widehat f_1,\dots,\widehat f_\kappa$, and returns $\widehat{\cZ}(x)=\{\CoT{\widehat f_j}{T}(x):j\in[\kappa]\}$. This implements the rule \ref{eq:kappa-cons-learning-rule}, so \Cref{thm:satistical_upper_bound} gives the $\passkappa$ guarantee. Since $\VCdim(\cF_d)\leq\poly(d,T)$, the required sample size is polynomial. The $\kappa$ oracle calls on at most $mT$ prefix--next-token pairs, with evaluating the fitted generators, give the claimed runtime.
\end{proof}
\begin{remark}[Improved sample complexity for E2E error when $\kappa\ll T$]
Beyond efficient $\passkappa$ learning, \Cref{thm:satistical_upper_bound} also gives improved sample complexity for E2E error when the common E2E agreement constraint is imposed. The bound scales with $\kappa\VCdim(\cF)\log(\kappa T)$, whereas imposing only E2E consistency and ignoring the CoTs can incur a dependence of $T\cdot\VCdim(\cF)$ in the worst case \cite[Theorem 3.3]{joshi2025theory}. This extends the statistical benefit of single-generator CoT supervision (\Cref{prop:VC-base-class-relation-ship}) to multiple generators, with a substantial improvement when $\kappa \ll T$. For fixed $\kappa$ and $\VCdim(\cF)$, the dependence on $T$ remains only logarithmic.
\end{remark}

\subsection{Hardness of E2E learning for short generation length \texorpdfstring{$T$}{}}\label{subsec:ete-hardness}

In this subsection, we establish hardness of E2E learning for short generation lengths through a reduction from learning small depth-$2$ threshold circuits (\Cref{thm:ete-hardness}), mentioned in \Cref{rem:e2e-hardness-for-small-T}. We first formally state the expressivity lemma that enables this reduction. 

\begin{lemma}[Formal version of \Cref{lem:informal-expressivity-wo-CoT}]\label{lemma:LT-expressivity-depth-2-d=n'}
For any input size $n$ and circuit size $s$, there is a $\poly(n,s)$-time computable feature map $\phi:\{0,1\}^n\to\{0,1\}^d$, with $d=n(2s-1)+4s-3$, such that every depth-$2$ threshold circuit $C$ with $s$ gates can be simulated by some $f_{\vw}\in\cFlin$ in $T=2s-1$ steps. Specifically, for every $\vx\in\{0,1\}^n$,
\begin{equation}\label{eq:CoT-computation}
\CoT{f_{\vw}}{T}(\phi(\vx))
=
[C_1(\vx),\dots,C_{s-1}(\vx),0^{s-1},C(\vx)],
\end{equation}
where $C_1,\dots,C_{s-1}$ are the hidden gates in a fixed order. The feature map $\phi$ depends only on $(n,s)$.
\end{lemma}

The construction also preserves CoT leakage, a property needed for our hardness results under heterogeneous CoT supervision (\Cref{thm:contribution-bias-hardness,thm:filtering-hardness}).
However, the E2E expressivity alone suffices to establish E2E hardness. Since E2E learning is a standard supervised learning problem, an efficient learner for $\ete{\cFlin}{T}$ would also learn any hard-to-learn circuit class expressed by the simulation.

\begin{tcolorbox}[colback=white]
\begin{theorem}\label{thm:ete-hardness}
Let $g(n)\leq\poly(n)$ be any non-decreasing sequence computable in $\poly(n)$ time. Assuming there is no $\poly(n)$-time algorithm that weakly learns depth-$2$ threshold circuits of size $g(n)$ over $\{0,1\}^n$, there is no $\poly(d)$-time algorithm that weakly learns $\ete{\cFlin}{T(d)}$ for all $T(d)\leq2g(d)$.
\end{theorem}
\end{tcolorbox}
We prove \Cref{thm:ete-hardness,lemma:LT-expressivity-depth-2-d=n'} in \Cref{sec:proof-of-ete-hardness}.
\paragraph{Implications to hardness for short generation length.}

Combining \Cref{thm:ete-hardness} with known hardness results for small depth-$2$ threshold circuits and their special cases, including DNFs and intersections of halfspaces \cite{klivans2009cryptographic,applebaum2010public,daniely2016complexity,daniely2021local,tiegel2024improved,applebaum2025structured}, yields hardness of E2E learning for short generation lengths.

\begin{corollary}\label{cor:ete-hardness-cor-1}
The following statements hold.
\begin{itemize}
    \item Assuming the subexponential-time quantum hardness of $\gapsvp$ or $\sivp$ (\Cref{assump:lattice}), for any non-decreasing sequence $g(d)=\omega_d(\log d\log\log d)$ computable in $\poly(d)$ time, there is no $\poly(d)$-time algorithm that weakly learns $\ete{\cFlin}{T(d)}$ for all $T(d)\leq2g(d)$.
    \item Assuming the random $K$-SAT hypothesis \cite{daniely2016complexity}, the same conclusion holds for any non-decreasing sequence $g(d)=\omega_d(\log d)$ computable in $\poly(d)$ time.
    \item Assuming the existence of local PRGs with polynomial stretch (\Cref{assumption:prg_exists}), the same conclusion holds for any non-decreasing sequence $g(d)=\omega_d(1)$ computable in $\poly(d)$ time.
\end{itemize}
\end{corollary}

The first part follows from the construction of \cite{klivans2009cryptographic} by padding the inputs with zeros and using the stronger subexponential-time hardness assumption (\Cref{assump:lattice}). We formalize this when proving \Cref{thm:hardness-with-two-gens} in \Cref{subsec:proof-of-two-hardness}, since hardness under CoT supervision also implies E2E hardness. The second and third parts follow from \cite{daniely2016complexity} and \cite{daniely2021local}, respectively. Under the sparse-LPN assumption in \cite[Corollary 5]{applebaum2025structured}, we also obtain the same conclusion of the third part with $g(d)=\omega_d(1)$; see \Cref{sec:related-work}.

\Cref{cor:ete-hardness-cor-1} improves upon the generation length $T=\Theta(d)$ for which \cite{joshi2025theory} established hardness. Standard padding extends their result to $T=\Theta(d^\epsilon)$ for any fixed $\epsilon>0$.

Moreover, a negative resolution to the problem of learning intersections of a constant number of halfspaces---a longstanding open problem in learning theory \cite{khot2008hardness,klivans2009cryptographic}---would imply hardness for constant generation length $T$. It remains open to directly establish hardness for constant generation length $T$, or even hardness of \emph{proper} learning in this regime.\footnote{Unlike the representation-independent hardness results, our expressivity lemma does not imply hardness of proper learning for constant $T$, even though hardness of (semi-)properly learning intersections of a constant number of halfspaces is known \cite{blum1988training,alekhnovich2004learnability,khot2008hardness}.}

\subsection{Formal algorithm and guarantee under active and adaptive CoT collection}\label{subsec:formal-boosting-alog}

In this subsection, we formally state our algorithm (\Cref{alg:adaboost-cot}) and a more detailed guarantee for \Cref{thm:boosting}. Recall the active and adaptive CoT collection protocol (\Cref{def:active-adaptive-collection}): the learner selects examples on which to request homogeneous CoTs, while the adversary chooses each generator after observing the requested examples and previous rounds.

\begin{algorithm}[!ht]
\caption{Boosting-based learner under the active and adaptive protocol}
\label{alg:adaboost-cot}
\begin{algorithmic}[1]
\Require Training set $S_{\sete}=\{(x^{(i)},y^{(i)})\}_{i=1}^m$, hyperparameters $K$ and per-source budget $\M$, and accuracy/confidence parameters $\eps,\delta\in(0,1)$.
\Ensure An (improper) boosted learner $\widehat h_K:\cX \to \{0,1\}$.

\State Initialize $D^{(1)} \gets (1/m,\dots,1/m)\in\mathbb{R}^m$.
\For{$k=1,\dots,K$}
    \For{$\ell=1,\dots,\lceil\log_2(2K/\delta)\rceil$}
        \State Draw $i_1,\dots,i_{\M}$ i.i.d.\ from $D^{(k)}$ and set $\cS_{k,\ell}=\{i_1,\dots,i_{\M}\}\subseteq[m]$.
        \State Obtain $\{z^{(i)}=\CoT{f_{k,\ell}}{T}(x^{(i)}):i\in\cS_{k,\ell}\}$ from $f_{k,\ell}$ chosen by the adversary.
        \State Find $\widehat f_{k,\ell}$ by applying $\Cons$ (\Cref{def:cons-oracle}) to the prefix--next-token pairs in these CoTs.
        \State Compute its end-to-end empirical error:
        \[
        \epsilon_{k,\ell} \gets \sum_{i=1}^m D_i^{(k)} \cdot \mathbf{1}\!\left[\ete{\widehat f_{k,\ell}}{T}(x^{(i)})\neq y^{(i)}\right].
        \]
        \If{$\epsilon_{k,\ell}\leq0.25$}
            \State Break out of the inner loop.
        \EndIf
    \EndFor
    \State Let $\widehat f_k$ be the predictor from the last attempt and $\epsilon_k$ its error on $D^{(k)}$.
    \If{$\epsilon_k>0.25$}
        \State \Return the constant-$0$ predictor.
    \EndIf
    \If{$\epsilon_k=0$}
        \State \Return $\ete{\widehat f_k}{T}$.
    \EndIf
    \State Set
    \[
    \alpha_k \gets \frac12\log\!\left(\frac{1-\epsilon_k}{\epsilon_k}\right).
    \]
    \State Update the distribution: for all $i\in[m]$,
    \[
    D_i^{(k+1)} \propto D_i^{(k)}\times
    \begin{cases}
    e^{+\alpha_k} & \text{if } \ete{\widehat f_k}{T}(x^{(i)})\neq y^{(i)},\\
    e^{-\alpha_k} & \text{if } \ete{\widehat f_k}{T}(x^{(i)})=y^{(i)},
    \end{cases}
    \]
    normalizing so that $\sum_{i=1}^m D_i^{(k+1)}=1$.
\EndFor
\State \Return the weighted-majority predictor (break ties in favor of $1$):
\[
\widehat h_K(x)=\arg\max_{b\in\{0,1\}}
\left[\sum_{k=1}^K \alpha_k\,\mathbf{1}\!\left[\ete{\widehat f_k}{T}(x)=b\right]\right],
\]

\end{algorithmic}
\end{algorithm}

\begin{theorem}[More technical version of \Cref{thm:boosting}]\label{thm:boosting-model-aggregation}
There is a universal constant $c>0$ such that the following holds. For any base class $\cF \subseteq \{0,1\}^{\{0,1\}^*}$, generation length $T$, domain $\cX\subseteq \{0,1\}^*$, distribution $\cD \in \Delta(\cX)$, target $h_*\in\ete{\cF}{T}$, and parameters $\eps,\delta \in (0,1)$, after drawing an end-to-end training set
\[
S_{\sete}=\{(x^{(i)},y^{(i)})\}_{i=1}^m,
\qquad\text{where } x^{(1)},\dots,x^{(m)} \simiid \cD
\quad\text{and}\quad
y^{(i)}=h_*(x^{(i)}) \text{ for all } i\in [m],
\]
of any size
\[
m \geq m(\eps,\delta)=
c\left(
\frac{
\VCdim(\ete{\cF}{T}) \log(2/\eps)
\log^2\!\left(\frac{2\VCdim(\ete{\cF}{T})}{\eps}\right)+\log \delta^{-1}
}{\eps}
\right)\,,
\]
\Cref{alg:adaboost-cot}, which then proceeds to collect CoTs under the protocol in \Cref{def:active-adaptive-collection}, with parameters
\[
K=\lceil8\log(2m)\rceil \qquad \text{and} \qquad \M=O(\VCdim(\ete{\cF}{T})),
\]
returns, with probability at least $1-\delta$ over the draw of $S_{\sete}$ and the algorithm's internal randomness, a predictor $\widehat h_K$ with \(
L_{\cD,h_*}(\widehat{h}_K) \leq \eps.
\)
The number of generators used by the algorithm satisfies
\[
\kappa \leq K\left\lceil\log_2(2K/\delta)\right\rceil.
\]
Moreover, for a hierarchy $(\cF_d)_d$ over the input domain $\cX_d\subseteq \{0,1\}^{\leq d}$, choosing $m=\lceil m(\eps,\delta)\rceil$, the overall runtime of the algorithm is bounded by $\poly(d,T,\eps^{-1},\log \delta^{-1})$, as long as $(\cF_d)_d$ has a tractable consistency oracle (\Cref{def:cons-oracle}) and $\VCdim(\ete{\cF_d}{T})\leq\poly(d,T)$.

Finally, in the relaxed protocol, in which the adversary chooses $f_k$ before observing the subset $\cS_k$ for each $k\in [\kappa]$, the same learning guarantee holds but with a per-source query budget of $\M=O(\VCdim(\CoT{\cL}{T}(\cF)))=O(\min\{\log|\cF|,\VCdim(\cF)\log T\})$, where $\VCdim(\cF)$ is over the domain $\cX \times \{0,1\}^{\leq T-1}$.
\end{theorem}

The proof follows the outline in \Cref{subsec:technical-boosting}, with full details in \Cref{sec:analysis-of-alogorithm}. See \Cref{alg:adaboost-cot}: each attempt produces a predictor of error at most $1/4$ under the current weighted distribution over the E2E sample with probability at least $1/2$, by uniform convergence in a realizable setting (\Cref{prop:fundamental-VC}), even when the generator is chosen after observing the queried examples. Making at most $\lceil\log_2(2K/\delta)\rceil$ attempts per boosting round ensures that all $K$ boosting rounds succeed with probability at least $1-\delta/2$. The standard AdaBoost guarantee \citep{freund1997decision} then gives consistency with the full E2E sample. Uniform convergence over weighted majorities of at most $K$ predictors gives population error at most $\eps$, with an additional failure probability of at most $\delta/2$.

\begin{remark}[Extension to stochastic classes]\label{rem:ub-stochastic}
The same approach extends to a finite class $\Pi$ of stochastic next-token policies, each with a deterministic E2E output. Replace $\Cons$ by next-token maximum likelihood. By realizability, the generating policy assigns positive likelihood to every observed trace, so the fitted policy does as well. Its deterministic E2E predictor is therefore consistent with the observed labels. Uniform convergence over the induced E2E class (\Cref{prop:fundamental-VC}) yields a constant-error predictor from $\M=O(\log|\Pi|)$ traces with constant probability, even when the source is chosen after observing the queried examples. Boosting proceeds as before. The per-source CoT budget remains independent of $\eps$. Assuming efficient maximum likelihood and policy evaluation, the algorithm is efficient. Our argument uses only uniform convergence for E2E prediction and neither guarantees nor requires matching the generating source's CoT distribution. More general guarantees for distribution matching \cite{foster2024behavior,sftjoshi2025learning} and stochastic E2E learning with infinite policy classes \cite{doron2026stochastic} are not required.
\end{remark}

\subsection{Learning noisy parities from heterogeneous supervision}\label{subsec:parity-main-paper}

In this subsection, we discuss the problem of \emph{learning noisy parities} from heterogeneous supervision to illustrate some important points concerning our reductions in \Cref{subsec:hard-to-learn-contribution-bias,subsec:hardness-without-filter}.

Our hardness reductions must account for the distinction between learning the contributing source and learning the target. In instance-dependent selection (\Cref{def:instance-dependent-bias}), establishing hardness of learning the routing rule is necessary but not sufficient to prove \Cref{thm:contribution-bias-hardness}. Similarly, under uniform unfiltered mixture supervision (\Cref{def:uniform-mixture}), establishing hardness of recovering source provenance is necessary but not sufficient to prove \Cref{thm:filtering-hardness}. In both cases, a learner may still use the information in the traces to predict the target. This distinction is easy to demonstrate with synthetic concept classes. Here, we demonstrate it explicitly on a natural problem of \emph{learning noisy parities}, which has also been studied in learning with CoT because of its E2E hardness \citep{WLS23,malach2023auto,kim2024transformers,tsilivis2025reinforcement}.

Inputs are drawn uniformly from $\{0,1\}^d$, and each CoT reveals parities of growing subsets of a hidden support of size $k$, with \emph{noise added at each generation step} (see \Cref{sec:parity} for formal details). Under E2E supervision, this gives the standard problem of learning noisy parities and is believed to be computationally hard \cite{blum2003noise}. The formal CoT supervision model appears in \Cref{sec:parity}. When the indices are added in a \emph{fixed unknown order}, the hidden support can be recovered exactly using $O(\log(d/\delta))$ samples and $\poly(d,\log\delta^{-1})$ time; see \Cref{rem:small-samples-suffice-from-single-thinker}. 

We consider heterogeneous supervision in which the order may vary across examples. Each permutation of the hidden support specifies the order in which its indices are added, computing parities of growing subsets, giving $k!$ possible generators.

We allow the contributing generators to be chosen \emph{\textbf{adversarially}} after observing the entire input sample $(x^{(1)},\dots,x^{(m)})$. Thus, the source indices $(j^{(1)},\dots,j^{(m)})$ may depend on the entire sample, allowing more general selection than the instance-dependent rule in \Cref{def:instance-dependent-bias}. We establish the following:
\begin{itemize}
   \item  \textbf{Algorithm under adversarial selection (\Cref{thm:propogating-noise-recovery}):} For any noise rate $\eta\in[0,1/2)$, a simple correlation-based algorithm exactly recovers the hidden support in polynomial time using $O\!\left(k\log(d/\delta)/(1-2\eta)^2\right)$ samples, with probability at least $1-\delta$, even under this adversarial selection of generators. This selection model is more adversarial than instance-dependent selection, since the sources may be chosen after observing the entire input sample, and there is no fixed routing rule to learn. Moreover, a generator may contribute only a single CoT, which is insufficient to learn the target from that trace alone, and filtering is unnecessary for learning the target. The power lies in the intermediate traces, which collectively reveal enough information to learn the target. This demonstrates why our reductions must also control the information revealed by the traces, which is where we use the CoT-leakage-preserving property of \Cref{lem:informal-expressivity-wo-CoT} along with the CoT leakage-hiding strategies in \Cref{subsec:prf-ove-hardness-of-two,subsec:flitering-pf-ovw}.
    \item \textbf{Nearly matching information-theoretic limit (\Cref{thm:it-limit-cot}):} Under the uniform mixture of generators, $\Omega(k\log(d/k)/\log(k+1))$ samples are information-theoretically necessary to recover the hidden support with probability at least $1/2$, even in the noiseless setting $\eta=0$.
\end{itemize}

We also train transformers with gradient descent using the next-token prediction objective on noiseless parity traces.\footnote{Even with $\eta=0$, E2E learning of parities is believed to be hard for gradient-based neural learners, as also observed in practice \cite[e.g.,][]{WLS23,kim2024transformers}. It is also hard for Statistical Query algorithms~\cite{Kea98} when $\min\{k,d-k\}=\omega_d(1)$.} Our experiments reveal a similar qualitative behavior as the theoretical guarantees. We compare training on sequences drawn from a single source, with a fixed order across all examples, against training on sequences drawn uniformly from the $k!$ generators. The goal is to predict the final bit, namely, the parity of the hidden support. Our experiments show that transformers can learn under heterogeneous supervision, with the required sample size and training time growing approximately linearly with $k$.

\bibliographystyle{alpha}  
\bibliography{general}
  
\clearpage
\tableofcontents
\clearpage

\appendix
\crefalias{section}{appendix}
\crefname{appendix}{appendix}{appendices}
\Crefname{appendix}{Appendix}{Appendices}


\newpage
\section{Standard Technical Lemmas}
\begin{lemma}[Bernstein]
\label{lem:bernstein-mean}
Let $X_1,\ldots,X_m$ be independent real-valued random variables with
$\E[X_i]=\mu$ and $|X_i-\mu|\le b$ almost surely. Let
\[
\sigma^2 := \frac{1}{m}\sum_{i=1}^m \Var(X_i).
\]
Then for every $\epsilon>0$,
\[
\Pr\!\left(\left|\frac{1}{m}\sum_{i=1}^m X_i - \mu\right|\ge \epsilon\right)
\;\le\;
2\exp\!\left(-\frac{m\epsilon^2}{2\sigma^2+\frac{2}{3}b\epsilon}\right).
\]
\end{lemma}

Recall the definition of the growth function (\Cref{def:growth-function}). The following classical result provides a sharp upper bound on this quantity in terms of the VC dimension.
\begin{lemma}[Sauer's Lemma]\label{lem:sauer'slemma} For a hypothesis class $\gH \subseteq \{0,1\}^{\cX}$, for every $m \in \naturals_{+}$, we have $ \Gamma_{\gH}(m) \leq (em)^{\VCdim(\gH)}$. Additionally, for $m \geq \VCdim(\gH) \geq 1$:
$$ \Gamma_{\gH}(m) \leq \left( \frac{em}{\VCdim(\gH)
} \right)^{\VCdim(\gH)}.$$  
\end{lemma}
Sauer's Lemma directly links combinatorial complexity to learnability. Building on this connection, the next result characterizes when a hypothesis class is learnable, using the consistent learning rule, in terms of its VC dimension in the realizable case (e.g., \cite[Thm.~6.8]{shalev2014understanding}). For a hypothesis class $\gH \subseteq \cY^{\cX}$ given a training set $S=((\vx_1,y_1),\dots,(\vx_m,y_m)) \in (\gX \times \gY)^m$,
    \begin{equation}\label{eq:consistent}
    \text{Return $\widehat{h}\in \cH $ such that }\,  \widehat{h}(\vx_i) = y_i, \forall i \in [m]. \tag{$\Cons_{\gH}(S)$} 
\end{equation}

\begin{proposition}[The Fundamental Theorem of Statistical Learning]\label{prop:fundamental-VC} There is a universal constant $c>0$ such that for any domain $\cX$, a hypothesis class $\gH \subseteq \cY^{\cX}$ with $\cY=\{0,1\}$ and $\VCdim(\gH)<\infty$, and any distribution $\cD$ over $\cX$ and any $h_* \in \gH$, the following holds: With probability at least $1-\delta$ over $\vx^{(1)}, \dots, \vx^{(m)} \iid \cD$ with $S = (\vx^{(1)},h_*(\vx^{(1)})),\dots,(\vx^{(m)},h_*(\vx^{(m)}))$ for any\ $m\geq m(\eps,\delta)$, we have that for any consistent $\widehat{h}$ (from \ref{eq:consistent})
$$ \cL_{\cD,h_*}(\widehat{h}) \leq  \eps\, \quad \textnormal{ where } \quad m(\eps,\delta) \leq c \left( \frac{\VCdim(\gH) \log(2/\eps)+\log(1/\delta)}{\eps} \right)\,,$$
and $\cL_{\cD,h_*}(h) := \E_{\vx\sim\cD}[\ind[h(\vx) \neq h_*(\vx)]]$.
\end{proposition}

\begin{lemma}[Lemma A.4 from \cite{joshi2025theory}]\label{lem:technical-corollary}
For any $N \in \naturals_{+},M \in \R_{+}$ with $NM \geq 1$, there exists $m \in \naturals_{+}$ such that
$$N \leq m \leq 3 N \log_2 \left( \frac{2NM}{\ln 2} \right) \text{ and } m > N \log_2 (emM) \,.$$
\end{lemma}

\begin{lemma}[Algebraic inequality]\label{lem:algebraic-condition}
For every \(k\geq 1\), there exists a constant \(c>0\), depending only on \(k\), such that for every \(a\geq e\), every \(b\geq 0\), and every \(x>0\), if
\[
x \geq c\left(a\log^k(a)+b\right) \quad \text{ then } \,\, x \geq a\log^k(x)+b .
\]
\end{lemma}

\begin{proof}
Fix \(k\geq 1\). We choose \(c\), depending only on \(k\), sufficiently large. Suppose \(a\geq e\), \(b\geq 0\), and
\( 
x \geq c\left(a\log^k(a)+b\right).
\)
We show that
\[ 
a\log^k(x)\leq \frac{x}{2}
\qquad\text{and}\qquad
b\leq \frac{x}{2}.
\]
The second inequality follows immediately from \(x\geq cb\), provided \(c\geq 2\). 

For the first inequality, let
\( 
r:=\frac{x}{a}
\qquad\text{and}\qquad
u:=\log a .
\) Since \(a\geq e\), we have \(u\geq 1\). Also, from \(x\geq c a\log^k(a)\), we get
\( 
r\geq c u^k .
\) Moreover,
\( 
\log x=\log a+\log r=u+\log r .
\)
We claim that, for \(c\) sufficiently large depending only on \(k\), the conditions \(u\geq 1\) and \(r\geq c u^k\) imply
\[
(u+\log r)^k\leq \frac r2 .
\]
Indeed, there are two cases. First, suppose \(\log r\leq u\). Then
\( 
(u+\log r)^k\leq (2u)^k=2^k u^k.
\) Since \(r\geq c u^k\), this is at most \(r/2\), provided
\( 
c\geq 2^{k+1}.
\) Second, suppose \(\log r>u\). Then
\( 
(u+\log r)^k\leq (2\log r)^k.
\)
Since \((2\log r)^k=o_r(r)\), there exists a constant \(R_k\), depending only on \(k\), such that for all \(r\geq R_k\),
\[
(2\log r)^k\leq \frac r2 .
\]
But \(u\geq 1\), so \(r\geq c u^k\geq c\). Hence \(r\geq R_k\), provided \(c\geq R_k\). Therefore,
\[
(u+\log r)^k\leq \frac r2 .
\]
Thus
\( 
\log^k(x)\leq \frac r2.
\)
Multiplying by \(a\), we obtain
\[ 
a\log^k(x)\leq \frac{ar}{2}=\frac{x}{2}.
\]
Combining this with \(b\leq x/2\), we get
\[
a\log^k(x)+b\leq x.\qed
\]

\end{proof}

\section{Generalization bounds for \texorpdfstring{$\passk$}{} error and E2E prediction}\label{sec:sample-complexity-kappa<<T}

We prove \Cref{thm:satistical_upper_bound} using the learning rule \eqref{eq:kappa-cons-learning-rule} and the loss class $\CoT{\cL}{T}(\cF;\kappa)$ from \eqref{eq:loss-class-kappa} defined in \Cref{subsec:passk-efficient}. 
We first bound the VC dimension of the loss class, then derive the $\mathsf{pass@}\kappa$ guarantee and the E2E guarantee under the additional agreement constraint.

\begin{proof}[Proof of \Cref{thm:satistical_upper_bound}] Given input $\vx \in \cX$, chain-of-thought $z \in \{0,1\}^T$ of length $T$, and $f_1,\dots,f_\kappa \in \gF$, define $\ell_{f_1,\dots,f_\kappa}: \cX \times \{0,1\}^T \rightarrow \{0,1\}$:
\begin{equation}\label{eq:loss-kappa}
    \ell_{f_1,\dots,f_\kappa}(\vx,z) = \ind \l\{\forall i\in[\kappa]
    : \CoT{f_i}{T}(\vx) \neq z \r\}.   
\end{equation}
Recall the binary loss class on the domain $\cX \times \{0,1\}^T$ from \eqref{eq:loss-class-kappa}, which can be written as
\begin{align*}
    \CoT{\cL}{T}(\cF;\kappa) = \{\ell_{f_1,\dots,f_\kappa}: f_1,\dots,f_\kappa \in \gF\}.
\end{align*}
Our goal is to upper bound $\VCdim(\CoT{\cL}{T}(\cF;\kappa))$, which yields, as we will show, an upper bound on the sample complexity of \eqref{eq:kappa-cons-learning-rule}. We start by upper-bounding the growth function of $\CoT{\cL}{T}(\cF;\kappa)$, denoted as $\Gamma_{\CoT{\cL}{T}(\cF;\kappa)}$, in terms of the growth function of $\gF$, denoted as $\Gamma_{\gF}$. Given $S=\left((\vx^{(1)},z^{(1)}),\dots,(\vx^{(m)},z^{(m)}) \right) \in (\cX\times \{0,1\}^T)^m$, we have

\begin{align} 
      |\CoT{\cL}{T}(\cF;\kappa)(S) |&= |\{ (\ell_{f_1,\dots,f_{\kappa}}(\vx^{(1)},z^{(1)}),\dots,\ell_{f_1,\dots,f_{\kappa}}(\vx^{(m)},z^{(m)})): f_1,\dots,f_{\kappa} \in \gF\} | \tag{By \Cref{def:growth-function}}\\
     &\leq | \{ (\ell_{f}(\vx^{(1)},z^{(1)}),\dots,\ell_{f}(\vx^{(m)},z^{(m)})): f \in \gF \} |^{\kappa}, \label{eq:growth_function_upper_bound_exp_kappa}
\end{align}
where the last inequality holds by the following observation: For any $f_1,\dots,f_{\kappa},f_1',\dots,f_{\kappa}' \in \gF$ such that 
\begin{align*}
    (\ell_{f_1,\dots,f_{\kappa}}(\vx^{(1)},z^{(1)}),\dots,\ell_{f_1,\dots,f_{\kappa}}(\vx^{(m)},z^{(m)})) \neq (\ell_{f_1',\dots,f_{\kappa}'}(\vx^{(1)},z^{(1)}),\dots,\ell_{f_1',\dots,f_{\kappa}'}(\vx^{(m)},z^{(m)})),
\end{align*}
there must exist $i\in [m]$ and $j\in[{\kappa}]$ such that $\ell_{f_j}(\vxi,z^{(i)}) \neq \ell_{f_j'}(\vxi,z^{(i)})$. 

Indeed, define
\[
\gU \;=\; \Big\{ \l(\ell_{f_1,\dots,f_{\kappa}}(\vx^{(1)},z^{(1)}),\dots,\ell_{f_1,\dots,f_{\kappa}}(\vx^{(m)},z^{(m)})\r) : f_1,\dots,f_{\kappa} \in \gF \Big\} \subseteq \{0,1\}^m.
\]
For every $u \in \gU$, fix an arbitrary representative $(f_1^u,\dots,f_{\kappa}^u) \in \gF^\kappa$ such that
\[
u = (\ell_{f_1^u,\dots,f_{\kappa}^u}(\vx^{(1)},z^{(1)}),\dots,\ell_{f_1^u,\dots,f_{\kappa}^u}(\vx^{(m)},z^{(m)})).
\]
Define $\phi:\gU \to (\{0,1\}^m)^\kappa$ by
\begin{align*}
\phi(u)
=
\Big(
(\ell_{f_1^u}(\vx^{(1)},z^{(1)}),\dots,\ell_{f_1^u}(\vx^{(m)},z^{(m)})),
\dots,
(\ell_{f_{\kappa}^u}(\vx^{(1)},z^{(1)}),\dots,\ell_{f_{\kappa}^u}(\vx^{(m)},z^{(m)}))
\Big).
\end{align*}
Then $\phi$ is injective, and the size of the range of this mapping can be upper bounded by the RHS of \eqref{eq:growth_function_upper_bound_exp_kappa}, which implies the desired inequality.

Next, consider the indexed sequence of all prefixes of the observed chains of thought in $S$,
\begin{align*}
    \prefix(S)=\left(\vp_{(i,t)}: 1 \leq i \leq m, 0 \leq t \leq T-1\right)
\end{align*}
where $\vp_{(i,t)}= (\vxi, z^{(i)}[:t]) \in \cX \times \{0,1\}^t$. Then, we have 
\begin{align}
&| \{ (\ell_f(\vx^{(1)},z^{(1)}),\dots,\ell_f(\vx^{(m)},z^{(m)})): f \in \gF \} | \nonumber\\
   &\leq |\{ \left( f(\vp_{(1,0)}),\dots,f(\vp_{(m,T-1)})  \right) : f \in \gF \}| \nonumber \\
    & = |\gF(\prefix(S))| \leq \, \Gamma_{\gF}(m T) \,,   &&\text{(As $|\prefix(S)|=mT$)} \label{eq:growth_function_upper_bound}
\end{align}
where the first inequality holds since for any two $f,g \in \gF$ such that
\begin{align*}
(\ell_{f}(\vx^{(1)},z^{(1)}),\dots,\ell_{f}(\vx^{(m)},z^{(m)})) \neq (\ell_g(\vx^{(1)},z^{(1)}),\dots,\ell_g(\vx^{(m)},z^{(m)})),
\end{align*}
there must exist $ 1 \leq i \leq m,\,  0 \leq t \leq T-1$ such that $f(\vp_{(i,t)}) \neq g(\vp_{(i,t)})$. 
Then, using an analogous argument based on fixing canonical representatives as above, define
\[
\cV
=
\Big\{
(\ell_f(\vx^{(1)},z^{(1)}),\dots,\ell_f(\vx^{(m)},z^{(m)})) : f \in \gF
\Big\}.
\]
For every $v \in \cV$, fix a representative $f^v \in \gF$ such that $v = (\ell_{f^v}(\vx^{(1)},z^{(1)}),\dots,\ell_{f^v}(\vx^{(m)},z^{(m)})).$
Define $\psi(v) = \big( f^v(\vp_{(1,0)}),\dots,f^v(\vp_{(m,T-1)}) \big).$
Then $\psi$ is injective, and hence $|\cV|
\;\le\;
|\gF(\prefix(S))|.$
Combining \eqref{eq:growth_function_upper_bound_exp_kappa} and \eqref{eq:growth_function_upper_bound}, we have that $|{\CoT{\cL}{T}(\cF;\kappa)}(S)|\leq \l(\Gamma_{\gF}(m T)\r)^{\kappa}$, and since $S$ is arbitrary, we have that $\Gamma_{\CoT{\cL}{T}(\cF;\kappa)}(m) \leq \l(\Gamma_{\gF}(m T)\r)^{\kappa}$. By Sauer's lemma\footnote{We assume $\VCdim(\gF)>0$; otherwise, the claim follows trivially.} (\Cref{lem:sauer'slemma}), for any $m \geq \VCdim(\gF)$,
\begin{align*}
\Gamma_{\CoT{\cL}{T}(\cF;\kappa)}(m) \leq \l(\Gamma_{\gF}(m T)\r)^{\kappa} \leq \left( \frac{emT}{\VCdim(\gF)} \right)^{{\kappa}\cdot\VCdim(\gF)}.
\end{align*}
Recall that if $\Gamma_{\CoT{\cL}{T}(\cF;\kappa)}(m)<2^m$, then by definition $\VCdim(\CoT{\cL}{T}(\cF;\kappa)) < m$. Therefore, if we can find $m$ such that
\begin{align*}
    {\kappa}\cdot\VCdim(\gF)\cdot \log\left( \frac{emT}{\VCdim(\gF)} \right) < m  \quad \text{and} \quad  \VCdim(\gF) \leq m,
\end{align*}
then $\VCdim(\CoT{\cL}{T}(\cF;\kappa)) < m$. By \Cref{lem:technical-corollary} with $N={\kappa}\cdot\VCdim(\gF)$ and $M=T/\VCdim(\gF)$, there exists a positive integer $m$ such that 
\begin{align*}
    {\kappa}\cdot\VCdim(\gF) \leq m \leq 3 {\kappa} \cdot \VCdim(\gF) \cdot \log \left( \frac{2{\kappa}T}{\ln 2 } \right) \text{ and } m > {\kappa}\cdot\VCdim(\gF) \log \l(\frac {emT} {\VCdim(\gF)}\r).
\end{align*}
We can conclude that
\begin{align}
    \VCdim(\CoT{\cL}{T}(\cF;\kappa)) \leq 3 {\kappa} \cdot \VCdim(\gF) \cdot \log \left( \frac{2{\kappa}T}{\ln 2 } \right) = O\l(  {\kappa} \cdot \VCdim(\gF) \cdot \log \left( 2{\kappa}T \right) \r). \label{eq:vcdim.upper.bound}
\end{align}
 
\paragraph{From VC dimension to CoT agreement.}
Let $f_1^*,\dots,f_\kappa^*$ denote the target generators from \Cref{thm:satistical_upper_bound}.
Let $\cD'$ be the distribution of $(x,z)$ obtained by sampling $x\sim\cD$, then $j\sim\cP(\cdot\mid x)$, and setting $z=\CoT{(f_j^*)}{T}(x)$. 
The observed training pairs are i.i.d. samples from $\cD'$. The target tuple satisfies CoT consistency in \eqref{eq:kappa-cons-learning-rule}, so a feasible tuple exists. When the E2E agreement constraint is imposed, the target tuple also satisfies this constraint because the target thinkers share a common E2E predictor. Moreover, $\ell^*=\ell_{f_1^*,\dots,f_\kappa^*}$ belongs to $\CoT{\cL}{T}(\cF;\kappa)$ and is zero $\cD'$-almost surely.

Apply \Cref{prop:fundamental-VC} to the binary hypothesis class $\CoT{\cL}{T}(\cF;\kappa)$ with target $\ell^*$ and the $\zo$ loss. All observed labels for this auxiliary learning problem are zero. The returned tuple defines $\ellcons=\ell_{\widehat{f}_1,\dots,\widehat{f}_\kappa}$, which is consistent with these labels by \eqref{eq:kappa-cons-learning-rule}. Thus, with probability at least $1-\delta$, its population loss satisfies
\begin{align*}
\E_{(x,z)\sim\cD'}\left[\ind\{\ellcons(x,z)\neq\ell^*(x,z)\}\right]
&=\E_{(x,z)\sim\cD'}[\ellcons(x,z)]\\
&=\Pr_{x\sim\cD,\,j\sim\cP(\cdot\mid x)}\left[\forall i\in[\kappa]:\CoT{\widehat{f}_i}{T}(x)\neq\CoT{(f_j^*)}{T}(x)\right]\leq\eps,
\end{align*}
using a sample size $m\geq m(\eps,\delta)$ with
\begin{align*}
m(\eps,\delta)&\leq c'\left(\frac{\VCdim(\CoT{\cL}{T}(\cF;\kappa))\log(2/\eps)+\log(1/\delta)}{\eps}\right)\\
&\leq c\left(\frac{\kappa\cdot\VCdim(\gF)\cdot\log(2\kappa T)\log(2/\eps)+\log(1/\delta)}{\eps}\right).
\end{align*}
The last inequality follows from \eqref{eq:vcdim.upper.bound}. This proves the claimed $\mathsf{pass@}\kappa$ guarantee with unknown provenance.
Now impose the E2E agreement constraint.
For every $x\in\cX$ and $j\in[\kappa]$, if $\CoT{\widehat{f}_i}{T}(x)=\CoT{(f_j^*)}{T}(x)$ for some $i\in[\kappa]$, then their final tokens agree. Since $\ete{\widehat{f}_i}{T}=\widehat{h}$ and $\ete{(f_j^*)}{T}=h_*$, this implies $\widehat{h}(x)=h_*(x)$. Hence,
\begin{align*}
\Pr_{x\sim\cD}[\widehat{h}(x)\neq h_*(x)]
\leq\Pr_{x\sim\cD,\,j\sim\cP(\cdot\mid x)}\left[\forall i\in[\kappa]:\CoT{\widehat{f}_i}{T}(x)\neq\CoT{(f_j^*)}{T}(x)\right]\leq\eps,
\end{align*}
where the last inequality holds on the $\mathsf{pass@}\kappa$ event established above.
\end{proof}



\begin{remark}
The $\passkappa$ guarantee does not require the target generators to share a common E2E predictor. It suffices to impose only CoT consistency. The target tuple remains feasible and has zero population loss, so the same proof and sample-complexity bound apply.
\end{remark}

\section{Proofs for E2E hardness with short generation length}\label{sec:proof-of-ete-hardness}

In this section, we prove the hardness of learning for short generation length (\Cref{thm:ete-hardness}) and, in \Cref{subsec:prf-of-expressivity-lemma}, the main expressivity lemma associated with it (\Cref{lemma:LT-expressivity-depth-2-d=n'}). We start by describing the class of depth-2 threshold circuits.

\paragraph{Depth-2 threshold circuits.}
Let $\vx = (x_1,\dots,x_n) \in \{0,1\}^n$ denote the input. A depth-$2$ linear threshold circuit $C:\{0,1\}^n \to \{0,1\}$ is defined as a computation over a directed acyclic graph (DAG) $G=(V,E)$ with the following structure:
\begin{itemize}
\item A set of $n$ input nodes $V_{\text{in}}=\{v_{\text{in},1},\dots,v_{\text{in},n}\}$, corresponding to the input bits $(x_1,\dots,x_n)$;
\item A set of $s-1$ hidden nodes $V_{\text{hid}}=\{v_1,\dots,v_{s-1}\}$, each receiving edges only from input nodes;
\item A single output node $v_{s}$, which receives edges only from the hidden nodes;
\item A weight assignment $w:E \to \R$ on the edges.
\end{itemize}

Each node $v \in V$ is assigned a Boolean value $\val(v) \in \{0,1\}$ computed as follows. For input nodes, $\val(v_{\text{in},k}) = x_k$ for all $k \in [n]$. Each other node $v_i \in V_{\text{hid}} \cup \{v_s\}$ computes a threshold function with bias $\theta_i \in \R$:
\begin{align*}\label{eq:value-of-bounded-depth-circuit}
    \val(v_i) = \th\left(\theta_i+\sum_{(u,v_i)\in E} w(u,v_i)  \,\val(u)\right) = \ind \l[ \theta_i+\sum_{(u,v_i)\in E} w(u,v_i)\,\val(u) \geq 0\r].
\end{align*}
The circuit output is given by $C(\vx) = \val(v_s)$. We define $C_i:\{0,1\}^n\rightarrow\{0,1\}$ to be the value of the $i$-th gate, that is $C_i(\vx) = \val(v_i)$. The \emph{size} of the circuit is the number of non-input nodes, i.e., $|V \setminus V_{\text{in}}|=s$. This class captures several well-known function classes as special cases, including DNFs, CNFs, and intersections of halfspaces. In particular, a DNF with $s$ terms, a CNF with $s$ clauses and an intersection of $s$ halfspaces, can be represented by a depth-$2$ threshold circuit of size $s+1$.  

\begin{proof}[Proof of \Cref{thm:ete-hardness}] 
Fix some non-decreasing $g(n)$ such that $ g(n) \leq \poly(n)$ and it is computable in time $\poly(n)$.
Let $C:\{0,1\}^n \to \{0,1\}$ be a target function from the hard class of depth-$2$ threshold circuits of size $g(n)$.
Then by \Cref{lemma:LT-expressivity-depth-2-d=n'}, there exists a $\poly(n)$-time computable mapping
$\phi:\{0,1\}^n \to \{0,1\}^d$, which depends only on $n$ and $g(n)$
, and a vector $\vw \in \R^d$, such that $\ete{f_{\vw}}{T}(\phi(\vx)) = C(\vx)$ for all $\vx \in \{0,1\}^n$,
where $d=d(n)=n(2g(n)-1)+4g(n)-3=O(n\cdot g(n))$ and $T=T(n)=2g(n)-1\leq 2g(n)$.
Since $g$ is non-decreasing and $d\geq n$, we also have that $T\leq 2g(n)\leq 2g(d)$.
Suppose toward contradiction that there exists a $\poly(d)$-time algorithm $\gA$ that weakly learns
$\ete{\cFlin}{T(d)}$ under E2E supervision for all $T(d)\leq2g(d)$.
We construct a $\poly(n)$-time weak learner for the original class as follows.
Given labeled samples $(\vx^{(i)}, y^{(i)})$, we transform them into $(\phi(\vx^{(i)}), y^{(i)})$.
Since $\phi$ is computable in $\poly(n)$ time and $d=\poly(n)$, this transformation is efficient.
By construction in \Cref{lemma:LT-expressivity-depth-2-d=n'}, the target satisfies
$C(\vx) = \ete{f_{\vw}}{T}(\phi(\vx))$ for some $T \leq 2g(d)$, so $\gA$ can be applied to weakly learn it in time
$\poly(d)=\poly(n \cdot g(n))$, which is also $\poly(n)$.
On a new input $\vx$, we apply the same feature map and evaluate the predictor returned by $\gA$ on $\phi(\vx)$.
For every input distribution, this preserves the prediction error and the learner's success probability.
Evaluating the resulting predictor takes polynomial time, since both $\phi$ and the predictor returned by $\gA$ are
computable in polynomial time.
This yields a $\poly(n)$-time weak learner for the class of depth-$2$ threshold circuits of size $g(n)$, a contradiction.
\end{proof}

\paragraph{Proof of \Cref{cor:ete-hardness-cor-1}.}
The first part follows from the construction of \cite{klivans2009cryptographic} after artificially padding zeros to the input, together with the stronger subexponential-time hardness assumption (\Cref{assump:lattice}). 
This will be directly formalized when proving \Cref{thm:hardness-with-two-gens} in \Cref{subsec:proof-of-two-hardness}, since the setting of CoT supervision there is more general than that of E2E supervision. 
The second and third parts follow, respectively, from \cite{daniely2016complexity} and \cite{daniely2021local}. We note that, under a certain Sparse LPN assumption \cite[Corollary 5]{applebaum2025structured}, we also obtain hardness as in the last part of the corollary for \(g(d)=\omega_d(1)\); refer to \Cref{sec:related-work}.

\subsection{Simulating depth-2 threshold circuits with short generation length}\label{subsec:prf-of-expressivity-lemma}
We now provide the proof of our expressivity lemma (\Cref{lemma:LT-expressivity-depth-2-d=n'}).
\begin{proof}[Proof of \Cref{lemma:LT-expressivity-depth-2-d=n'}]
If $s=1$, take $\phi(\vx)=(1,\vx[1],\dots,\vx[n])\in\{0,1\}^{n+1}$ and $\vw=(\theta_1,\vw')\in\R^{n+1}$,
where $\theta_1$ is the bias of the sole gate and $\vw'\in\R^n$ contains its input weights, with zeros for absent edges.
Then $\phi$ depends only on $n$, and $\ete{f_{\vw}}{1}(\phi(\vx))=C(\vx)$, as required with $d=n+1$ and $T=1$.
Thus, we can assume that $s\geq2$. For simplicity, we assume that the first layer contains $s \geq 1$ nodes (instead of $s-1$), so the total number of nodes in the circuit is $s+1$. The desired result will follow at the end of the proof after replacing $s$ by $s-1$ in the final bounds on $T,d$. Let $v_1,\dots,v_s,v_{s+1}$ be the nodes of the circuit, where $v_{s+1}$ is the output node. Let $
    \theta_1,\dots,\theta_{s+1}$ be the corresponding biases. We denote all the weights associated with incoming edges for the $\ord{i}$ node as a vector $\vw_i$. Also, w.l.o.g, we assume that each node is connected to all nodes from the previous layer, as we can add the edges with weight $0$, without affecting the output of the circuit. As a result, we have that $\vw_1,\dots,\vw_s \in \R^n$ and $\vw_{s+1} \in \R^{s}$. We now specify our feature map $\phi:\{0,1\}^n \to \{0,1\}^{d}$. For any $\vx = (\vx[1],\dots,\vx[n])\in \{0,1\}^n$, the feature $\phi(\vx) \in \{0,1\}^{d}$ written as a string takes the following form 
 \begin{equation}\label{eq:feature-map}
       \phi(\vx)= (0^{2s},1,0^{2s},\vx[n],0^{2s},\vx[n-1],\dots,0^{2s},\vx[1],0^{2s}),
    \end{equation} 
where $0^{t}$ denotes the vector with $t$ entries, all equal to zero. Clearly, we have that $d = n(2s+1) + 4s+1 $. The purpose of padding zeros between the coordinates of $\vx$ is to allow us to compute the internal nodes’ values consecutively and ensure that $T$ will not depend on $n$, as we will see later. Our goal is to construct a linear predictor that, when applied autoregressively to $\phi(\vx)$, produces the output of each internal node sequentially (with some zeros separated between the internal nodes' values to the output value). More precisely, we will construct a linear threshold $w$ in terms of $w_1,\dots,w_{s+1}$ and $\theta_1,\dots,\theta_{s+1}$ such that
\begin{align}\label{eq:cot.lt.depth.2}
        \CoT{f_w}{T}(\phi(\vx))=[y_{1},y_2,\dots,y_s,0^s,y_{s+1}], 
    \end{align}
where $y_i = C_i(\vx) = \val(v_i)$.
Establishing \eqref{eq:cot.lt.depth.2} gives the trace required by \Cref{lemma:LT-expressivity-depth-2-d=n'}
after replacing $s$ by $s-1$.
The idea is to combine all the weights from the circuit into one linear threshold. For any $i \in [n]$, let $$\vv_i = (\vw_s[i],\vw_{s-1}[i],\dots,\vw_1[i]) \in \R^s$$ be the vector composed of the $\ord{i}$ coordinates of $\vw_s,\dots, \vw_1$. 
Define 
\begin{align*}
\vw := [\vv,\theta_s,\dots,\theta_1, 0^{s+1},\vv_n,0^{s+1},\vv_{n-1},0^{s+1},\dots,\vv_1,\vw_{s+1},0^{s}] \in \R^{d}    
\end{align*}
where $\vv\in \R^{s+1}$ is a vector to be specified later. Therefore, we have that $\vw$ and $\phi(\vx)$ are initially aligned as follows:
\begin{align}
\vw &= [\underbrace{\vv ,\quad \theta_s ,\dots,\theta_2}_{2s \text{ coordinates}}, \ \theta_1,\underbrace{0^{s+1},\vv_n}_{2s+1 \text{ coordinates}},\dots, \underbrace{0^{s+1},\vv_1}_{2s+1 \text{ coordinates}},\quad \underbrace{\vw_{s+1},0^{s}}_{2s \text{ coordinates}} ] \notag \\
\phi(\vx) &=   \hspace{6pt}  [\underbrace{0^{2s}}_{2s \text{ coordinates}} \quad,1,\underbrace{0^{2s},\vx[n]}_{2s+1 \text{ coordinates}},\dots,\underbrace{0^{2s},\vx[1]}_{2s+1 \text{ coordinates}},\quad  \underbrace{0^{2s}}_{2s \text{ coordinates}}].
\end{align}
That is, the $(0^{s+1},\vv_i)$ slice of $\vw$ is aligned with the $(0^{2s},\vx[i])$ slice of $\phi(\vx)$, for any $i\in[n]$. 
Since the last coordinate of $\vv_i$ is $\vw_1[i]$, the dot product of these slices equals $\vw_1[i]\vx[i]$.
Moreover, $1$ from $\phi(\vx)$ is aligned with $\theta_1$, while the other coordinates of $\vw$ aligned with $0$'s. 
We therefore conclude that $\thr(\langle \vw,\phi(\vx) \rangle) = \ind \l[\summ i n \vw_1[-i]\vx[-i] + \theta_1\geq0\r] = y_1$. 
Similarly, at step $t \in \{2,\dots,s\}$, the coordinate $\vx[i]$ is aligned with $\vv_i[-t]=\vw_t[i]$,\footnote{Given $\vx\in\R^n$, we denote that $t^\mathrm{th}$ entry from the end by $\vx[-t]:=\vx[n+1-t]$} and the coordinate $1$ is aligned with $\theta_t$. 
Moreover, since $\phi(\vx)[1:2s]=\phi(\vx)[-2s:]= 0^{2s}$,\footnote{Given $\vx\in \R^n$, we denote by $\vx[-t:]$ the vector of the last $t$ entries of $\vx$, i.e. $\vx[-t:]=(\vx[n+1-t],\dots,\vx[n])$} during these steps, $\vv$ and $\vw_{s+1}$ are aligned with $0^{s+1}$ and $0^s$, respectively. This implies that $f_{\vw}$ generates $y_1,y_2,\dots,y_s$ on $\phi(\vx)$ in the first $s$ steps.

Next, we want to ``wait'' $s$ steps so $y_1,y_2,\dots,y_s$ align perfectly with $\vw_{s+1}$, while during each of those steps the generated bit will be $0$, as in \Cref{eq:cot.lt.depth.2}. The idea is to choose the vector $\vv \in \R^{s+1}$ whose last $s$ coordinates equal $-B$, to be specified later, such that for every step $t \in \{s+1,s+2,\dots,2s\}$, the $1$ in $\phi(\vx)$ is aligned with $-B$ in $\vv$, which on thresholding gives us the output $0$. Formally, we define $\vv$ to be
\begin{equation}\label{eq:u}
    \vv = [\theta_{s+1}, \underbrace{-B, -B,\dots,-B}_{s \text{ coordinates}}],
\end{equation}
where $B=1+\summ i {s+1} \norm{\vw_{i}}_1$ is a real number greater than the total $\ell_1$ norm of the weights in the circuit. Overall, after $s$ steps, we have that $\vw$ and $\phi(\vx),y_1,\dots,y_s$ are aligned as follows: 
\begin{align}
\vw   &= [\theta_{s+1},\underbrace{-B,\dots,-B}_{s \text{ coordinates}}  ,\underbrace{\theta_s,\dots,\theta_1}_{s \text{ coordinates}},\underbrace{0^{s+1},\vv_n}_{2s+1 \text{ coordinates}},\dots, \quad \underbrace{0^{s+1},\vv_1}_{2s+1 \text{ coordinates}},\quad \quad \underbrace{\vw_{s+1},0^{s}}_{2s \text{ coordinates}} ] \notag \\
\phi(\vx),y_1,\dots,y_s &=   \hspace{15pt}  [\quad 0\dots,0 \quad ,1,\underbrace{0^{s}}_{s \text{ coordinates}}, \underbrace{0^{s},\vx[n],0^{s}}_{2s+1 \text{ coordinates}},\dots,\quad \underbrace{0^{s},\vx[1],0^{s}}_{2s+1 \text{ coordinates}}\quad ,\quad  \underbrace{0^{s},y_1,\dots,y_s]}_{2s \text{ coordinates}}.\label{eq:alignment-after-s}
\end{align}
Since in steps $\{s+1,\dots,2s\}$ we have that $1$ from $\phi(\vx)$ is aligned with $-B$, which implies that the output must be $0$ in those steps, by the choice of $B$ and because the biases $\theta_1,\dots,\theta_{s+1}$ only align with zeros. 

For the final $2s+1^\mathrm{th}$ step, we note that all the $x[i]$ are aligned with $0$s because in \Cref{eq:alignment-after-s}, we had that $(0^{s+1},\vv_i)$ slice from $\vw$ is aligned with the $(0^{s},\vx[i],0^s)$ slice from $\phi(\vx)$, for any $i\in[n]$. 
This implies that 
even after shifting the vector $w$ to the right for $s$ locations, we still have that $x[i]$ align with 0.
Therefore, the final output $\ete{f_{\vw}}{2s+1}(\phi(\vx))$ is obtained by thresholding the linear combination
of the weights aligned with the constant $1$ in $\phi(\vx)$ and the generated bits $y_1,\dots,y_s$.
Note that in step $2s+1$, we have that $y_1,\dots,y_s$ is aligned with $\vw_{s+1}$ and $1$ is aligned with $\theta_{s+1}$ because the entire $w$ is further shifted by $s$ locations to the right than in \Cref{eq:alignment-after-s}. Finally, replacing $s$ with $s-1$ in our bounds $d=n(2s+1)+4s+1, T=2s+1$ completes the proof.
\end{proof}

\section{Proof of hardness under instance-dependent contributions}\label{sec:hardness-with-two-thinkers}
In this section, we prove \Cref{thm:contribution-bias-hardness}: the hardness of learning under instance-dependent contribution biases, even with known provenance (\Cref{def:instance-dependent-bias}). 

We refer to \Cref{subsec:prf-ove-hardness-of-two} for an overview of the proof. The key idea is to construct distribution forks and realize the same end-to-end function using two generators, so that the CoT leakage of the first generator is completely trivial on one part of the distribution, while the CoT leakage of the second generator is completely trivial on the complementary part. Under the hardness of lattice problems, this is done by implementing the decryption function of Regev’s \(\gapsvp\)- and \(\sivp\)-based cryptosystem \cite{regev2009lattices} using two depth-\(2\) threshold-circuit representations (\Cref{subsec:decrypting-the-function-in-two-ways-regev}). Under the existence of local PRGs, we prove a lemma in the same spirit (\Cref{subsec:distribution-fork-under-prg}). In \Cref{subsec:proof-of-two-hardness}, we combine these lemmas with our CoT-leakage-preserving expressivity lemma (\Cref{lemma:LT-expressivity-depth-2-d=n'}) to transform such circuits into generators in \(\cFlin\), whose E2E predictors belong to \(\ete{\cFlin}{T}\), under an appropriate mapping of parameters.



\subsection{Implementing decryption functions of Regev's cryptosystem in two ways}\label{subsec:decrypting-the-function-in-two-ways-regev}

As we discussed, the main component in the proof is to implement the decryption functions of the $\gapsvp$- and $\sivp$-based cryptosystems due to Regev in two ways with the desired property. We refer to \Cref{app:primer} for an overview of these cryptosystems. Here, we directly start by specifying the decryption functions of these cryptosystems. The presentation is based on \cite[Section 2.3]{klivans2009cryptographic}.


 \paragraph{Decryption functions of the cryptosystem.} We refer the reader to \Cref{app:primer} for the decryption procedure. Let $n$ be the security parameter. Given a ciphertext $(a,b)\in \Z_p^n \times \Z_p$, the decryption procedure computes
\( 
(b-\langle a,s\rangle) \hspace{-5pt}\mod p,
\)
and outputs \(0\) if this quantity is closer to \(0\) than to \(\lfloor p/2\rfloor\), and outputs \(1\) otherwise. 

We view a ciphertext \((a,b)\in \Z_p^n\times \Z_p\) as a single input \(x\), which can be encoded using \((n+1)\log p\) bits. Klivans and Sherstov \cite[Lemma 4.4]{klivans2009cryptographic} already established that the decryption functions with secret keys $(s_1,\dots,s_n)\in\Z_p^n$, denoted by,  
$d_{s_1,\ldots,s_n}:\bigl(\{0,1\}^{\log p}\bigr)^{n+1}\to\{0,1\}$, can be written as
\begin{equation}\label{eq:decryption-function}
d_{s_1,\ldots,s_n}(x)
=
\NEARMID_p\!\left(
\sum_{i=0}^{\log p-1} 2^i x_{0,i}
-
\sum_{j=1}^n s_j \sum_{i=0}^{\log p-1} 2^i x_{j,i}
\right)\,.
\end{equation}
The predicate $\NEARMID_p$ is defined as follows: for any $a$,
\begin{equation}\label{eq:nea}
    \NEARMID_p(a) \iff \left|b - \left\lfloor \frac{p}{2} \right\rfloor \right|
\le \min\{b,\, p-b\}\,,
\end{equation}
where $b \in \{0,1,\ldots,p-1\}$ is the integer such that $a \equiv b \pmod p$. Also, following the notation in \cite{klivans2009cryptographic}, and for ease of proof verification, we use the indexing $x_{j,i}$ for $0 \leq j \leq n$ and $0 \leq i \leq \log p -1$ for the coordinates of the input $x$ to $d_{s_1,\dots,s_n}$. We define a concept class of all possible decryption functions (when the security parameter is $n$) as follows:
\begin{equation}
    \cC_n=\{d_{s_1,\ldots,s_n}:\bigl(\{0,1\}^{\log p}\bigr)^{n+1}\to\{0,1\} \mid (s_1,\dots,s_n) \in \Z_p^n \}\,.
\end{equation}
We show the following lemma: for every decryption function, there exist two circuits such that, in one of them, the output of every intermediate node is $1$ whenever the final function output is $1$, while in the other, the output of every intermediate node is $0$ whenever the final function output is $0$. Moreover, in the second circuit, exactly one intermediate node evaluates to $1$ whenever the final function output is $1$. This additional property will be useful later in \Cref{sec:unknown.id} when establishing hardness of learning from an unfiltered mixture (\Cref{thm:filtering-hardness}). 
\begin{lemma}[Implementing decryption functions in two ways]\label{lem:two-ways=for-regev}
For any private key $(s_1,\dots,s_n)\in  \Z_p^n$, the decryption function \(
d_{s_1,\ldots,s_n} \in \cC_n
\) can be implemented with two depth-$2$ threshold circuits $C^0,C^1$, each with $n\log p+2$ gates in the intermediate layer, and $\poly(n)$ time computable feature map with input dimension $2 (n+1)^2 \log^2 p$, such that
\begin{itemize}
    \item $C^1_k(x)=1$ for all nodes $k\in [n\log p+2]$ whenever $d_{s_1,\dots,s_n}(x)=1$ (all gates evaluate to 1) and the 2nd layer computes AND of intermediate gates;
    \item $C^0_k(x)=0$ for all nodes $k\in [n\log p+2]$ whenever $d_{s_1,\dots,s_n}(x)=0$ (all gates evaluate to 0) and the 2nd layer computes OR of intermediate gates;
    \item Finally, for every $x$ with $d_{s_1,\dots,s_n}(x)=1$, there is exactly one intermediate-layer node in $C^0$ that evaluates to $1$.
\end{itemize}
Though not needed for our purpose, we note that the weights of the circuits are bounded by $O\bigl((pn\log p)^2\bigr)$.
\end{lemma}

\begin{proof} We will realize each case from the above as an intersection and union of degree-$2$ polynomial threshold functions (PTFs). The intersection was already shown by \cite{klivans2009cryptographic} which we note, and we here also show the same can be done with the union of degree-$2$ PTFs. This directly corresponds to two circuits with the desired properties in dimension
\( \binom{2+(n+1)\log p}{2}\le 2(n+1)^2\log^2 p,
\)
obtained by applying the quadratic feature map to the input in dimension $(n+1)\log p$, and padding zeros to make the dimension exact. Denote
\[
\dsum(x)
\triangleq
\sum_{i=0}^{\log p-1} 2^i x_{0,i}
-
\sum_{j=1}^n \sum_{i=0}^{\log p-1} (2^i s_j \bmod p)\,x_{j,i}.
\]
Thus, $\dsum(x)$ is the original weighted sum
\[
\left(
\sum_{i=0}^{\log p-1} 2^i x_{0,i}
-
\sum_{j=1}^n s_j \sum_{i=0}^{\log p-1} 2^i x_{j,i}
\right)
\]
with the coefficients reduced modulo $p$. Using the definition of $\NEARMID_p$ in \Cref{eq:nea}, we have
\[
d_{s_1,\ldots,s_n}(x)=\NEARMID_p(\dsum(x)).
\]
The integer $\dsum(x)$ ranges between $-(p-1)n\log p$ and $p-1$, a total range of length $<p(n\log p+1)$. As \cite{klivans2009cryptographic} noted, the range can be divided into consecutive intervals on which $d_{s_1,\ldots,s_n}(x)$ is constant (i.e., does not change value within an interval).

These intervals alternate in label, and intervals of the same label repeat with period $p$. Thus, the range intersects at most $n\log p+2$ intervals of each label. Below are two implementations of the same function.
\begin{itemize}
    \item By picking out the at most $n\log p+2$ intervals on which
$d_{s_1,\ldots,s_n}(x)=0$ and AND'ing their negations, we can compute
$d_{s_1,\ldots,s_n}$ exactly. As \cite{klivans2009cryptographic} noted, it remains to note that the negation of an
interval $[a,b]$ can be computed by a degree-$2$ weight-$O((pn\log p)^2)$ PTF
of the form
\(
\left(\dsum(x)-\frac{a+b}{2}\right)^2
>
\left(\frac{b-a}{2}\right)^2.
\)
As a result, the output of each gate in the corresponding resulting circuit will be $1$ whenever the final output is $1$. The circuit is in dimension $N=O((n\log p)^2)$ after applying a quadratic feature map.
    \item By picking out the at most $n\log p+2$ intervals on which
$d_{s_1,\ldots,s_n}(x)=1$ and OR'ing them, we can compute
$d_{s_1,\ldots,s_n}$ exactly. Similarly, the interval $[a,b]$ can be computed by a degree-$2$ weight-$O((pn\log p)^2)$ PTF
of the form
\(
\left(\dsum(x)-\frac{a+b}{2}\right)^2
\leq
\left(\frac{b-a}{2}\right)^2.
\)
As a result, the output of each gate in the corresponding resulting circuit will be $0$ whenever the final output is $0$.
    \item Moreover, in the second construction, whenever $d_{s_1,\ldots,s_n}(x)=1$, the value $\dsum(x)$ belongs to exactly one of the one intervals labeled 1, since these intervals are non-overlapping. Thus, exactly one of the degree-$2$ PTFs computing their membership tests evaluates to $1$, while all the others evaluate to $0$.\qed
\end{itemize}

\end{proof}

\subsection{Creating distribution forks from local PRGs with polynomial stretch}\label{subsec:distribution-fork-under-prg}
We now show a similar property under \Cref{assumption:prg_exists}. The key idea is to use a hard-to-learn function class whose labels are indistinguishable from independent Bernoulli labels and whose functions admit both DNF and CNF representations. The DNF representation was established in \cite{daniely2021local}; here, we show that the same functions also admit CNF representations.

These representations have the desired properties: whenever the final output is $0$, all DNF terms evaluate to $0$, and whenever the final output is $1$, all CNF clauses evaluate to $1$. Moreover, we additionally show that in our DNF construction, exactly one term evaluates to $1$ whenever the final output is $1$. The construction and its structural properties are given in \Cref{subsec:hardness_of_dnf}. Here, we directly apply \Cref{lem:hardness_of_dnf} to obtain the following lemma, which we shall use in the proof of \Cref{thm:contribution-bias-hardness,thm:filtering-hardness}.

\begin{lemma}\label{lem:hardness_of_dnf_adversarial} Assuming the existence of local PRGs with polynomial stretch (\Cref{assumption:prg_exists}), for any constant $c>1$, there is a binary hypothesis class $\scrC=\cup_{n\geq 1}\cC_n $ and a distribution $\gD_n$ over $\{0,1\}^n$ such that for any $\poly(n)$-time algorithm $\gA$:
\begin{enumerate}
\item $\gA$ does not weakly learn $\cC_n$ with respect to $\gD_n$ with sample size at most $n^c$;
\item for any integer-valued $g(n)=\omega_n(1)$ and any $\psi \in \cC_n$, there exist depth-$2$ threshold circuits $C^0,C^1$ of size exactly $g(n)$ for sufficiently large $n$ such that
\begin{align*}
    \Pr_{\vx \sim \gD_n}\big[C^0(\vx)=C^1(\vx)=\psi(\vx)\big]=1,
\end{align*}
and moreover, for every $\vx\in\operatorname{supp}(\gD_n)$:
\begin{itemize}
    \item if $\psi(\vx)=1$, then all gates of $C^1$ evaluate to $1$ on $\vx$, and the 2nd layer computes AND of intermediate gates; 
    \item if $\psi(\vx)=0$, then all gates of $C^0$ evaluate to $0$ on $\vx$, and the 2nd layer computes OR of intermediate gates;
    \item if $\psi(\vx)=1$, exactly one intermediate-layer gate of $C^0$ evaluates to $1$ on $x$.
\end{itemize}
\end{enumerate}
\end{lemma}

\begin{proof}
Fix $c>1$ and apply \Cref{lem:hardness_of_dnf} with exponent $c+1$. This gives a constant \(\klocal\), a class \(\classdnfhard\) of DNF formulas over \(n_0=N\cdot\klocal\) variables (each with at most \(2^{\klocal}\) terms), and a distribution \(\disthard\) such that no $\poly(n_0)$-time learner \(\gA\) can weakly learn \(\classdnfhard\) with respect to \(\disthard\) using at most $n_0^{c+1}$ samples. For every sufficiently large ambient dimension $n$, set $N=\lfloor n/\klocal\rfloor$ and $n_0=N\klocal$. Pad each instance with $n-n_0<\klocal$ zero coordinates and extend each formula to ignore these coordinates; continue to denote the padded class and distribution by $\classdnfhard$ and $\disthard$. Since $n=\Theta(n_0)$ and $n^c\leq n_0^{c+1}$ for sufficiently large $n$, a $\poly(n)$-time learner using at most $n^c$ samples for the padded problem would contradict the above hardness. The finitely many remaining dimensions may be defined arbitrarily.

We let \(\cC_n=\classdnfhard\) and \(\gD_n=\disthard\). Any \(\psi\in\classdnfhard\) has at most \(2^{\klocal}\) terms and can be implemented by a depth-\(2\) threshold circuit of size at most \(2^{\klocal}+1\). Let \(C^0\) be the circuit computing \(\psi\) via its DNF representation, with OR in the second layer. If \(\psi(\vx)=0\), then all terms evaluate to \(0\), and hence all gates in \(C^0\) output \(0\). Moreover, \Cref{lem:hardness_of_dnf} guarantees that whenever \(\psi(\vx)=1\), exactly one DNF term evaluates to \(1\), establishing the third property.

By \Cref{lem:hardness_of_dnf}, \(\psi\) also admits a CNF representation \(\phi\) with at most \(2^{\klocal}\) clauses that agrees with \(\psi\) on \(\disthard\). Let \(C^1\) be the corresponding depth-\(2\) threshold circuit computing \(\phi\), with AND in the second layer. On the support of \(\disthard\), if \(\psi(\vx)=1\), then all clauses evaluate to \(1\), and hence all gates in \(C^1\) output \(1\).

Finally, both \(C^0\) and \(C^1\) have size at most \(2^{\klocal}+1\leq g(n)\) for sufficiently large \(n\), and both agree with \(\psi\) on \(\gD_n=\disthard\). Pad \(C^0\) with constant-$0$ gates and \(C^1\) with constant-$1$ gates to make their sizes exactly $g(n)$. This preserves all three properties, completing the proof.
\end{proof}

\subsubsection{DNF hardness from local PRGs with additional structural properties}\label{subsec:hardness_of_dnf}
Following \cite{daniely2021local}, we derive hardness of learning DNFs from the existence of local PRGs. We establish two additional properties: single-term satisfaction, that is crucial for the cyclic-shift reduction in \Cref{subsec:cyclic-shift-construction}, and a CNF representation on the support of the hard distribution, used to construct distribution forks in \Cref{subsec:distribution-fork-under-prg}. The following lemma provides the detailed construction; in the proof of our main theorem, we use the corollary stated in \Cref{lem:hardness_of_dnf_adversarial}.

\begin{lemma}
\label{lem:hardness_of_dnf}
Under \Cref{assumption:prg_exists}, for any constant $t>1$ there exists a constant $\klocal>1$ and a family $\classdnfhard$
of DNF formulas over $N\cdot\klocal$ variables, each with at most $2^\klocal$ terms,
together with a distribution $\disthard$ over $\{0,1\}^{N\cdot\klocal}$,
such that \textbf{no} $\poly(n)$-time algorithm over the domain $\{0,1\}^{n}$ (where $n=N\cdot\klocal$)
with sample size at most ${n}^{\,t}$, can learn the class $\classdnfhard$ with respect to $\disthard$,
even within constants $\eps<0.5,\delta < 1$, and even if $\klocal$ is provided as input.

Moreover, $\disthard$ and $\classdnfhard$ satisfy the following properties:
\begin{enumerate}

\item \textbf{Single-Term Satisfaction:} For every $\psi\in\classdnfhard$, write
\begin{align*}
    \psi = c_1 \lor c_2 \lor \dots \lor c_\ell \quad \text{with } \ell \leq 2^\klocal.
\end{align*}
Every input in the support of $\disthard$ satisfies at most one term of $\psi$.
In particular, if $\psi(\vx)=1$, exactly one term is satisfied.
\item \textbf{CNF representation:} For any $\psi \in \classdnfhard$, there exists a CNF formula $\phi$, with at most $2^{\klocal}$ clauses, such that for any $\vx \sim \disthard$, we have that $\psi(\vx) = \phi(\vx)$.
\end{enumerate}
\end{lemma}
Following \cite{daniely2021local}, we encode a hyperedge $S = (i_1,\ldots,i_\klocal)$ by $\vz^S \in \{0,1\}^{N\cdot\klocal}$,
where $\vz^S$ is the concatenation of $\klocal$ vectors in $\{0,1\}^N$,
such that the $j$-th vector has $0$ in the $i_j$-th component and $1$ elsewhere.
Thus, $\vz^S$ consists of $\klocal$ size-$N$ slices, each encoding a member of $S$.
For $\vz \in \{0,1\}^{N\cdot\klocal}$, $i \in [\klocal]$ and $j \in [N]$, we denote $z_{i,j}=z_{(i-1) \cdot N + j}$. That is, $z_{i,j}$ is the $j$-th component in the $i$-th slice in $\vz$.
For a predicate $P:\{0,1\}^\klocal \rightarrow \{0,1\}$ and $\vx \in \{0,1\}^N$, let $P_\vx:\{0,1\}^{N\cdot\klocal} \rightarrow \{0,1\}$ be a function such that for every hyperedge $S$ we have $P_\vx(\vz^S) = P(\vx_{S})$.
\begin{lemma}
\label{lemma:from P to DNF}
For every predicate $P:\{0,1\}^\klocal \rightarrow \{0,1\}$ and $\vx \in \{0,1\}^N$, there is a DNF formula $\psi_{\vx}$ over $\{0,1\}^{N \cdot \klocal}$ with at most $2^\klocal$ terms, such that for every hyperedge $S$ we have $P_\vx(\vz^S)=\psi_\vx(\vz^S)$. Moreover, for any assignment $\vz^S$, at most one term of $\psi_{\vx}$ is satisfied. Furthermore, there exists a CNF formula $\phi_{\vx}$  over $\{0,1\}^{N \cdot \klocal}$ with at most $2^\klocal$ clauses such that $\phi_{\vx}(\vz^S)=\psi_{\vx}(\vz^S)$, for any assignment $\vz^S$.
\end{lemma}
\begin{proof}
The formula $\psi_\vx$ is such that for each satisfying assignment $\vb \in \{0,1\}^\klocal$ of $P$ there is exactly one term in $\psi_\vx$ that checks whether $\vx_S = \vb$.
We denote by $\gB \subseteq \{0,1\}^\klocal$ the set of satisfying assignments of $P$. Note that the size of $\gB$ is at most $2^\klocal$.
Consider the following DNF formula over $\{0,1\}^{N \cdot \klocal}$:
\begin{align*}
\psi_\vx(\vz)
= \bigvee_{\vb \in \gB} c_\vb, \ \ \ \text{where} \ \  \ c_\vb = \bigwedge_{j \in [\klocal]} \bigwedge_{\{l:x_l \neq b_j \}} z_{j,l}~.
\end{align*}
We show that $c_\vb(\vz^S) = 1 \iff \vx_S=\vb$, for any $\vb \in \gB$.
For a hyperedge $S=(i_1,\ldots,i_\klocal)$ and $\vb \in \gB$, we have
\begin{align}
c_\vb(\vz^S) = 1
&\iff \forall j \in [\klocal], \; \forall x_l \neq b_j, \; z^S_{j,l}=1 \notag
\\
&\iff \forall j \in [\klocal], \; \forall x_l \neq b_j, \; i_j \neq l \notag
\\
&\iff \forall j \in [\klocal], \; x_{i_j} = b_j \notag
\\
&\iff \; \vx_S = \vb~. \label{eq:dnf_xs_equals_b}
\end{align}
Therefore,
\begin{align*}
    \psi_\vx(\vz^S) = 1 &\iff \exists \vb \in \gB, c_\vb(\vz^S) = 1 \iff \exists \vb \in \gB, \vx_S=\vb \iff P(\vx_S)=1 \\
    &\iff P_\vx(\vz^S)=1.
\end{align*}
Moreover, for any assignment $\vz^S$, at most one term of $\psi_{\vx}$ is satisfied. Indeed, if $\psi_\vx(\vz^S) = 0$, then clearly $c_\vb(\vz^S) = 0$ for any $\vb\in\gB$. If $\psi_\vx(\vz^S) = 1$, only for $\vb = \vx_S$, we have that $c_\vb(\vz^S) = 1$. This completes the proof of the first part of the lemma.

Similarly, the CNF formula $\phi_{\vx}$ is such that for any unsatisfying assignment $\vg \in \{0,1\}^\klocal$ of $P$
there is exactly one clause in $\phi_\vx$ that checks whether $\vx_S \neq \vg$.
Let $\gG \subseteq \{0,1\}^\klocal$ be the set of \emph{unsatisfying} assignments of $P$. Note that the size of $\gG$ is at most $2^\klocal$.
Therefore,
\begin{align*}
    \psi_{\vx}(\vz^S) = 1 \iff \exists \vb \in \gB: \vx_S = \vb \iff \forall \vg \in \gG: \vx_S \neq \vg \iff \phi_{\vx}(\vz^S) = 1.
\end{align*}
The CNF $\phi_\vx$ takes the following form:
\begin{align*}
\phi_\vx(\vz)
= \bigwedge_{\vg \in \gG} c_{\vg}, \ \ \ \text{where} \ \  \ c_\vg = \bigvee_{j \in [\klocal]} \bigvee_{\{l:x_l \neq g_j \}} \neg{z_{j,l}}~.
\end{align*}
By De Morgan and similarly to \eqref{eq:dnf_xs_equals_b}, we have that
$c_\vg(\vz^S) = 1 \iff \vx_S \neq \vg$. This completes the proof.
\end{proof}
\begin{proof}[Proof of \Cref{lem:hardness_of_dnf}]
Fix a constant $t>1$ and set $s=\lceil t\rceil+2$.
By \Cref{assumption:prg_exists}, there exist a constant $\klocal$ and a predicate
$P:\{0,1\}^{\klocal}\to\{0,1\}$ such that $\cF_{P,N,M}$, with $M=N^s$, is a $\frac13$-PRG.
In particular, $\klocal$ depends only on $t$.
Set $n=N\klocal$ and let $\classdnfhard=\{\psi_\vx:\vx\in\{0,1\}^N\}$,
where $\psi_\vx$ is the DNF from \Cref{lemma:from P to DNF}.
Let $\disthard$ be the distribution of $\vz^S$ when $S$ is a uniformly random ordered $\klocal$-tuple of distinct vertices.
The single-term satisfaction and CNF representation properties follow from \Cref{lemma:from P to DNF}.

Suppose, toward a contradiction, that a $\poly(n)$-time algorithm $\gL$ learns $\classdnfhard$ under $\disthard$
using at most $m(n)=\lfloor n^t\rfloor$ examples, even when given $\klocal$.
Let $\eps<1/2$ and $\delta<1$ be constants such that, for every target in $\classdnfhard$,
the learner returns a hypothesis with error at most $\eps$ with probability at least $1-\delta$.

\paragraph{The distinguisher.}
Given a PRG challenge $(S_1,y_1),\dots,(S_M,y_M)$, form
$\gS=((\vz^{S_1},y_1),\dots,(\vz^{S_M},y_M))$.
Since $\klocal$ is constant and $s>t+1$, for all sufficiently large $N$,
\begin{align*}
    N m(n)+N\leq N^s=M.
\end{align*}
Run $\gL$ independently $N$ times, using disjoint blocks of $m(n)$ examples and fresh internal randomness.
Denote the returned hypotheses by $h_1,\dots,h_N$.
Reserve the next $N$ examples for validation, with index set
$I=\{N m(n)+1,\dots,N m(n)+N\}$.
For each returned hypothesis, compute
\begin{align*}
    \ell_I(h_j)=\frac{1}{N}\sum_{i\in I}\ind\bigl[h_j(\vz^{S_i})\neq y_i\bigr].
\end{align*}
The distinguisher $\gA$ returns $1$ if $\ell_I(h_j)\leq(\eps+1/2)/2$ for some $j\in[N]$,
and returns $0$ otherwise.
It runs in polynomial time, including evaluating all hypotheses on the validation examples.

\paragraph{Pseudorandom labels.}
Condition on the seed $\vx$.
By \Cref{lemma:from P to DNF}, the labels satisfy $y_i=P(\vx_{S_i})=\psi_\vx(\vz^{S_i})$.
Thus, each training block consists of i.i.d.\ examples from the target $\psi_\vx$ under $\disthard$.
Conditional on $\vx$, the training blocks and learner randomness are independent.
Each run succeeds with probability at least $1-\delta$, so the probability that all $N$ runs fail is at most $\delta^N$.

Condition further on the training blocks and learner randomness.
If a successful run exists, fix its first returned hypothesis $h_j$.
The validation examples remain independent of the training blocks, and $h_j$ has error at most $\eps$.
Hoeffding's inequality gives
\begin{align*}
    \Pr\left[\ell_I(h_j)>\frac{\eps+1/2}{2}\right]
    \leq
    \exp\left(-\frac{N(1/2-\eps)^2}{2}\right).
\end{align*}
Consequently, averaging over the seed, training blocks, and learner randomness,
\begin{align*}
    \Pr[\gA\text{ returns }1\mid\text{pseudorandom labels}]
    \geq
    1-\delta^N-\exp\left(-\frac{N(1/2-\eps)^2}{2}\right).
\end{align*}

\paragraph{Uniform labels.}
Condition on all training blocks and learner randomness, so that $h_1,\dots,h_N$ are fixed.
The validation labels are independent uniform bits, independent of the hypotheses and validation inputs.
For each fixed $j$, the validation errors are therefore independent Bernoulli random variables with mean $1/2$.
Hoeffding's inequality and a union bound over the $N$ hypotheses give
\begin{align*}
    \Pr[\gA\text{ returns }1\mid\text{uniform labels}]
    \leq
    N\exp\left(-\frac{N(1/2-\eps)^2}{2}\right).
\end{align*}
The distinguishing advantage is therefore at least
\begin{align*}
    1-\delta^N-(N+1)\exp\left(-\frac{N(1/2-\eps)^2}{2}\right)
    >\frac13
\end{align*}
for sufficiently large $N$, contradicting \Cref{assumption:prg_exists}.
\end{proof}

\subsection{Proof of \texorpdfstring{\Cref{thm:hardness-with-two-gens}}{}}\label{subsec:proof-of-two-hardness}
We now prove our theorem by combining the developments in \Cref{lem:two-ways=for-regev,lem:hardness_of_dnf_adversarial}.
\begin{proof}[Proof of \Cref{thm:hardness-with-two-gens}]
We will respectively prove both statements in the theorem under the two assumptions. First, define
\[
\bar g(a):=\min\left\{a,\left\lfloor\nicefrac{g(a)}{2}\right\rfloor\right\}.
\]
The function \(\bar g\) is non-decreasing, computable in \(\poly(a)\) time, and satisfies \(\bar g(a)\le a\). It also satisfies \(\bar g(a)=\omega_a(\log a\log\log a)\) under \Cref{assump:lattice} and \(\bar g(a)=\omega_a(1)\) under \Cref{assumption:prg_exists}.
\paragraph{Under \Cref{assump:lattice}.} We refer to \Cref{app:primer} for the following lemma, which follows from the standard connection between cryptography and learning under \Cref{assump:lattice}. 
\begin{restate}[\Cref{lem:final-decryption-function-hard-lemma}]
    Let \(\cC=\cup_{n \geq 1}\cC_n\) be the concept class from \Cref{eq:decrypt-class-app}, where \(\cC_n\) consists of the decryption functions of Regev's cryptosystem at security parameter \(n\). Assuming the quantum hardness of \(\gapsvp\) or \(\sivp\) against subexponential-time adversaries (\Cref{assump:lattice}), there is no algorithm that runs in time \(2^{o(n)}\) and uses \(2^{o(n)}\) samples that weakly learns the concept class \(\cC\).
\end{restate}
We will show that the existence of a \(\poly(d)\)-time algorithm for learning the class \(\ete{\cFlin}{T(d)}\), for all \(T(d)\le g(d)\), from CoT supervision of \(\kappa=2\) thinkers, where the contributing thinker is selected by an unknown identity function \(\id:\{0,1\}^d\to\{0,1\}\), would imply a \(2^{o(n)}\)-time algorithm for learning the concept class \(\cC_n\), a contradiction.

\noindent\emph{Reduction:} Given samples \((x^{(i)},y^{(i)})_{i=1}^m\) for the problem of learning \(\cC_n\), the reduction constructs samples \((\Tilde{x}^{(i)},j^{(i)},z^{(i)})_{i=1}^m\), where these respectively denote the input instances, identities, and CoTs associated with the instances.
\begin{enumerate}
    \item First, compute \(N=N(n)\) which is the minimum integer such that \(n\log p+3 \leq \bar g(N)\). Define a map \(x\mapsto \psi(x)\) that first applies the quadratic feature map to \(x\), and then pads zeros (say, to the left) artificially inflating the dimension to \(N\) when $N$ is greater than the size of the quadratic feature map. Apply this map to each example \(x^{(i)}\).
    \item Next, let \(s=s(N)=\bar g(N)\), and apply the feature map from \Cref{lemma:LT-expressivity-depth-2-d=n'} to obtain examples \(\Tilde{x}^{(i)}=\phi\circ\psi(x^{(i)})\) for all \(i\in[m]\). Let \(d=d(N)\) denote the resulting dimension.

    \item For each \(i\in[m]\), set \(j^{(i)}=y^{(i)}\), and define \(z^{(i)}\) to be the length-\(T=T(N)=2s(N)-1\) sequence whose bit at the \(t^\mathrm{th}\) location is exactly \(y^{(i)}\) if \(t\in I\), and \(0\) otherwise. Recall from \Cref{lemma:LT-expressivity-depth-2-d=n'} that the set of special indices is
    \( 
    I=\{1,\dots,s(N)-1\}\cup\{2s(N)-1\}.
    \)
    \item Finally, invoke the weak learning algorithm for \(\cFlin\), and use it to make predictions on new instances by applying the same transformation \(\phi\circ\psi(x)\) to a test instance \(x\) for the problem of learning \(\cC_n\).
\end{enumerate}

\noindent \emph{Correctness:} We now argue that the resulting algorithm indeed weakly learns the concept class \(\cC_n\); the runtime analysis is deferred to after this. It suffices to show that whenever the data-generating process \((x^{(i)},y^{(i)})\) is realizable by \(\cC_n\), the transformed data obtained from the above reduction is realizable by the class \(\cFlin\) with \(T\leq g(d)\), under CoT supervision from \(\kappa=2\) thinkers, where the identity of the contributing thinker is selected by some function \(\id:\{0,1\}^d\to\{0,1\}\). This follows directly from \Cref{lem:two-ways=for-regev} and \Cref{lemma:LT-expressivity-depth-2-d=n'}.

Indeed, by \Cref{lem:two-ways=for-regev}, for every decryption function \(h=d_{s_1,\dots,s_n}\in \cC_n\), there exist two circuits with exactly \(n\log p+3\) total gates such that the intermediate nodes have the following property: whenever the final output is \(h(x)=0\), all intermediate nodes in the first circuit evaluate to \(0\); and whenever the final output is \(h(x)=1\), all intermediate nodes in the second circuit evaluate to \(1\). Since \(n\log p+3 \leq \bar g(N)\), we may pad these circuits so as to obtain two depth-2 threshold circuits with the same property and with exactly \(\bar g(N)\) total gates (including the output gate), by adding gates in the intermediate layer that always evaluate to \(0\) or \(1\), respectively.

Applying \Cref{lemma:LT-expressivity-depth-2-d=n'} now yields two thinkers \(f_0,f_1\in\cFlin\) such that whenever \(h(x)=0\), the CoT \(z\) produced by our reduction exactly matches the CoT of \(f_0\), and whenever \(h(x)=1\), it exactly matches the CoT of \(f_1\). Also, note that $T=T(N)=2\bar g(N)-1<g(N)\leq g(d)$, because $d\geq N$ and $g$ is non-decreasing.
 Finally, letting \(\id(x)=h(x)\), we obtain the desired realization. Therefore, any weak learner for the class \(\ete{\cFlin}{T(d)}\) for all $T(d) \leq g(d)$ in this setting would also yield a weak learner for the class \(\cC_n\).

\noindent \emph{Runtime:} In our reduction, an instance with parameter \(n\) is mapped to an instance with size parameter \(d\) through the sequence
\( n \to N(n) \to d(N).\)\footnote{Here, w.l.o.g., we consider the case in which $\bar g$ is such that $N(n)$ is larger than the size of the quadratic feature map. So size of $\psi(x)$ is $N(n)$. Otherwise, the proof can be continued by taking $N(n)=O(n^2 \log^2 n)$, i.e., the size of the quadratic feature map.} We first argue that as long as the choice of \(N(n)\) in the first step of the reduction satisfies \(N(n)=2^{o(n)}\), the overall algorithm runs in time \(2^{o(n)}\). Indeed, \(N\) can be found by checking all values in \(\{1,\dots,N\}\) until the defining condition is met, and this takes time at most \(\poly(N)\), since \(\bar g(N)\) is computable in \(\poly(N)\) time. Further, by \Cref{lemma:LT-expressivity-depth-2-d=n'}, the resulting dimension satisfies
\( d(N)=\Theta(N\cdot s(N))=\Theta(N\cdot \bar g(N)).
\) Since, by assumption, \(\bar g(N)\le N\), it follows that
\( 
d(N)=O(N^2).\)
Hence, the learning algorithm runs in time
\( 
\poly(d)=\poly(N).\)
The step of applying feature map $\phi\circ \psi$ is also $\poly(N)$ time by \Cref{lemma:LT-expressivity-depth-2-d=n',lem:two-ways=for-regev}. Therefore, the entire reduction runs in time
\( 
\poly(N)\le N^{O(1)} \le (2^{o(n)})^{O(1)} = 2^{o(n)}.
\)

It remains to show that \(N(n)=2^{o(n)}\). Suppose otherwise. Then there exists \(\epsilon>0\) and infinitely many \(n\) such that \(  N(n)\ge 2^{\epsilon n}.\)
By minimality of \(N(n)\), this means that for every \(a<2^{\epsilon n}\),
\( \
\bar g(a)<n\log p+3=O(n\log n).
\)
In particular, taking \(a=\lfloor 2^{\epsilon n/2}\rfloor\), we obtain
\( 
\bar g(a)= O(n\log n)=O(\log a \log \log a),
\) where we used that \(\log a=\Theta(n)\). This contradicts the assumption \(\bar g(a)=\omega_a(\log a \log \log a)\). Therefore \(N(n)=2^{o(n)}\).

 \noindent\textbf{Under \Cref{assumption:prg_exists}.} The proof follows the same reduction idea, but uses \Cref{lem:hardness_of_dnf_adversarial} with growth function $\bar g$. Let $\cA$ be a learner that learns $\ete{\cFlin}{T(d)}$ for all $T(d)\leq g(d)$ in time $\poly(d)$ under the supervision model of \Cref{thm:hardness-with-two-gens}. Suppose its sample complexity is bounded by $d^{c/3}$ for some $c>1$. We will show that this yields a learner contradicting \Cref{lem:hardness_of_dnf_adversarial} for the same constant $c$.

Apply the same Steps 2, 3, and 4 from the reduction already described, with $s(n)=\bar g(n)$, when given samples for the concept class $\cC=\cup_{n \geq 1}\cC_n$ over the distribution $\cD_n$, for the choice of $c$ under consideration. We will show that using $\cA$ in this way yields a $\poly(n)$-time weak learner for the class $\cC=\cup_{n\geq 1}\cC_n$ over the distribution $\cD_n$, using at most $n^c$ samples.

\noindent\emph{Correctness and Runtime:} We observe that the realizable data distribution $(x,\psi(x))$ for $x\sim\cD_n$ and any unknown $\psi \in \cC_n$, when combined with our reduction applied to each sample, is realizable by the concept class $\ete{\cFlin}{T}$ with supervision of some thinkers $f_{w_0}$ and $f_{w_1}$, i.e., $\kappa=2$, where the identity of the thinker is given by $\id(\phi(x))=\psi(x)$. 

Similar to the analysis under the other assumption, we have $d=d(n)=O(n\cdot \bar g(n))$ and $T=T(n)=2\bar g(n)-1<g(n)\leq g(d)$, since $d \geq n$ and $g$ is non-decreasing. Moreover, by \Cref{lem:hardness_of_dnf_adversarial}, for any $\psi\in \cC_n$, there exist depth-$2$ threshold circuits $C^0,C^1$ of size exactly $\bar g(n)$ such that $C^0(\vx) = C^1(\vx) = \psi(\vx)$ with probability $1$ over $\vx\sim\gD_n$. Moreover, if $\psi(\vx)=0$, then all gates of $C^0$ evaluate to $0$ on $\vx$, and if $\psi(\vx)=1$, then all gates of $C^1$ evaluate to $1$ on $\vx$. By \Cref{lemma:LT-expressivity-depth-2-d=n'}, this implies that there are two linear thresholds $\vw_0,\vw_1 \in \R^d$ such that $\ete{f_{\vw_0}}{T}(\phi(\vx))=C^0(\vx)$ and $\ete{f_{\vw_1}}{T}(\phi(\vx))=C^1(\vx)$ for all $\vx \in \{0,1\}^n$, and the respective CoTs generated in our reduction are exactly as desired. Finally, letting $\id(\phi(x))=\psi(x)$ specifies which thinker's CoT matches. Therefore, if $\cA$ is a weak learner for our learning problem, then the algorithm obtained after this reduction is also a weak learner for the problem of learning $\cC=\cup_{n \geq 1} \cC_n$ with respect to $\cD_n$.

Now, we have $d=O(n\cdot \bar g(n))=O(n^2)$, recalling that $\bar g(a)\leq a$. Since $\cA$ runs in time $\poly(d)=\poly(n)$, and the other steps of the reduction also run in $\poly(n)$ time (using \Cref{lemma:LT-expressivity-depth-2-d=n'}), the overall runtime is $\poly(n)$. The number of samples is bounded by $d^{c/3}=O(n^{2c/3})\ll n^c$, contradicting \Cref{lem:hardness_of_dnf_adversarial}. This completes the proof under both assumptions.
\end{proof}

\begin{remark}\label{rem:tiegel-conclusion}
Our reduction in Step 1, used under \Cref{assump:lattice}, formalizes Tiegel's zero-padding argument \cite{tiegel2024improved} and the implied relationship between the number of halfspaces and the input dimension. However, we need $g(N)=\omega_N(\log N\log\log N)$ to ensure that the overall runtime remains $2^{o(n)}$, whereas \cite[Section 2]{tiegel2024improved} states that $g(N)=\omega_N(\log N)$ suffices. The discrepancy arises from distinguishing the security parameter $n$ from the ciphertext length, which is $O(n\log p)=O(n\log n)$. The assumption concerns subexponential-time hardness in $n$, rather than in the ciphertext length. Indeed, $2^{O(n)}$-time algorithms are known \cite[e.g.,][]{regev2009lattices}. We have confirmed this parameter accounting and the resulting requirement on $g(N)$ with Stefan Tiegel in private communication \cite{tiegel_private_communication}.
\end{remark}

\section{Proof of hardness of learning from unfiltered uniform mixture}\label{sec:unknown.id}
In this section, we prove \Cref{thm:filtering-hardness}, which establishes the hardness of learning from an unfiltered uniform mixture of CoTs from multiple sources (\Cref{def:uniform-mixture}). We build on the structural property established in \Cref{lem:two-ways=for-regev,lem:hardness_of_dnf_adversarial} in the previous subsection. Combining this property with a common cyclic shift of circuit gates, we construct a collection of generators whose uniform mixture of CoTs can be efficiently sampled using only E2E labels (\Cref{subsec:cyclic-shift-construction}). We then combine these ingredients to prove the theorem in \Cref{subsec:proof-of-filtering hardness}.

\subsection{A cyclic-shift strategy to hide CoT leakage}\label{subsec:cyclic-shift-construction}

Our reduction exploits a structural property shared by the hard circuit classes considered above. We give a common reduction for depth-$2$ threshold circuits with an OR output gate, requiring that, with probability $1$ over the input distribution, exactly one hidden gate outputs $1$ whenever the final output is $1$. Both the hard circuit classes with the OR representation in \Cref{lem:two-ways=for-regev} and \Cref{lem:hardness_of_dnf_adversarial} satisfy this condition.

\begin{lemma}\label{lem:cyclic-shift-mixture}
Let $C$ be a depth-$2$ threshold circuit with $s\geq2$ gates on $\{0,1\}^n$ and an OR output gate.
Let $\cD$ be an input distribution such that, with probability $1$ over $\vx\sim\cD$,
at most one hidden gate outputs $1$.
Let $\phi:\{0,1\}^n\to\{0,1\}^d$ be the feature map from \Cref{lemma:LT-expressivity-depth-2-d=n'},
where $d=n(2s-1)+4s-3$.
There exist $\kappa=s-1$ generators $f_1,\dots,f_\kappa\in\cFlin$ that all compute $C$ end-to-end on embedded inputs in $T=2s-1$ steps; that is, $\ete{f_j}{T}(\phi(\vx))=C(\vx)$ for every $\vx\in\{0,1\}^n$ and every $j\in[\kappa]$.
Given $(\vx,C(\vx))$, $n$, and $s$, a randomized algorithm running in $\poly(n,s)$ time outputs $(\phi(\vx),z)$ such that, when $\vx\sim\cD$,
\[
(\phi(\vx),z)
\overset{\textnormal{dist}}{=}
(\phi(\vx),\CoT{f_j}{T}(\phi(\vx))),
\]
where $j\sim\Unif[\kappa]$ is independent of $\vx$.
The algorithm does not require a description of $C$.
\end{lemma}

\begin{proof} Let $C_1,\dots,C_{s-1}$ denote the hidden-gate functions of $C$. For each $r\in\{0,\dots,s-2\}$, cyclically shift the hidden gates of $C$ by $r$ positions, keeping the output gate unchanged.
Since OR is invariant under permutations, each resulting circuit computes $C$ on every input.
Apply \Cref{lemma:LT-expressivity-depth-2-d=n'} to these circuits.
The feature map $\phi$ is common to all of them, since it depends only on $(n,s)$.
By \eqref{eq:CoT-computation}, the corresponding generator $f_{r+1}$ satisfies
\begin{align*}
    \CoT{f_{r+1}}{T}(\phi(\vx))
    =
    [C_{r+1}(\vx),\dots,C_{s-1}(\vx),C_1(\vx),\dots,C_r(\vx),0^{s-1},C(\vx)].
\end{align*}
For $\cD$-almost every $\vx$, the hidden-gate output vector is all zeros if $C(\vx)=0$ and has exactly one $1$ if $C(\vx)=1$. In the latter case, the cyclic shifts place this $1$ in each of the $s-1$ positions exactly once. Thus, a uniformly chosen shift produces either the all-zero vector or a uniformly random standard basis vector, according to the label $C(\vx)$.

The sampler constructs this vector from $C(\vx)$, appends $s-1$ zeros and the label to obtain $z$, and returns $(\phi(\vx),z)$. For $\cD$-almost every $\vx$, the conditional distribution of $z$ is exactly that of $\CoT{f_j}{T}(\phi(\vx))$ for $j\sim\Unif[\kappa]$. Hence, when $\vx\sim\cD$, the output has the required joint distribution. The sampler runs in $\poly(n,s)$ time and does not require a description of $C$.
\end{proof}

\begin{remark}
The dependence on $\kappa$ in \Cref{thm:filtering-hardness} arises from our cyclic-shift strategy for hiding trace leakage, which uses one generator per first-layer gate. Finding a strategy that hides this leakage with only $\kappa=2$ generators and establishes analogous hardness, as conjectured in \Cref{sec:discussion-future-work}, remains open.
\end{remark}


\subsection{Proof of \texorpdfstring{\Cref{thm:filtering-hardness}}{}}\label{subsec:proof-of-filtering hardness}

We follow the same technical outline as the proof for \Cref{thm:contribution-bias-hardness} in \Cref{subsec:proof-of-two-hardness}, with one conceptual change: we use \Cref{lem:cyclic-shift-mixture} to simulate unfiltered uniform-mixture supervision from end-to-end labeled examples.

\begin{proof}[Proof of \Cref{thm:filtering-hardness}]
We prove the two statements separately, following the reductions in \Cref{subsec:proof-of-two-hardness}. Again, define
\( 
\bar g(a):=\min\left\{a,\left\lfloor\nicefrac{g(a)}{2}\right\rfloor\right\}.
\)
 Under the lattice assumption, it satisfies $\bar g(a)=\omega_a(\log a\log\log a)$; under the local PRG assumption, it satisfies $\bar g(a)=\omega_a(1)$.

\paragraph{Under \Cref{assump:lattice}.}
Suppose that $\cA$ is a $\poly(d)$-time weak learner in the setting of the theorem. We show that $\cA$ yields a $2^{o(n)}$-time weak learner for the Regev decryption class $\cC_n$ from \Cref{lem:final-decryption-function-hard-lemma}, a contradiction.

\noindent\emph{Reduction:}
Let $q(n)=O(n\log n)$ bound the size of the OR-output circuit representation from \Cref{lem:two-ways=for-regev}, and let $m(n)=\poly(n)$ be the dimension of the quadratic feature map used in \Cref{subsec:proof-of-two-hardness}. Given labeled examples $(x^{(i)},y^{(i)})$, construct examples $(\tilde x^{(i)},z^{(i)})$ as follows:
\begin{enumerate}
    \item Compute
    \[
    N=\min\{a\geq m(n):\bar g(a)\geq q(n)\}.
    \]
    Let $\psi$ apply the quadratic feature map and then pad with zeros to dimension $N$.

    \item Set $s=\bar g(N)$ and apply the feature map $\phi$ from \Cref{lemma:LT-expressivity-depth-2-d=n'} to obtain
    \[
    \tilde x^{(i)}=\phi(\psi(x^{(i)})).
    \]
    The resulting dimension and generation length are
    \[
    d=N(2s-1)+4s-3,
    \qquad
    T=2s-1.
    \]

    \item Construct a vector $b^{(i)}\in\{0,1\}^{s-1}$ from the label. If $y^{(i)}=0$, set $b^{(i)}=0^{s-1}$. Otherwise, place a single $1$ at a uniformly random position, independently for each example. Set
    \[
    z^{(i)}=[b^{(i)},0^{s-1},y^{(i)}].
    \]

    \item Invoke $\cA$ on the examples $(\tilde x^{(i)},z^{(i)})$. To predict on a new input $x$, apply the returned predictor to $\phi(\psi(x))$.
\end{enumerate}

\noindent\emph{Correctness:}
Fix a target $h\in\cC_n$. By \Cref{lem:two-ways=for-regev}, after the quadratic feature map and padding, $h$ is represented by a depth-$2$ threshold circuit $C$ with an OR output gate and at most one active hidden gate. Since $s\geq q(n)$, we may add all-zero hidden gates to obtain exactly $s$ gates, preserving this property and ensuring
\[
C(\psi(x))=h(x).
\]

By \Cref{lem:cyclic-shift-mixture}, there are $\kappa=s-1$ generators in $\cFlin$ that all compute $C$ end-to-end on embedded inputs. Moreover, Step~3 produces exactly their uniform-mixture trace distribution, with the source chosen uniformly and independently of the input. Thus, the transformed examples are realizable under the unfiltered uniform-mixture model (\Cref{def:uniform-mixture}).

Since $d\geq N$ and $g$ is non-decreasing,
\[
T=2s-1\leq g(N)\leq g(d),
\qquad
\kappa=s-1\leq T\leq g(d).
\]
Hence, $\cA$ applies to these examples. Composing its predictor with $\phi\circ\psi$ therefore yields a weak learner for $\cC_n$.

\noindent\emph{Runtime:}
The same minimal-padding argument as in \Cref{subsec:proof-of-two-hardness}, applied to $\bar g$, gives $N=2^{o(n)}$, since
\( \bar g(a)=\omega_a(\log a\log\log a).
\)
Furthermore,
\[
d=O(Ns)=O(N\bar g(N))=O(N^2) \text{ since } \bar{g} \leq a.
\]
Computing $N$, applying the feature maps, and simulating each trace take $\poly(N)$ time. Consequently, the resulting learner runs in $2^{o(n)}$ time and uses $2^{o(n)}$ samples, contradicting \Cref{lem:final-decryption-function-hard-lemma}.

\paragraph{Under \Cref{assumption:prg_exists}.}
The proof follows the same reduction idea, using \Cref{lem:hardness_of_dnf_adversarial}. Let $\cA$ be a $\poly(d)$-time weak learner in the setting of the theorem, and choose $c>1$ large enough that its sample complexity is at most $d^{c/3}$. We show that this yields a learner contradicting \Cref{lem:hardness_of_dnf_adversarial} for the same constant $c$.

Apply Steps~2--4 of the reduction above with $N=n$, $\psi$ equal to the identity map, and $s(n)=\bar g(n)$, to samples from the hard class $\cC=\bigcup_{n\geq1}\cC_n$ over $\cD_n$ supplied by \Cref{lem:hardness_of_dnf_adversarial}. We show that the resulting algorithm weakly learns $\cC$ in $\poly(n)$ time using at most $n^c$ samples.

\noindent\emph{Correctness and runtime:}
By \Cref{lem:hardness_of_dnf_adversarial}, each target $h\in\cC_n$ admits an OR-output depth-$2$ threshold representation $C$ with $s=s(n)$ gates such that $C(x)=h(x)$ and at most one hidden gate is active, almost surely over $x\sim\cD_n$. By \Cref{lem:cyclic-shift-mixture}, there are $\kappa=s-1$ generators in $\cFlin$ that all compute $C$ end-to-end on embedded inputs. Moreover, the traces constructed in Step~3 have exactly their uniform-mixture distribution, with the source chosen uniformly and independently of the input.

The resulting parameters satisfy
\[
d=n(2s-1)+4s-3,
\qquad
\kappa=s-1\leq T=2s-1\leq g(n)\leq g(d).
\]
Thus, the transformed examples are realizable in the supervision model of the theorem, and composing the predictor returned by $\cA$ with $\phi$ yields a weak learner for $\cC_n$ over $\cD_n$.

Finally, since $\bar g(n)\leq n$, we have
\( 
d=O(n\bar g(n))=O(n^2).
\)
As in \Cref{subsec:proof-of-two-hardness}, the reduction and learner therefore run in $\poly(n)$ time. The number of original labeled examples used is at most
\[
d^{c/3}=O(n^{2c/3})<n^c
\]
for sufficiently large $n$, contradicting \Cref{lem:hardness_of_dnf_adversarial}.
\end{proof}

\section{Proof of \texorpdfstring{\Cref{thm:boosting-model-aggregation}}{}: analyzing \texorpdfstring{\Cref{alg:adaboost-cot}}{}}\label{sec:analysis-of-alogorithm}
In this section, we prove our main algorithmic result in \Cref{thm:boosting-model-aggregation}.
\paragraph{Overview of the steps in the analysis.} We will keep \Cref{alg:adaboost-cot} in mind for the proof. We begin with an overview, dividing the argument into three parts, as is standard in boosting analyses.
\begin{itemize}
\item First, we show that the inner for loop ensures that the predictor $\widehat f_k$ has error $\epsilon_k \leq 0.25$ simultaneously for all $k$, with high probability. We establish this for the rule in \eqref{eq:CoT-Conc} trained on a small sample drawn from $D^{(k)}$, even when the generator was chosen adaptively.
\item Second, we show that if, for each iteration $k \in [K]$ of the outer loop in \Cref{alg:adaboost-cot}, we obtain a predictor with error $\epsilon_k \leq 0.25$ on the distribution $D^{(k)}$ constructed by the algorithm, then the final predictor $\widehat h_K$ returned has zero empirical error on the training set $S_{\sete}$.
\item Finally, to establish \Cref{thm:boosting-model-aggregation}, we will show that the final predictor $\widehat h_K$ is a zero-error predictor on a large training set of size $m$ by combining only $K=\lceil8\log(2m)\rceil$ predictors from the small-capacity hypothesis class $\ete{\cF}{T}$.
\end{itemize}

The first step of the analysis establishes a distribution-free guarantee that holds even when the CoTs are provided by a single generator after observing the E2E training set. This will be important for proving guarantees under our general protocol in \Cref{def:active-adaptive-collection}.
\paragraph{Selection of a single generator after observing E2E data.}
Formally, we prove a guarantee for the setting in which the generator $f_* \in \cF$ is chosen adaptively after observing the E2E training set. We show the following.

\begin{lemma}[Generator chosen after samples]\label{lem:adv-chosen-single-thinker}
There exists a universal constant $c>0$ such that the following holds. For any base class $\cF \subseteq \{0,1\}^{\{0,1\}^*}$, thinking length $T$, domain $\cX\subseteq \{0,1\}^*$, choice of adversary $(\cD,h_*)$, and sample size
$$
m \geq c \left( \frac{\min\{\log |\cF|,\VCdim(\ete{\cF}{T})\log(2/\eps)\}+\log \delta^{-1}}{\eps}\right)\,,
$$
with probability at least $1-\delta$ over the draw of
$S_{\sete}=\{(x^{(i)},h_*(x^{(i)}))\mid i \in [m]\}$,
where $x^{(i)}$ are drawn independently from $\cD$, the following holds simultaneously for every generator $f_*\in \cF$ such that $\ete{f_*}{T} \equiv h_*$ (selected after sampling $S_{\sete}$): a predictor $\widehat{h}$ returned by the rule in \eqref{eq:CoT-Conc} on the training set
$S_{\sCoT}=\{(x^{(i)},z^{(i)})\mid \forall\,{i \in [m]},\,\, z^{(i)}=\CoT{f_*}{T}(x^{(i)})\}\,$
has $L_{\cD,h_*}(\widehat{h}) \leq \eps$.
\end{lemma}

\begin{proof}
This follows from \Cref{prop:UC-of-e2e-CoT}, whose guarantee does not use CoT information. For any $f_*\in \cF$ such that $\ete{f_*}{T} \equiv h_*$, any predictor returned by the rule in \eqref{eq:CoT-Conc} on $S_{\sCoT}$ also satisfies \eqref{eq:ete-consistent}.
\end{proof}
Although, from a purely statistical perspective, the guarantee above may seem to leave the CoT information unused, its main advantage is computational here as we shall see. In particular, the rule in \eqref{eq:CoT-Conc} is efficiently implementable given a tractable $\Cons$ oracle (cf.~\Cref{def:cons-oracle}), whereas learning from E2E labels alone may be computationally hard (cf.~\Cref{thm:ete-hardness}).

\paragraph{Intermediate predictors are weak learners.} The next step of the analysis is to establish that \Cref{alg:adaboost-cot} returns a sequence of weak predictors $(\widehat{f}_1,\dots,\widehat{f}_K)$ such that, with high probability, each satisfies $\epsilon_k \leq 0.25$ for the corresponding distribution in the sequence $D^{(1)},\dots,D^{(K)}$. 

\begin{lemma}[$\widehat f_k$ are weak learners for $D^{(k)}$]
\label{lem:weak-learner-generalization}
Consider any fixed training set $S_{\sete}=\{(x^{(i)},y^{(i)}): i \in [m]\}$. There is a universal constant $c>0$ such that for any $\delta \in (0,1)$, there exists a choice of $\M$ satisfying
\[
\M \leq c\,\VCdim(\ete{\cF}{T})
\]
such that \Cref{alg:adaboost-cot}, invoked with this choice of $\M$, enjoys the following guarantee with probability at least $1-\delta/2$ over its internal randomness of subsampling: for all $k\in[K]$ simultaneously,
\[
\epsilon_k:=\Pr_{x\sim D^{(k)}}\!\big[\ete{\widehat f_k}{T}(x)\neq h_*(x) \big] \;\le\; 0.25.
\]
Moreover, in the relaxed protocol in which $f_{k,\ell}$ is fixed by the adversary before observing the subsample $\cS_{k,\ell}$, the same guarantee holds with $\M\leq c \VCdim(\CoT{\cL}{T}(\cF))=O(\VCdim(\cF)\log T)$ (recall \Cref{eq:loss-class-main} and \Cref{prop:VC-base-class-relation-ship}).
\end{lemma}
\begin{proof} Fix $k\in[K]$ and condition on the history before iteration $k$, so that $D^{(k)}$ is fixed. With probability at least $0.9$ over the randomness over the draw of $\cS_{k,\ell}$ i.i.d. from $D^{(k)}$, any \ref{eq:CoT-Conc} predictor $\widehat f_{k,\ell}$, trained on the CoTs given by a generator $f_{k,\ell}$ has error of disagreement of at most $0.25$ on a new test point $x\sim D^{(k)}$. This follows by applying the guarantee from \Cref{lem:adv-chosen-single-thinker} to constant accuracy $\eps=0.25$ and $\delta=0.1$. The same holds under the relaxed protocol by invoking \Cref{prop:UC-of-e2e-CoT} with $\M\leq c \VCdim(\CoT{\cL}{T}(\cF))=O(\VCdim(\cF)\log T)$ by \Cref{prop:VC-base-class-relation-ship}. Using respective bounds under either protocol, with probability at least $0.9$ over a random draw of subset $\cS_{k,\ell}$ of size $\M$, the predictor consistent with CoT  $\widehat{f}_{k,\ell}$ has 
$$\epsilon_{k,\ell} =\Pr_{x\sim D^{(k)}}\left( \ete{\widehat{f}_{k,\ell}}{T}(x) \neq h_*(x)\right) \leq 0.25\,.$$
At each repetition $\ell$, conditional on all previous randomness, the probability that $\epsilon_{k,\ell}>0.25$ is at most $0.1$. Thus, after $L=\lceil\log_2(2K/\delta)\rceil$ repetitions:
$$\Pr\left( \epsilon_k>0.25\right)=\Pr\left( \forall \ell\in[L],\,\, \epsilon_{k,\ell}>0.25\right) \leq (0.1)^L \leq 2^{-L} \leq \frac{\delta}{2K}\,. $$
 By a union bound over all $k \in [K]$, over the randomness of subsampling in \Cref{alg:adaboost-cot} 
 $$\Pr\left( \forall k \in [K]:\,\, \epsilon_k \leq 0.25 \right)\geq 1-\delta/2 \,. \qed$$
 
\end{proof}
We now apply the above weak learning guarantee in the standard analysis of Ada-Boost to conclude that the final predictor $\widehat{h}_K$ has zero training error.
\begin{lemma}[Zero training error via exponential decay]
\label{lem:zero-training-error}
For every training set $S_{\sete}=\{(x^{(i)},y^{(i)}): i \in [m]\}$, as long as $\epsilon_k \leq 0.25$ for all $k\in [K]$ when running \Cref{alg:adaboost-cot} on $S_{\sete}$, for any
$$K \geq 8 \log m\,,$$
we ensure that the predictor $\widehat h_K$ achieves \emph{zero training error} on $S_{\sete}$, i.e.,
\[
\sum_{i=1}^m \mathbf{1}\!\left[\widehat h_K(x^{(i)})\neq y^{(i)}\right]=0.
\]
If the outer for-loop in \Cref{alg:adaboost-cot} has already terminated early, then the training error must be zero. This is because early termination occurs only when $\epsilon_k=0$, which implies zero error on the dataset, since every training point has positive mass by the update rule of the algorithm.
\end{lemma}
\begin{proof}
Fix a training set $S_{\sete}=\{(x^{(i)},y^{(i)}): i\in [m]\}$, and run \Cref{alg:adaboost-cot}.
Using the standard boosting calculations (see all the details in \Cref{clm:emprical-erro-bound}), we can bound the empirical error by
  \[
\frac{1}{m}\sum_{i=1}^m \mathbf{1}\!\left[\widehat h_K(x^{(i)})\neq y^{(i)}\right]
\;\le\;
\prod_{k=1}^K 2\sqrt{\epsilon_k(1-\epsilon_k)}.
\]
Under the event that $\epsilon_k\le 0.25$ for every round of the outer loop $k\in [K]$, each factor satisfies
\[
2\sqrt{\epsilon_k(1-\epsilon_k)} \;\le\; 2\sqrt{0.25\cdot 0.75} \;=\; \sqrt{0.75}.
\]
Hence
\[
\frac{1}{m}\sum_{i=1}^m \mathbf{1}\!\left[\widehat h_K(x^{(i)})\neq y^{(i)}\right]
\;\le\;
(\sqrt{0.75})^{K}.
\]
Therefore, for any
\[ K \geq 8 \log m >\frac{2\log m}{\log(4/3)}, \quad \text{ we have }
\big(\sqrt{0.75}\big)^K \;<\; \frac{1}{m}\,,
\]
then the training error is strictly smaller than $1/m$. This implies that it must equal $0$, as the training error is always an integral multiple of $1/m$, concluding the proof. 
\end{proof}
Now the final step is to use uniform convergence between empirical and population error for predictors that are weighted votes of $K$ predictors from the class $\ete{\cF}{T}$. Note that the final predictor comes from the class
\begin{equation}\label{eq:boosted-class}
   \cH^{\mathrm{boost}}(K)
\;\coloneqq\;
\Big\{
x \mapsto \arg \max_{b\in \{0,1\}} \Big(\sum_{k=1}^K \alpha_k\, \mathbf{1}[\ete{f_k}{T}(x)=b]\Big)
\;:\;
f_1,\ldots,f_K \in \mathcal{F},\ \alpha_1,\ldots,\alpha_K \in \mathbb{R}_{>0}
\Big\}\,.
\end{equation}
Recoding both the base predictors and the output labels via $h\mapsto 2h-1$, and retaining the same notation, we can equivalently write
\begin{equation}\label{eq:boosted-class-pm 1}
\cH^{\mathrm{boost}}(K)
=
\Big\{
x \mapsto \sign\!\Big(\sum_{k=1}^K \alpha_k\, \ete{f_k}{T}(x)\Big)
\;:\;
f_1,\ldots,f_K \in \mathcal{F},\
\alpha_1,\ldots,\alpha_K \in \mathbb{R}_{>0}
\Big\}\,.
\end{equation}
with breaking ties in favor $+1$. We must control the capacity of this class. The VC dimension of this class is well-known in terms of the capacity of the base class of weak learners, which in our case is $\ete{\cF}{T}$. Using \cite[Lemma 10.3]{shalev2014understanding}, we have that
\begin{equation}\label{eq:VC-of-boosted-class}
\VCdim(\mathcal{H}^{\mathrm{boost}}(K))
=O\left(
K \cdot \VCdim(\ete{\cF}{T})
\log\left(K \cdot \VCdim(\ete{\cF}{T})\right)
\right)\,.
\end{equation}
By standard uniform convergence on the class $\cH^\mathrm{boost}(K)$ (\Cref{prop:fundamental-VC}), we have the following guarantee.
\begin{lemma}\label{lem:uc-on-boosted}
There exists a universal constant $c>0$, such that, for any distribution $\cD$ over $\cX$ and $\delta\in (0,1)$, with probability at least $1-\delta/2$ over a draw $S_{\sete}=\{(x^{(i)},y^{(i)}): x^{(i)} \simiid \cD, y^{(i)}=h_*(x^{(i)})\}$ of size $m$, with
$$
m \geq c\left( \frac{K \cdot \VCdim(\ete{\cF}{T}) \log \left(K \cdot \VCdim(\ete{\cF}{T})\right)\log(2/\eps)+\log\delta^{-1}}{\eps}\right)\,,
$$
any E2E consistent predictor $\widehat{h}\in\cH^{\mathrm{boost}}(K)$ satisfies
$$
L_{\cD,h_*}(\widehat{h}) \leq \eps\,.
$$
\end{lemma}
\begin{proof}
This follows from standard uniform convergence (\Cref{prop:fundamental-VC}) in the realizable setting over the class $\mathcal{H}^{\mathrm{boost}}(K)$ and the VC-dimension bound from \Cref{eq:VC-of-boosted-class}.
\end{proof}
We are now ready to prove our main \Cref{thm:boosting-model-aggregation}.

\begin{proof}[Proof of \Cref{thm:boosting-model-aggregation}]  We start by noting that the final predictor lies in the class $\cH^\mathrm{boost}(K)$ with $K=\lceil8\log(2m)\rceil=O(\log(2m))$, and hence the effective complexity of the class depends on the size of the end-to-end training set. To use \Cref{lem:uc-on-boosted}, thus, we must choose $m$ satisfying:
$$m \geq c\left( \frac{K \cdot \VCdim(\ete{\cF}{T}) \cdot \log\left(K \cdot \VCdim(\ete{\cF}{T})\right) \log(2/\eps)+ \log\delta^{-1}}{\eps}\right),$$
to ensure that the final predictor $\widehat{h}_K$ which has zero error on end-to-end dataset satisfies the desired guarantee. By \Cref{lem:algebraic-condition}, there exists another universal constant $c>0$ such that
\[
m
\geq
c\left(
\frac{
\VCdim(\ete{\cF}{T}) \log(2/\eps)
\log^2\!\left(\frac{2\VCdim(\ete{\cF}{T})}{\eps}\right)+\log\delta^{-1}
}{\eps}
\right)\,,
\]
always satisfies the desired bound. Therefore, an application of \Cref{lem:uc-on-boosted} yields that, with probability at least $1-\delta/2$ over the randomness of a training set $S_{\sete}$ of size $m$, any consistent predictor from the class $\cH^\mathrm{boost}(K)$ with $K=\lceil8\log(2m)\rceil$ has error at most $\eps$.

By \Cref{lem:weak-learner-generalization}, we can choose
$\M\leq c\,\VCdim(\ete{\cF}{T})$, independently of $\eps$. Combining \Cref{lem:weak-learner-generalization,lem:zero-training-error}, we have that with probability $1-\delta/2$ over its own randomness, the predictor $\widehat{h}_K$ returned by \Cref{alg:adaboost-cot} is a valid zero training error solution, and the predictor $\widehat{h}_K$ indeed belongs to the class $\cH^\mathrm{boost}(K)$.

A union bound over the two events gives us that with probability at least $1-\delta$,
 $$L_{\cD,h_*}(\widehat{h}_K):=\Pr_{x\sim \cD}\left( \widehat{h}_K(x)\neq h_*(x)\right) \le \eps\,.$$
 The total number of rounds in which CoTs are collected in \Cref{alg:adaboost-cot}, counting both the inner and outer loops, is clearly bounded by \(\kappa \leq K\lceil\log_2(2K/\delta)\rceil=O\!\left(\log m\,\log\frac{\log m}{\delta}\right)\).\\
\emph{Runtime.} The total runtime is polynomial in \(T\) and in the size \(m=m(\eps,\delta)\) of the initial end-to-end training set, provided that the base class has tractable $\Cons$ oracle (\Cref{def:cons-oracle}). Indeed, given CoT samples, constructing a predictor consistent with them only requires decomposing each CoT into all prefix--next-token pairs, which can be done in time polynomial in \(m\) and \(T\). Thus, as long as \(m(\eps,\delta)\) is bounded by \(\poly(d,T,\eps^{-1},\log\delta^{-1})\), already ensured by the fact that \(\VCdim(\ete{\cF_d}{T})=\poly(d,T)\), the overall runtime is polynomial in \(d\), \(T\), \(\eps^{-1}\), and \(\log\delta^{-1}\).
\end{proof}
\paragraph{Proof of a Technical Claim.} We finally return to the proof of the technical claim that the error of the aggregated predictor at the end of $K$ rounds is bounded as follows. 
\begin{claim}\label{clm:emprical-erro-bound}
In \Cref{alg:adaboost-cot}, for every input training set $S_{\sete}$, the empirical error of the final predictor $\widehat{h}_K$ is bounded as
\[
\frac{1}{m}\sum_{i=1}^m \mathbf{1}\!\left[\widehat h_K(x^{(i)})\neq y^{(i)}\right]
\;\le\;
\prod_{k=1}^K 2\sqrt{\epsilon_k(1-\epsilon_k)}.
\]
\end{claim}

\begin{proof}
For simplicity, we take the labels and predictions to be in $\{\pm 1\}$ instead of $\{0,1\}$. First consider the case where the algorithm terminates at some round $k'$ because $\epsilon_{k'}=0$. Since the weights are initialized uniformly and each multiplicative update is by a positive factor, we have $D_i^{(k')}>0$ for every $i\in[m]$. Therefore, $\epsilon_{k'}=0$ implies that $\widehat{h}_K=\ete{\widehat f_{k'}}{T}$ correctly classifies every training example. Hence the predictor returned by the algorithm has zero empirical error, and the claim follows.

Thus, without loss of generality, we may assume that the algorithm does not terminate because of a zero-error weak learner. In particular, for every executed round $k$, we have $\epsilon_k>0$, so $\alpha_k=\frac12\log((1-\epsilon_k)/\epsilon_k)$ is well-defined. We now prove the standard AdaBoost bound.

Define the real-valued ensemble score
\[
F_K(x) \;\coloneqq\; \sum_{k=1}^K \alpha_k\,\ete{\widehat f_k}{T}(x),
\qquad
\widehat h_K(x) \;=\; \sign\!\big(F_K(x)\big).
\]
For any example $(x^{(i)},y^{(i)})$, if $\widehat h_K(x^{(i)})\neq y^{(i)}$ then $y^{(i)}F_K(x^{(i)})\le 0$, hence
\[
\mathbf{1}\!\left[\widehat h_K(x^{(i)})\neq y^{(i)}\right]
\;\le\;
\exp\!\big(-y^{(i)}F_K(x^{(i)})\big).
\]
Averaging over $i\in[m]$ and using $D_i^{(1)}=1/m$ gives
\begin{equation}
\label{eq:err-exp-upper}
\frac{1}{m}\sum_{i=1}^m \mathbf{1}\!\left[\widehat h_K(x^{(i)})\neq y^{(i)}\right]
\;\le\;
\sum_{i=1}^m D_i^{(1)} \exp\!\big(-y^{(i)}F_K(x^{(i)})\big).
\end{equation}
Now let us relate the RHS to the product of normalization factors. For each $k$, define the normalization constant
\[
Z_k \;\coloneqq\; \sum_{i=1}^m D_i^{(k)}\exp\!\big(-\alpha_k\,y^{(i)}\,\ete{\widehat f_k}{T}(x^{(i)})\big).
\]
The weight update in \Cref{alg:adaboost-cot} can be written as
\[
D_i^{(k+1)}
\;=\;
\frac{D_i^{(k)}\exp\!\big(-\alpha_k\,y^{(i)}\,\ete{\widehat f_k}{T}(x^{(i)})\big)}{Z_k}.
\]
Iterating this identity over $k=1,\dots,K$ yields, for every $i$,
\[
D_i^{(K+1)}
\;=\;
\frac{D_i^{(1)}\exp\!\Big(-\sum_{k=1}^K \alpha_k\,y^{(i)}\,\ete{\widehat f_k}{T}(x^{(i)})\Big)}{\prod_{k=1}^K Z_k}
\;=\;
\frac{D_i^{(1)}\exp\!\big(-y^{(i)}F_K(x^{(i)})\big)}{\prod_{k=1}^K Z_k}.
\]
Summing both sides over $i$ and using $\sum_i D_i^{(K+1)}=1$ gives
\[
\sum_{i=1}^m D_i^{(1)}\exp\!\big(-y^{(i)}F_K(x^{(i)})\big)
\;=\;
\prod_{k=1}^K Z_k.
\]
Combining with \eqref{eq:err-exp-upper} yields the standard AdaBoost training-error bound
\begin{equation}
\label{eq:train-prodZ}
\frac{1}{m}\sum_{i=1}^m \mathbf{1}\!\left[\widehat h_K(x^{(i)})\neq y^{(i)}\right]
\;\le\;
\prod_{k=1}^K Z_k.
\end{equation}

Now we compute $Z_k$ as the factor $2\sqrt{\epsilon_k(1-\epsilon_k)}$.
Let $C_k$ be the set of indices correctly classified by $\ete{\widehat f_k}{T}$ at round $k$ and $W_k$ the misclassified set:
\[
C_k \coloneqq \{i:\ete{\widehat f_k}{T}(x^{(i)})=y^{(i)}\},\qquad
W_k \coloneqq \{i:\ete{\widehat f_k}{T}(x^{(i)})\neq y^{(i)}\}.
\]
Then for $i\in C_k$ we have $y^{(i)}\,\ete{\widehat f_k}{T}(x^{(i)})=+1$, and for $i\in W_k$ we have
$y^{(i)}\,\ete{\widehat f_k}{T}(x^{(i)})=-1$. Therefore
\[
Z_k
=\sum_{i\in C_k} D_i^{(k)}e^{-\alpha_k}+\sum_{i\in W_k} D_i^{(k)}e^{+\alpha_k}
=(1-\epsilon_k)e^{-\alpha_k}+\epsilon_k e^{\alpha_k}.
\]
With $\alpha_k=\frac12\log\!\left(\frac{1-\epsilon_k}{\epsilon_k}\right)$,
\[
e^{\alpha_k}=\sqrt{\frac{1-\epsilon_k}{\epsilon_k}}
\quad\text{and}\quad
e^{-\alpha_k}=\sqrt{\frac{\epsilon_k}{1-\epsilon_k}}.
\]
Plugging these into the expression for $Z_k$ gives
\[
Z_k
=(1-\epsilon_k)\sqrt{\frac{\epsilon_k}{1-\epsilon_k}}
+\epsilon_k\sqrt{\frac{1-\epsilon_k}{\epsilon_k}}
=
2\sqrt{\epsilon_k(1-\epsilon_k)}.
\]
Substituting back into \eqref{eq:train-prodZ} yields
\[
\frac{1}{m}\sum_{i=1}^m \mathbf{1}\!\left[\widehat h_K(x^{(i)})\neq y^{(i)}\right]
\;\le\;
\prod_{k=1}^K 2\sqrt{\epsilon_k(1-\epsilon_k)}. \,\,\qed 
\]

\end{proof}


\section{A case study on learning noisy parities}\label{sec:parity}
In this section, we present a case study on learning noisy parities with Chain-of-Thought under the supervision of multiple generators. Recall the discussion in \Cref{subsec:parity-main-paper}. Our goal is to show that a simple polynomial-time algorithm (\Cref{alg:support-recovery-parity}) recovers the hidden support even when the CoTs come from \emph{super-exponentially} many generators (\Cref{subsec:parity-algorithm}). This holds even when the generator is chosen adversarially on each example without the provenance (\Cref{thm:propogating-noise-recovery}).

Moreover, the sample complexity of the algorithm nearly matches the information-theoretic limit of \Cref{thm:it-limit-cot} under the supervision of these generators sampled uniformly for each example, even in the case when there is no noise in the CoT generation process (\Cref{subsec:parity-main-paper}). Interestingly, this information-theoretic limit is genuinely larger than in the single-generator setting (\Cref{rem:small-samples-suffice-from-single-thinker}). Finally, we show numerical simulations of transformers trained with the next-token prediction objective on noiseless parities (\Cref{subsec:transformers-expriment}).

\paragraph{CoT for learning parities.} For learning parities, there is a natural CoT that reveals parities on a hierarchy of growing partial subsets, by xoring in one hidden bit at a time. Consider the class of all size-$k:=k(d)$ subset parities over $\{0,1\}^d$. For a fixed hidden support $S_\star$, a single generator produces a CoT by successively incorporating coordinates of $S_\star$ one by one, thereby revealing parities of nested subsets. Different generators correspond to different orderings of these coordinates, yielding up to $k!$ possible generators. This is formalized below.

\paragraph{Generators.} Fix parameters $(d,k)$. Let $x=(x_1,\dots,x_d)\in\{0,1\}^d$.
Fix a subset $S \subseteq [d]$ with $|S|=k$, and let
\[ \Bij([k],S):=\{\pi:[k]\to S, \,\mathrm{ s.t., } \,\pi \text{ is a bijection} \}
\]
be the set of bijection (orderings of the elements of $S$). For each $\pi\in \Bij([k],S)$, we have a generator $f_{S,\pi}$, so the total number of generators is given by $|\Bij([k],S)|=k!\,\,$ 

The generator $f_{S,\pi}$ produces a length-$k$ chain-of-thought as follows:
\[
z_0 := 0,
\qquad
z_t := z_{t-1} \oplus x_{\pi(t)},
\quad t=1,\dots,k\,.
\]
Note that the final token computes the parity of the subset $S$:
\[
y:=z_k = \bigoplus_{j\in S} x_j.
\]
Note that all generators produce the same final output, though their CoTs may differ. In this sense, the setting is similar to the general framework introduced in \Cref{subsec:CoT-Framework,subsec:multiple-CoT-supervision-intro}. However, to make end-to-end learning from only the final outputs $(x,y)$ nontrivial, we must introduce noise; otherwise, the target parity can be recovered efficiently from end-to-end labels alone by solving a system of linear equations over $\F_2$. Our interest is instead in a regime where learning from the final output alone is computationally hard, and where access to the CoT trace enables tractable learning.

To this end, we consider a noisy generation model in which, at each step of the CoT, the generated bit is flipped with some probability. Because the noise is introduced during generation itself, any error may propagate to subsequent steps. We observe samples generated according to this noisy process, potentially from different generators. This noise model has been studied experimentally before for a combination of next-token prediction and reinforcement learning algorithms applied to transformers~\cite{tsilivis2025reinforcement}.

Formally, for a noise level $\eta \in (0,1/2)$, we have the following observation model.
\paragraph{Propagating noise during generation.}
Let $(\xi_1,\dots,\xi_k)\simiid \Bern(\eta)$.
The noisy chain is generated recursively as
\[
\tilde z_0 := 0,
\qquad
\tilde z_t := (\tilde z_{t-1} \oplus x_{\pi(t)}) \oplus \xi_t,
\quad t=1,\dots,k\,.
\]
Equivalently, the noisy chain satisfies
\[
\tilde z_t
=
z_t
\;\oplus\;
\bigoplus_{i=1}^t \xi_i\,.
\]
The last token $\tilde y := \tilde z_k$ satisfies
\[
\eta_{\text{eff}}=\Pr[\tilde y \neq y]
=\Pr\left(  \bigoplus_{t=1}^k \xi_{t}=1\right)=
\frac12 \bigl(1 - (1-2\eta)^k\bigr)\,,
\]
where the last equality can be verified inductively.
\begin{remark}[Noise Accumulation]
Let $\eta_{\mathrm{eff}}:=\Pr[\tilde y\neq y]$ denote the effective noise at the final output. First, even when $\eta_{\mathrm{eff}}$ is a sufficiently small absolute constant, learning from end-to-end examples alone is still believed to be hard when $k(d)=\Omega(d)$: for parity learning with noise, the best known algorithm runs in time $2^{O(d/\log d)}$~\citep{blum2003noise}. Thus, hardness already persists at constant effective noise.

Second, $\eta_{\mathrm{eff}}$ may be much larger than the per-step generation noise $\eta$, since errors accumulate over the $k$ steps of the CoT. Concretely, if $\eta \ll 1/k$, then $\eta_{\mathrm{eff}}=\Theta(k\eta)$; if $\eta=1/(2k)$, then $\eta_{\mathrm{eff}}\to \frac12(1-e^{-1})$; and if $\eta \gg 1/k$, then $\eta_{\mathrm{eff}}\to \frac12$. In particular, when $\eta=1/4$ and $k=d/2$, we have $\eta_{\mathrm{eff}}=\Pr[\tilde y\neq y]=\frac12-\frac{1}{2^{d/2+1}}$, which is exponentially close to $1/2$. Therefore, even very small per-step noise can induce substantial effective noise at the final output, rendering end-to-end learning difficult.
\end{remark}

\paragraph{Learning problem.} For parameters $(d,k)$, the adversary chooses an unknown support $S_\star\subseteq[d]$ with $|S_\star|=k$. We observe $m$ samples generated as follows. First,
\[
(x^{(1)},\dots,x^{(m)})\simiid \Unif(\{0,1\}^d).
\]
Then, for each example $x^{(i)}$, the adversary chooses a generator $f_{S_\star,\pi^{(i)}}$ from the collection $\{f_{S_\star,\pi}:\pi\in\Bij([k],S_\star)\}$ of $k!$ possible generators, corresponding to different orders on the support coordinates. The choice of generator may depend adversarially on the entire input sequence $(x^{(1)},\dots,x^{(m)})$.

Under the noisy generation model with noise level $\eta$, the learner receives CoTs $(\tilde{z}^{(1)},\dots,\tilde{z}^{(m)})$ generated from $(f_{S_\star,\pi^{(1)}},\dots,f_{S_\star,\pi^{(m)}})$. Thus, the observed dataset is
\[
((x^{(1)},\tilde{z}^{(1)}),\dots,(x^{(m)},\tilde{z}^{(m)})),
\]
and the goal of the learner is to recover $S_\star$. By Stirling's approximation, $k(d)! = 2^{\Theta(k(d)\log k(d))}$. Hence, whenever $k(d)=\omega(d/\log d)$, the number of possible orders/CoTs satisfies $k(d)! = 2^{\omega(d)}$, and is therefore super-exponential in $d$.

\subsection{Recovery algorithm to learn from multiple generators}\label{subsec:parity-algorithm}
 We now present an algorithm (\Cref{alg:support-recovery-parity}) that recovers the support tractably, even when the Chain-of-Thoughts are generated adversarially among the $k!$ possible generators. The algorithm forms, for each example, a scalar statistic from the adjacent differences of the observed CoT and then computes coordinate-wise correlations between that statistic and the input bits across the dataset to infer the $k$ indices of the hidden support.

 \begin{algorithm}[H]
\caption{Parity support recovery algorithm under heterogeneous CoT supervision.}
\label{alg:support-recovery-parity}
\begin{algorithmic}
\State \textbf{Input:} Samples $\{(x^{(i)}, \tilde z^{(i)})\}_{i=1}^m$,
where $x^{(i)} \in \{0,1\}^d$ and 
$\tilde z^{(i)} \in \{0,1\}^k$; assume $|S_\star|=k$.
\State \textbf{Output:} Estimated support $\widehat S \subseteq [d]$.
\begin{itemize}
    \item For each $i\in[m]$ and $t\in[k]$, define increments (with $\tilde z^{(i)}_0=0$) by
    \[
    \tilde\delta^{(i)}_t 
    \;:=\;1-
    2\cdot\bigl(\tilde z^{(i)}_t \oplus \tilde z^{(i)}_{t-1}\bigr),
    \qquad\text{and set}\qquad
    \tilde s^{(i)} := \sum_{t=1}^k \tilde\delta^{(i)}_t.
    \]
    \item For each coordinate $j\in[d]$, compute
    \[
    \mathrm{score}(j)
    =
    \frac{1}{m}
    \sum_{i=1}^m
    \left(1-2x^{(i)}_j\right)\, \tilde s^{(i)}.
    \]
    \item Output $\widehat S$ as the $k$ indices with largest $\mathrm{score}(j)$.
\end{itemize}
\end{algorithmic}
\end{algorithm}

We show the following guarantee 
 
 \begin{theorem}[Guarantee for \Cref{alg:support-recovery-parity}]
\label{thm:propogating-noise-recovery}
Fix $d\in\mathbb{N}$ and $k\le d$. Let $S_\star\subseteq[d]$ be unknown with $|S_\star|=k$.
Assume the samples $\{(x^{(i)},\tilde z^{(i)})\}_{i=1}^m$ are generated under the propagating noise during generation with the noise level $\eta\in[0,1/2)$. The generators $(\pi^{(1)},\dots,\pi^{(m)})$ can be chosen even adversarially based on $(x^{(1)},\dots,x^{(m)})$. Then the output of $\widehat S$ of \Cref{alg:support-recovery-parity} satisfies the following guarantee. There exists a universal constant $C>0$ such that for every $\delta\in(0,1)$,
if
\[
m \;\ge\; C\cdot \frac{k}{(1-2\eta)^2}\,\log\!\Big(\frac{d}{\delta}\Big), \quad \mathrm{then } \quad \Pr\!\big[\widehat S = S_\star\big]\;\ge\; 1-\delta\,.
\]
The algorithm runs in time $\poly(d,1/(1-2\eta),\log \delta^{-1})$.
\end{theorem}

\begin{proof}[Proof of \Cref{thm:propogating-noise-recovery}] 
The runtime bound follows directly from \Cref{alg:support-recovery-parity} once the sample bound is established, which we do now. Under our noise model:
    $$ \tilde z^{(i)}_t \oplus \tilde z^{(i)}_{t-1} = x_{\pi^{(i)}(t)}^{(i)} \oplus \xi^{(i)}_{t}\,.$$
Therefore,
$$\tilde\delta_t^{(i)}=1-2\left(\tilde z^{(i)}_t \oplus \tilde z^{(i)}_{t-1}\right) = \left(1-2x_{\pi^{(i)}(t)}^{(i)}\right)\left(1- 2 \xi_{t}^{(i)}\right)=\left(1-2x_{\pi^{(i)}(t)}^{(i)} \right)\gamma_{t}^{(i)}\, ,$$
where we defined  
$\gamma^{(i)}_t=\left(1- 2 \xi_{t}^{(i)}\right) \in \{\pm 1\}$ for which $ \Pr[\gamma_t^{(i)}=+1]=1-\eta\,.$ Therefore, the statistic can be simplified as follows:
\begin{align}
    \tilde{s}^{(i)}=\sum_{t=1}^k  \tilde\delta^{(i)}_t &= \sum_{t=1}^k\left(1-2x_{\pi^{(i)}(t)}^{(i)}\right) \cdot \gamma^{(i)}_t \,\nonumber
     =\sum_{\ell \in S_\star}\left(1-2x_{\ell}^{(i)}\right) \cdot\gamma^{(i)}_{\pi^{(i)}_{-1}(\ell)} \,. 
\end{align}
We claim that the statistics $\tilde s^{(1)},\dots,\tilde s^{(m)}$ are independent, even though the permutations $\pi^{(1)},\dots,\pi^{(m)}$ may be chosen adversarially after observing the entire input sequence $(x^{(1)},\dots,x^{(m)})$. Indeed, conditional on the inputs $(x^{(1)},\dots,x^{(m)})$ and the adversarially chosen permutations $(\pi^{(1)},\dots,\pi^{(m)})$, the random variables $\tilde s^{(i)}$ depend only on the noise vectors $\gamma_{1:k}^{(i)}$. Since the coordinates $\gamma_1^{(i)},\dots,\gamma_k^{(i)}$ are i.i.d., permuting them does not change their joint distribution; hence for each fixed $i$, the conditional law of $\tilde s^{(i)}$ depends only on $x^{(i)}$ and not on the particular order $\pi^{(i)}$. Because the noise vectors are independent across $i$, it follows that $\tilde s^{(1)},\dots,\tilde s^{(m)}$ are independent.


For any $j \in [d]$, we define:
\begin{align*}
    Z^{(i)}_j:=\left(1-2x^{(i)}_j\right)\tilde{s}^{(i)}\,.
\end{align*}
We have $\{Z_{j}^{(i)}; i \in [m]\}$ are also independent. The rest of the proof follows by applying Bernstein's lemma with mean and variance calculations. 

\paragraph{Mean:} Recall that $\E[\gamma^{(i)}_t]=(1-\eta)-\eta=1-2\eta$. And $\E\left[\left(1-2x_j^{(i)}\right)\left(1-2x_\ell^{(i)}\right)\right]=\mathbf{1}_{j=\ell}$. Averaging first over the generation noise and then reindexing over $S_\star$, we obtain:
\begin{align}
    \E[Z^{(i)}_j]
    &=\E\!\left[\left(1-2x^{(i)}_j\right)\tilde{s}^{(i)}\right]\nonumber\\
    &=(1-2\eta)\E\!\left[\left(1-2x_j^{(i)}\right)
    \sum_{t=1}^k\left(1-2x_{\pi^{(i)}(t)}^{(i)}\right)\right]\nonumber\\
    &=(1-2\eta)\sum_{\ell\in S_\star}
    \E\!\left[\left(1-2x_j^{(i)}\right)\left(1-2x_\ell^{(i)}\right)\right]\nonumber\\
    &=(1-2\eta)\sum_{\ell\in S_\star}\mathbf{1}_{j=\ell}\nonumber\\
    &=(1-2\eta)\mathbf{1}_{j\in S_\star}
    =:\mu\mathbf{1}_{j\in S_\star}\,. \label{eq:mean}
\end{align}

\paragraph{Variance:} Since $(1-2x_j^{(i)})^2=1$, averaging first over the generation noise and then reindexing the full cross-term sum over pairs in $S_\star$, we obtain
\begin{align*}
\E\bigl[(Z_j^{(i)})^2\bigr]
&=\E\bigl[(\tilde s^{(i)})^2\bigr]
=k+2\mu^2\E\!\left[
\sum_{1\le t<u\le k}
(1-2x_{\pi^{(i)}(t)}^{(i)})
(1-2x_{\pi^{(i)}(u)}^{(i)})
\right]\\
&=k+2\mu^2
\sum_{\substack{\ell,r\in S_\star\\\ell<r}}
\E\!\left[(1-2x_\ell^{(i)})(1-2x_r^{(i)})\right]
=k.
\end{align*}
Here, the diagonal terms equal $1$, and the reindexing is valid because $\pi^{(i)}$ is a bijection onto $S_\star$. The fixed-coordinate cross terms have zero expectation. Thus,
\[
\Var(Z_j^{(i)})
\le \E\bigl[(Z_j^{(i)})^2\bigr]
=k.
\]
\medskip\noindent
\textbf{Applying Bernstein.} We now apply Bernstein's lemma (\Cref{lem:bernstein-mean}). Note that for any fixed $j\in [d]$, we have $|Z^{(i)}_j-\E[Z^{(i)}_j]| \leq 2k$ almost surely. The value of $\sigma^2=\frac{1}{m}\sum_{i=1}^m\Var[Z^{(i)}_j] \leq k$. Applying \Cref{lem:bernstein-mean}: 

For any $j \in [d]$, setting $\epsilon=\mu/4$ yields,
\[
\Pr\!\left(
\left|\Score(j)-\E[Z^{(i)}_j]\right|
\ge \frac{\mu}{4}
\right)
\le
2\exp\!\left(-c\frac{m\mu^2}{k}\right)
\]
for a universal constant $c>0$. Doing a union bound over $j \in [d]$: with probability at least
\[
\Pr\left(\forall j \in [d], \, |\Score(j)-\E[Z^{(i)}_j]| \leq \frac{\mu}{4}\right) \geq 1-2d\exp\!\left(-c\frac{m\mu^2}{k}\right)
\]
Under this event, we have simultaneously:
\[
\Score(j)\ge \frac{3\mu}{4}
\quad \forall j\in S_\star,
\qquad
\Score(j)\le \frac{\mu}{4}
\quad \forall j\notin S_\star.
\]
Thus, the top-$k$ rule of \Cref{alg:support-recovery-parity} returns $\widehat S=S_\star$ with probability at least $1-\delta$, as long as
\[
2d\exp\!\left(-c\frac{m\mu^2}{k}\right)
\le \delta\,, \quad \text{ which is implied by } \quad
\quad
m \ge C\cdot \frac{k}{\mu^2}\log\!\Big(\frac{d}{\delta}\Big),
\]
for some $C>0$. Noting $\mu= (1-2\eta)$ completes the proof.  
\end{proof}

\subsection{Information theoretic limit of learning from uniform mixture}\label{subsec:it-limit-parity}
We now show that the $\Omega(k\log d)$ bound in the theorem above is also nearly necessary information-theoretically, even in the more favorable case where each generator is chosen uniformly from $\Bij([k],S_\star)$. By contrast, for any fixed single generator, $O(\log(d/\delta))$ samples suffice under constant per-step noise $\eta$ (cf.~\Cref{rem:small-samples-suffice-from-single-thinker}).

\sloppy
We consider the noiseless case $\eta=0$, in which the learner observes CoTs of the form $(x_1,\dots,x_d,z_1,\dots,z_k)$ that reveal the partial subset parities exactly.  The proof uses Fano's inequality.

\begin{theorem}[Information-theoretic limit from a random CoT generator]
\label{thm:it-limit-cot}
Fix $d\in\mathbb{N}$ and $1\le k\le d-1$. Let $S_\star$ be drawn uniformly from
$\binom{[d]}{k}$. Given $S_\star$, draw $m$ i.i.d.\ samples $(x^{(i)}, z^{(i)})_{i=1}^m$, where $z^{(i)}$ is a CoT generated by a random generator $\pi^{(i)}\simiid \Bij([k],S_\star)$.  

Then for any (possibly randomized) estimator $\widehat S=\widehat S\big((x^{(i)}, z^{(i)})_{i=1}^m\big)$, achieving $\Pr[\widehat S=S_\star]\ge 1-\delta$ requires
\[
m \;\ge\; \frac{(1-\delta)\log \binom{d}{k}-1}{\log(k+1)}.
\]
(where the probability is over $S_\star$ and samples). Moreover, for $k\leq d/2$, using $\log\binom{d}{k}\ge k\log(d/k)$, the bound becomes 
\[
m \;\ge\; \frac{(1-\delta)\,k\log(d/k)-1}{\log(k+1)}.
\]
\end{theorem}
\begin{proof}
Let $M:=\binom{d}{k}$ and let $S_\star$ be uniform over the $M$ possible supports.
Let the full dataset be
\[
D := \big((x^{(1)}, z^{(1)}),\ldots,(x^{(m)}, z^{(m)})\big),
\]
where $x^{(i)}=(x^{(i)}_1,\ldots,x^{(i)}_d)$ and $z^{(i)}=(z^{(i)}_1,\ldots,z^{(i)}_k)$.

\paragraph{The sum is a sufficient statistic.}
For each sample $i\in[m]$, define
\[
U^{(i)} \;:=\; \big|\{j\in S_\star : x^{(i)}_j=1\}\big|\in\{0,1,\ldots,k\}.
\]
We claim that, conditional on $x^{(i)}$, the conditional law of $z^{(i)}$ depends on $S_\star$
only through $U^{(i)}$, i.e., for every measurable set $A\subseteq \mathcal{Z}^k$,
\begin{equation}
\label{eq:suff}
\Pr\!\big(z^{(i)}\in A \mid x^{(i)},S_\star\big)
\;=\;
\Pr\!\big(z^{(i)}\in A \mid x^{(i)},U^{(i)}\big).
\end{equation}
Equivalently, $S_\star \perp z^{(i)} \mid (x^{(i)},U^{(i)})$. Indeed, conditioned on $x^{(i)}$, the multiset $\{x^{(i)}_j : j\in S_\star\}$ contains exactly $U^{(i)}$
many $1$'s and $k-U^{(i)}$ many $0$'s. Since $\pi^{(i)}$ is uniform over permutations of $S_\star$, the
increment sequence
\[
\big(x^{(i)}_{\pi^{(i)}(1)},\ldots,x^{(i)}_{\pi^{(i)}(k)}\big)
\]
is a uniformly random ordering of this multiset, and therefore its conditional distribution given $x^{(i)}$
depends on $S_\star$ only through $U^{(i)}$. The sequence
\(
\big( z^{(i)}_1,\dots,z^{(i)}_k\big)_{t=1}^k
\)
is a deterministic function of these increments,
hence \eqref{eq:suff} holds.

\paragraph{Step 2 (mutual information upper bound).}
By \eqref{eq:suff} and the data-processing inequality, conditioning on the entire input sequence
$x^{(1:m)}:=(x^{(1)},\ldots,x^{(m)})$,
\begin{align}
   I(S_\star;D \mid x^{(1:m)})
\;&\le\;
I(S_\star;U^{(1:m)}\mid x^{(1:m)}),
\qquad
U^{(1:m)}:=(U^{(1)},\ldots,U^{(m)}). \nonumber\\
& \leq
H(U^{(1:m)}\mid x^{(1:m)})
\;\le\;
\sum_{i=1}^m H(U^{(i)}\mid x^{(i)})
\;\le\;
m\log(k+1),\label{eq:MI-upper}
\end{align}
since each $U^{(i)}$ takes values in $\{0,\ldots,k\}$.
Finally, since $x^{(1:m)}$ is independent of $S_\star$, by chain-rule
\begin{equation}\label{eq:MI-uncond}
  I(S_\star;D)
\;\le\;
I(S_\star;D,x^{(1:m)})
=
I(S_\star;x^{(1:m)}) + I(S_\star;D\mid x^{(1:m)})
=
I(S_\star;D\mid x^{(1:m)}) \leq m \log (k+1),  
\end{equation}
where the last inequality uses \eqref{eq:MI-upper}. 
\paragraph{Applying Fano.}
Let $\widehat S=\widehat S(D)$ be any estimator and let $P_e:=\Pr[\widehat S\neq S_\star]$.
Fano's inequality in its standard form gives
\begin{equation}
\label{eq:fano-general}
P_e \;\ge\; 1-\frac{I(S_\star;D)+1}{\log M}.
\end{equation}

Combining \eqref{eq:MI-uncond} and \eqref{eq:fano-general} yields the following \emph{general} bound:
\begin{equation}
\label{eq:general-bound}
\Pr[\widehat S=S_\star]
=1-P_e
\;\le\;
\frac{I(S_\star;D)+1}{\log M}
\;\le\;
\frac{m\log(k+1)+1}{\log\binom{d}{k}}.
\end{equation}
Thus, if $\Pr[\widehat S=S_\star]\ge 1-\delta$, then by \eqref{eq:general-bound} we must have
\[
1-\delta \;\le\; \frac{m\log(k+1)+1}{\log\binom{d}{k}},
\]
which rearranges to
\[
m \;\ge\; \frac{(1-\delta)\log\binom{d}{k}-1}{\log(k+1)}.
\]
This is exactly the stated information-theoretic necessary condition.
\end{proof}
We conclude by observing that, under a single-generator supervision, one can still recover the parity from only $O(\log(d/\delta))$ samples whenever the per-step noise level $\eta<1/2$ is constant.
\begin{remark}[Learning with $O(\log(d/\delta))$ samples from a single generator]\label{rem:small-samples-suffice-from-single-thinker}
Fix any single generator, corresponding to an ordering $(j_1^\star,\dots,j_k^\star)$ of the support $S_\star=\{j_1^\star,\dots,j_k^\star\}$. Under the noisy generation model, the increments
$\tilde\delta_t^{(i)}:=1-2(\tilde z_t^{(i)}\oplus \tilde z_{t-1}^{(i)})$
satisfy
\[
\tilde\delta_t^{(i)}=(1-2x_{j_t^\star}^{(i)})\gamma_t^{(i)},
\]
where $\gamma_t^{(i)}\in\{\pm1\}$ is independent of $x^{(i)}$ and has mean $\E[\gamma_t^{(i)}]=1-2\eta$. Thus, for every fixed $t\in[k]$, the variable $\tilde\delta_t^{(i)}$ is simply a noisy copy of the bit $x_{j_t^\star}^{(i)}$ in $\{\pm1\}$ notation.

It follows that for each $j\in[d]$, the correlation statistic
\[
\Score_t(j):=\frac1m\sum_{i=1}^m (1-2x_j^{(i)})\tilde\delta_t^{(i)}
\]
has expectation
\[
\E[\Score_t(j)]
=
\begin{cases}
1-2\eta, & j=j_t^\star,\\
0, & j\neq j_t^\star.
\end{cases}
\]
Hence the correct coordinate $j_t^\star$ has a constant correlation gap over all incorrect coordinates, as long as $\eta<1/2$. Since each summand is bounded in $[-1,1]$, a standard Hoeffding or Bernstein bound implies that for any fixed $t\in[k]$, with probability at least $1-\delta'$, one can identify $j_t^\star$ correctly from $m=O(\log(d/\delta'))$ samples by selecting the coordinate with the largest empirical correlation.

Finally, since $k(d)\le d$, we may recover all coordinates $(j_1^\star,\dots,j_k^\star)$ simultaneously by a union bound over the $k$ positions. Taking $\delta'=\delta/d$ therefore yields that
\[
m = O(\log(d/\delta))
\]
samples from any fixed single generator suffice to recover the entire support $S_\star=\{j_1^\star,\dots,j_k^\star\}$ with probability at least $1-\delta$. Thus, the \emph{information-theoretic} limit indeed shifts when the learner receives CoTs from a uniform mixture of generators (cf.~\Cref{thm:it-limit-cot}).
\end{remark}

\subsection{Experiments with transformers trained via next token prediction}\label{subsec:transformers-expriment}
We finally provide experiments on transformers trained with the next-token prediction objective on noiseless parities. We compare training on sequences drawn from a single source (a fixed CoT for all samples) versus training on data sampled uniformly across the $k!$ generators. The goal is to predict correctly the final bit, i.e., the parity of the hidden set. This corresponds to the setting we studied theoretically with noise $\eta$ set to 0. Note that, even though end-to-end learning is possible in the noiseless case, as noted earlier, it remains hard for the natural class of Statistical Query algorithms~\cite{Kea98} as long as $\min\left\{k, d-k\right\} = \omega_d(1)$. Thus, we do not expect transformer-based end-to-end learning to succeed.

We train a depth-4 autoregressive transformer $f_\theta$ with the next-token prediction objective applied to the CoT of each sequence:
\begin{equation}
\min_{\theta} \mathbb{E}_{(x, z)} \left[ - \sum_{t = 1}^k \log f_{\theta} \left( z_t \mid x, z_{<t} \right) \right]
\end{equation}
In the case of a single generator, the CoT $z$ always corresponds to a fixed permutation of the indices of the support. In the case of multiple generators, each CoT $z$ is generated by sampling a permutation uniformly at random. We sample a new batch of 256 examples per training iteration and train using the AdamW optimizer~\cite{LoHu19} with a fixed learning rate of $10^{-4}$. Evaluation is performed by sampling the most probable next token (greedy decoding).

First, we show test accuracy versus training iterations for the case of a single versus multiple generators in Figure~\ref{fig:ntp_single_vs_many_thinkers_acc}. We observe that, even in the presence of $k!$ generators and when $k = d/2$, the test accuracy reaches 100\%, i.e., learning succeeds (Figure~\ref{fig:ntp_single_vs_many_thinkers_acc}, right). However, we also observe that when only a single generator is available, training succeeds in fewer iterations (Figure~\ref{fig:ntp_single_vs_many_thinkers_acc}, left -- note the different x-axes).

\begin{figure}[H]
    \centering
    \includegraphics[scale=0.45]{Fig/ntp_mixture_single_vs_many_gens_acc.pdf}
    \caption{Test accuracy of a transformer trained on sequences encoding noiseless parities ($d = 100$, $k \in \{50, 75, 90\}$) versus training iterations. \textbf{Left:} Single-thinker supervision (a unique CoT permutation across training samples). \textbf{Right:} Multiple-thinker supervision (each CoT comes from a randomly sampled permutation). We show the mean and standard deviation across three random seeds per configuration. Note the different x-axes.}
    \label{fig:ntp_single_vs_many_thinkers_acc}
\end{figure}

\begin{figure}[H]
    \centering
    \includegraphics[scale=0.6]{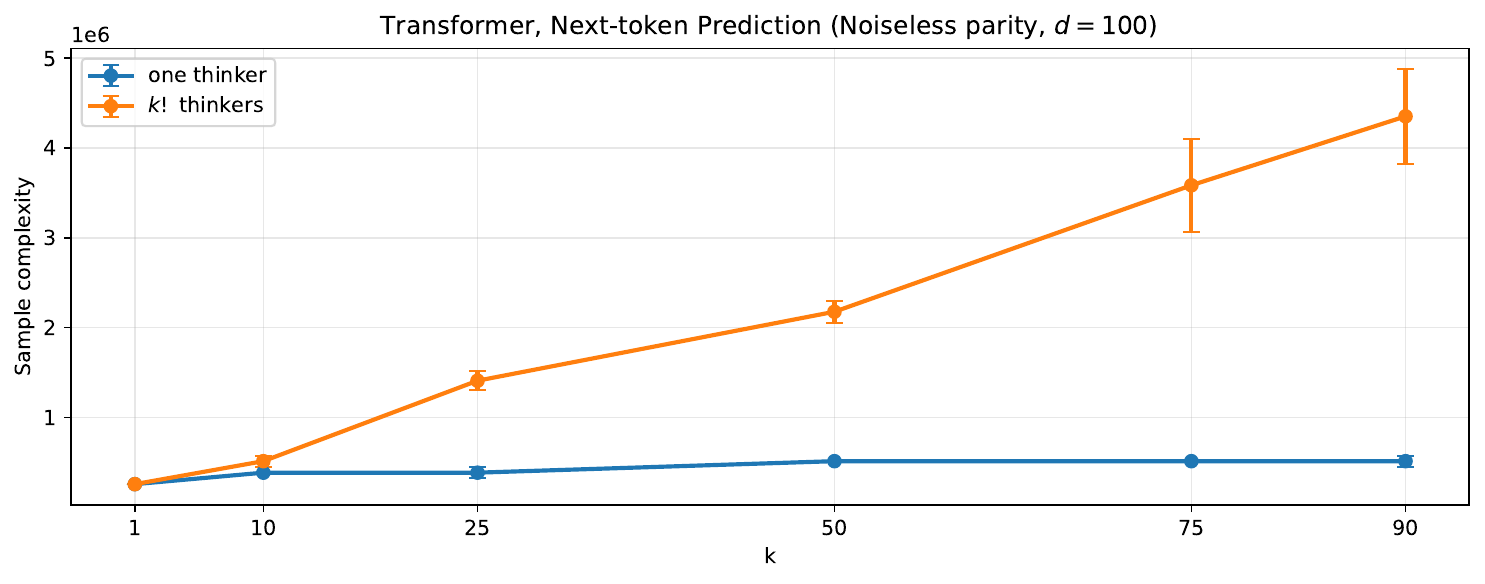}
    \caption{Sample complexity (defined as number of samples required to reach $>95\%$ test accuracy) of a transformer trained on sequences encoding noiseless parities ($d = 100$) versus support size $k$. We compare the single thinker to the multiple thinkers case.}
    \label{fig:ntp_single_vs_many_thinkers_m}
\end{figure}

In fact, in Figure~\ref{fig:ntp_single_vs_many_thinkers_m} we probe the sample complexity of next-token prediction in the two cases of either single-thinking supervision or access to multiple thinkers. In particular, we plot the number of samples required to reach greater than 95\% test accuracy versus the size of the parity support $k$. We observe that, while in the single-thinker case the sample complexity appears to be almost independent of $k$, in the multiple-thinkers case it scales (almost linearly) with $k$. This appears to be consistent with the lower bounds of Theorem~\ref{thm:it-limit-cot} and Remark~\ref{rem:small-samples-suffice-from-single-thinker}, and highlights that the presence of multiple thinkers renders learning more difficult than having access to only one of them, yet still tractable and more efficient than in the absence of thinking supervision.

\section{Regev's \texorpdfstring{$\gapsvp$}{}- and \texorpdfstring{$\sivp$}{}-based cryptosystems}\label{app:primer}
In this section, we review the \(\gapsvp\)- and \(\sivp\)-based cryptosystems due to Regev \cite{regev2009lattices}, as well as the implication of \Cref{assump:lattice} for these cryptosystems in \Cref{thm:regev's-threorem} and \Cref{cor:regev's-crypto-safe-aganist-subexp}, extending the original theorem of \cite{regev2009lattices}.

Let $n$ be the security parameter. Let $p$ be a prime satisfying
\( n^2 < p < 2n^2\), and let
\( m = 5(n+1)(1+2\log n)
\). Let $
\gamma(n) = n \log n$, which is a parameter that controls the correctness of the cryptosystem.\footnote{In both \cite{regev2009lattices,klivans2009cryptographic}, the parameter $\gamma(n)=\sqrt{n}\log^2 n$. For our purposes, however, we require a slightly faster-growing choice, since we need security against subexponential-time adversaries. So we also need the decryption error to decay at least exponentially, which is controlled by this parameter.}

\paragraph{Private key.}
A vector
\( 
s \in \Z_p^n.
\)
\paragraph{Public key.}
A sequence of pairs
\( 
(a_1,b_1),\dots,(a_m,b_m),
\)
where each $a_i \in \Z_p^n$ and each $b_i \in \Z_p$.

\paragraph{Encryption.}
To encrypt the bit $0$, choose a random subset $S \subseteq [m]$ and output
\( 
\left(\sum_{i\in S} a_i,\; \sum_{i\in S} b_i\right)
\) in $\Z_p^n \times \Z_p$.
To encrypt the bit $1$, choose a random subset $S \subseteq [m]$ and output
\( 
\left(\sum_{i\in S} a_i,\; \left\lfloor \frac{p}{2}\right\rfloor + \sum_{i\in S} b_i\right),
\)
with all arithmetic performed modulo $p$.

\paragraph{Decryption.}
Given a ciphertext $(a,b) \in \Z_p^n \times \Z_p$, compute
\( 
b - \langle a,s\rangle \pmod p
\). Decrypt to $0$ if this quantity is closer to $0$ than to $\lfloor p/2 \rfloor$ modulo $p$, and decrypt to $1$ otherwise.

\paragraph{Correctness.}
The probability of decryption error, over the choice of keys and encryption randomness, is
\( 
2^{-\Omega(\gamma(n)^2/m)}=2^{-\Omega(n \log n)}.
\)

Regev \cite{regev2009lattices} showed that any polynomial-time algorithm that distinguishes encryptions of \(0\) from encryptions of \(1\) in this scheme implies a polynomial-time quantum algorithm for \(\gapsvp\) and \(\sivp\) with certain approximation factors. Here, we note a slight generalization that also follows from his proof, which will be useful when we invoke a stronger hardness assumption against subexponential-time quantum adversaries for \(\gapsvp\) and \(\sivp\) (\Cref{assump:lattice}).
\begin{theorem}[Regev {\cite{regev2009lattices}}]\label{thm:regev's-threorem}
Consider the above cryptosystem when \( 
\gamma(n)=n \log n
\). If there is a \(\tau(n)\)-time (possibly quantum) algorithm that distinguishes
encryptions of \(0\) and \(1\) in this cryptosystem with non-negligible advantage,
then for some \(\rho(n)=\Tilde{O}(n^2)\), there is a \(O(\tau(n)\poly(n))\) time quantum algorithm for
\( \gapsvp_{\rho(n)}\) (cf. \Cref{prb:gap-SVP}) and \(\sivp_{\rho(n)}\) (cf. \Cref{prb:sivp}).
\end{theorem}

We make two remarks regarding the variation of the theorem stated above.

First, \cite{regev2009lattices,klivans2009cryptographic} use \(\gamma(n)=\sqrt{n}\log^2 n\), which yields an approximation factor of \(\Tilde{O}(n\cdot \gamma(n))=\Tilde{O}(n^{1.5})\). In our case, we use \(\gamma(n)=n\log n\), and hence obtain the worse approximation factor \(\Tilde{O}(n\cdot \gamma(n))=\Tilde{O}(n^2)\). We note that \cite[Theorem 2.3]{klivans2009cryptographic} already provides a statement of this general form, where the resulting algorithm for the lattice problem has approximation factor \(\Tilde{O}(n\cdot \gamma(n))\). In the notation of Regev \cite[Theorem 3.1]{regev2009lattices}, \(\alpha(n)=\gamma(n)^{-1}\in(0,1)\), and we note that his theorem requires only \(\alpha(n)\cdot p>\sqrt{n}\). Indeed, we have
\( 
\alpha(n)\cdot p = p\cdot \gamma(n)^{-1} > n^2\cdot (n\log n)^{-1} \gg \sqrt{n}
\).

Second, both \cite{regev2009lattices,klivans2009cryptographic} state the special case with \(\tau(n)=\poly(n)\). Here, we simply observe that the more general statement follows from the same chain of reductions \cite[Theorem~3.1 and Lemmas~4.4,~5.4]{regev2009lattices}: namely, by \cite[Lemmas~4.4 and~5.4]{regev2009lattices}, breaking the above public-key cryptosystem yields an algorithm for the relevant distinguishing problem for LWE, while \cite[Theorem~3.1]{regev2009lattices} then gives a quantum algorithm for \(\gapsvp\) and \(\sivp\). Importantly, throughout this reduction chain, the reduction makes only polynomially many oracle calls to the algorithm breaking the cryptosystem, and incurs only additional polynomial-time quantum overhead. The problem-size parameter \(n\) is preserved throughout.
 

We have the following immediate corollary of \Cref{thm:regev's-threorem} by setting \(\tau(n)=2^{o(n)}\).
\begin{corollary}\label{cor:regev's-crypto-safe-aganist-subexp}
Assuming the quantum hardness of $\gapsvp$ or $\sivp$ against subexponential time adversaries (\Cref{assump:lattice}), there is no algorithm running in time \(2^{o(n)}\) that distinguishes encryptions of \(0\) and \(1\) with non-negligible advantage.
\end{corollary}

\paragraph{Decryption functions of the cryptosystem.} We view a ciphertext \((a,b)\in \Z_p^n\times \Z_p\) as a single input \(x\), which can be encoded using \((n+1)\log p=\Theta(n \log n)\) bits. Like \cite{klivans2009cryptographic}, we denote the decryption function with secret key \((s_1,\dots,s_n)\in\Z_p^n\) by
\[
d_{s_1,\ldots,s_n}:\bigl(\{0,1\}^{\log p}\bigr)^{n+1}\to\{0,1\}.
\]
We consider the concept class of all possible decryption functions (for security parameter \(n\)) defined by
\begin{equation}\label{eq:decrypt-class-app}
    \cC_n=\{d_{s_1,\ldots,s_n}\mid (s_1,\dots,s_n) \in \Z_p^n \}.
\end{equation}

\paragraph{Cryptography and learning.}
We next note the standard proposition that learning such a concept class implies breaking the cryptosystem \cite{kearns1994cryptographic}. Here we use the precise variant stated in \cite{klivans2009cryptographic}.
\begin{proposition}[\cite{klivans2009cryptographic}, Lemma 3.1]\label{prop:crypto-learning}
Consider a public-key cryptosystem for encrypting individual bits by \(N\)-bit strings.
Let \(\mathcal{C}\) be a concept class that contains all the decryption functions
\( 
d_K : \{0,1\}^N \to \{0,1\}
\) of the cryptosystem, one for each choice of key \(K=(K_{\mathrm{priv}},K_{\mathrm{pub}})\).
Let
\( 
\epsilon(N)
=
\Pr_{K,r}\!\bigl[d_K(e_{K,r}(0)) \neq 0 \text{ or } d_K(e_{K,r}(1)) \neq 1\bigr]
\) be the probability of decryption error (over the choice of keys and randomization in the encryption).
If \(\mathcal{C}\) is weakly PAC-learnable in time \(\tau(N)\) with
\( 
\tau(N)\epsilon(N) \leq \frac{1}{N^{\omega(1)}},
\)
then there is a distinguisher between the encryptions of \(0\) and \(1\) that runs in time \(O(\tau(N))\).    
\end{proposition}
As a corollary, we obtain the following lemma.
\begin{lemma}\label{lem:final-decryption-function-hard-lemma}
Let \(\cC=\cup_{n \geq 1}\cC_n\) be the concept class from \Cref{eq:decrypt-class-app}, where \(\cC_n\) consists of the decryption functions of Regev's cryptosystem at security parameter \(n\). Assuming the quantum hardness of \(\gapsvp\) or \(\sivp\) against subexponential-time adversaries (\Cref{assump:lattice}), there is no algorithm that runs in time \(2^{o(n)}\) and uses \(2^{o(n)}\) samples that weakly learns the concept class \(\cC\).
\end{lemma}
\begin{proof}
This is an immediate corollary of \Cref{prop:crypto-learning}, \Cref{cor:regev's-crypto-safe-aganist-subexp}, and the fact that the decryption error rate is \(2^{-\Omega(n \log n)}\). Formally, recall that the ciphertexts of the cryptosystem have length \(N(n)=\Theta(n\log p)=\Theta(n\log n)\). Now suppose there is a weak learning algorithm for \(\cC\) that runs in time \(\tau(N(n))=2^{o(n)}\). Then
\( 
\tau(N(n))\epsilon(N(n))
=
2^{o(n)}\cdot 2^{-\Omega(n\log n)}
=
2^{-\Omega(n\log n)}=2^{-\Omega(N)}
<
N^{-\omega_N(1)}.
\)
Hence, by \Cref{prop:crypto-learning}, this would yield a distinguisher for the cryptosystem running in time \(O(\tau(N(n)))=2^{o(n)}\). This is a contradiction under \Cref{assump:lattice} in light of \Cref{cor:regev's-crypto-safe-aganist-subexp}, and we are done.
\end{proof}

\end{document}